\documentclass[lettersize,journal]{IEEEtran}
\usepackage{amsmath,amsfonts}
\usepackage{algorithmic}
\usepackage{algorithm}
\usepackage{array}
\usepackage{caption}
\usepackage[labelfont=sf, textfont=sf]{subcaption}
\usepackage{textcomp}
\usepackage{stfloats}
\usepackage{url}
\usepackage{verbatim}
\usepackage{graphicx}
\usepackage{cite}
\usepackage{soul}
\newcommand{\refframe}[1]{\mathcal{#1}}
\renewcommand{\vec}[1]{\boldsymbol{#1}}
\newcommand{\unitvec}[1]{\hat{\vec{#1}}}
\newcommand{\mat}[1]{\mathbf{#1}}
\newcommand{\tpose}{\mathsf{T}}
\newcommand{\vecframe}[2]{\vec{#1}^{\refframe{#2}}}
\newcommand{\rotation}[2]{\mat{R}_{\refframe{#1}}^{\refframe{#2}}}

\DeclareMathOperator{\diag}{diag}

\begin{document}

\title{Learning robotic milling strategies based on passive variable operational space interaction control}

\author{Jamie Hathaway$^{1,2,\dagger}$, Alireza Rastegarpanah$^{1,2, \dagger}$, Rustam Stolkin$^{1,2}$
\thanks{1 Department of Metallurgy \& Materials Science, University of Birmingham, Birmingham, UK, B15 2TT.}
\thanks{2 The Faraday Institution, Quad One, Harwell Science and Innovation Campus, Didcot, UK, OX11 0RA.}
\thanks{$\dagger$ These authors contributed equally to this work.}
\thanks{This research was funded by the project called  ``Reuse and Recycling of Lithium-Ion Batteries'' (RELiB) [RELIB2 grant number FIRG005, RELIB3 grant number FIRG057], and the project called ``Research and development of a highly automated and safe streamlined process for increased Lithium-ion battery repurposing and recycling'' (REBELION) [grant number 101104241].}
\thanks{The authors would like to acknowledge Ali Aflakian, Carl Meggs and Christopher Gell respectively for assistance with experimental validation, design of material holder and cutter tool for experiments herein.}
\thanks{Manuscript received October 31, 2022; revised March 22, 2023; accepted May 18, 2023.}
\thanks{© 2023 IEEE.  Personal use of this material is permitted.  Permission from IEEE must be obtained for all other uses, in any current or future media, including reprinting/republishing this material for advertising or promotional purposes, creating new collective works, for resale or redistribution to servers or lists, or reuse of any copyrighted component of this work in other works.}}

\markboth{IEEE Transactions on Automation Science \& Engineering}{Hathaway \emph{et al.}: Learning robotic milling using passive OSC}


\maketitle

\begin{abstract}
This paper addresses the problem of robotic cutting during disassembly of products for materials separation and recycling. Waste handling applications differ from milling in manufacturing processes, as they engender considerable variety and uncertainty in the parameters (e.g. hardness) of materials which the robot must cut. To address this challenge, we propose a learning-based approach incorporating elements of interaction control, in which the robot can adapt key parameters, such as feed rate, depth of cut, and mechanical compliance during task execution. We show how a mathematical model of cutting mechanics, embedded in a simulation environment, can be used to rapidly train the system without needing large amounts of data from physical cutting trials. The simulation approach was validated on a real robot setup based on four case study materials with varying structural and mechanical properties. We demonstrate the proposed method minimises process force and path deviations to a level similar to offline optimal planning methods, while the average time to complete a cutting task is within 25\% of the optimum, at the expense of reduced volume of material removed per pass. A key advantage of our approach over similar works is that no prior knowledge about the material is required.
\end{abstract}

\def\abstractname{Note to Practitioners}
\begin{abstract}
This work is motivated by challenges in emerging fields such as recycling of electric vehicles, where products such as batteries adopt a range of designs with varying physical geometry and materials. More generally, this applies when considering robotic disassembly of any unknown component where semi-destructive operations such as cutting are required. Product-to-product variation introduces challenges when planning cutting processes required to disassemble a component, as contemporary planning approaches typically require advance knowledge of the material properties, shape and desired path to select tool speed, feed and depth of cut. In this paper, we show a mathematical model of milling force embedded in a simulation environment can be used as a relatively inexpensive approach to simulate a broad spectrum of cutting processes the robot may encounter. This allows the robot to learn from experience a strategy that can select these key parameters of a milling task online without user assistance. We develop a framework for controlling a robot using this strategy that allows the stiffness of the robot arm to be modulated over time to best satisfy metrics of productivity (e.g. required cutting time), while maintaining safe interaction of the robot with its environment (e.g. by avoiding force limits), similarly to how a human operator can vary muscular tension to accomplish different tasks. We posit that the proposed method can substitute a trial-and-error strategy of selecting process parameters for disassembly of novel products, or integrated with existing planning approaches to adjust the parameters of milling tasks online.
\end{abstract}


\section{Introduction}
\IEEEpubidadjcol
\IEEEPARstart{S}{emi}-destructive disassembly processes such as cutting feature extensively in numerous applications, including end of life product disassembly, nuclear decommissioning, earthquake/disaster response, demolition with roboticised construction site machinery, or even applications to robotic surgery, in which tissue can have variable properties e.g. as a blade passes through muscle, fat, ligaments and connective tissue. For robotic disassembly of unknown products, challenges are presented in developing appropriate process plans due to extensive variations in the target environment, owing to a variety of object models, conditions and materials. While works such as \cite{CognitiveRoboticsBasic} aim to address uncertainty on a product level by altering product and operation-level plans, there remain difficulties in adapting plans for individual processes to uncertainties in the environment on a case-by-case basis. Frequent revisions to the original plan are also required due to uncertainties in initial process parameter estimates, of the component identification or product-to-product variation \cite{CognitiveRoboticsRevision}, but also to handle new product models while maintaining generality to older models, covering potentially decades of design iterations. Handling uncertainty on a process level remains a challenge that is addressed by few works \cite{DisassemblyRoboticsReview}. In particular, destructive tasks such as cutting are necessary, either as recourse if identification or removal of fasteners is impossible, or if the design prohibits non-destructive disassembly. While this extends to a wide range of product types, this is of particular interest in the field of disassembly of electric vehicle (EV) batteries due to the notable lack of standardisation, sensitivity of information regarding battery designs, and limited design for disassembly. 

In this work, we consider cutting using a rotary machine tool for disassembly as a subset of the more general family of milling processes, which implies separation, rather than shaping of material. Notably, the requirements of cutting, or milling for disassembly applications contrast with those of manufacturing, which are carried out in controlled environments, motivated by stringent limits on dimensional tolerance. For disassembly, the precise cutting path is less important, however, variation between products imposes much greater demands on the flexibility of the system to select appropriate process parameters, such as feed rate, depth of cut and tool speed. Simultaneously, this variation results in advance knowledge, such as product specifications, models, geometry and materials becoming difficult to obtain in a disassembly context, complicating the use of offline process planning approaches. Although previous works such as \cite{TommasoCuttingPath} and \cite{rastegarpanah2021vision} aim to address issues of path planning and interaction with uncertain environments, selection of these process parameters online remains largely unaddressed.

Learning-based approaches have proven to be effective at accomplishing a wide range of tasks in unstructured or unknown environments. These have been demonstrated extensively for various interaction control tasks \cite{ComplianceControlPegInHoleRL, HybridTrajectoryForceLearningAssembly, PegInHoleDDPGVIC, SimToRealDomainRandomisationPushingTask}, however, applications for destructive tasks remain limited. In particular, the advantages of randomised simulations for reducing overhead of costly data collection, while learning robust control strategies over a distribution of potential environments are compelling for addressing uncertainty in tasks such as milling.

This paper proposes a domain generalisation approach to learning cutting tasks based on a mechanistic model-based simulation framework. Leveraging the success of approximate Model Predictive Control (MPC) and reinforcement learning (RL) for manipulation applications in unknown environments, we propose a zero-shot system for optimising a cutting task in the context of robotic disassembly of unknown single-material components, such as removing a cover from an electric vehicle battery module or separation of nuclear waste. In addition, we address limitations of variable operational space control (OSC) in a RL-based manipulation context by combining RL with passivity-based control to ensure the closed-loop stability of the controlled system in the sense of Lyapunov. In contrast with previous schemes that guarantee the stability of the controlled system with policy in the case of RL, the proposed method is independent of learning strategy and thus can be employed even if the policy has already been trained, simply by applying the proposed modifications to the existing OSC strategy. Moreover, the proposed controller can optionally be incorporated into the training process, allowing the agent to learn to manage the tank energy if passed to the agent as observations.

The remainder of the paper is structured as follows: in Section \ref{sec:Related-Works} we enumerate previous studies in the area of robotic and Computer Numerical Control (CNC) milling, relating this to state-of-the-art approaches for interaction control. Section \ref{sec:Method} introduces the contemporary milling force modelling approach and proposed operational space control framework based on energy tanks (ET-OSC). This is then related to the overall framework for learning a milling task over a wide domain of materials, before evaluation of the modelling and framework in Section \ref{sec:Results}. Section \ref{sec:Conclusion} concludes the paper.

\section{Related work\label{sec:Related-Works}}
\subsection{Robotic Milling \& Milling Parameter Optimisation}

In recent years, research into using industrial robots for subtractive operations such as milling, drilling and grinding has gained much attention, particularly in the sphere of manufacturing. Such applications are driven by a demand for low-volume, highly flexible production with high dimensional tolerance. It is thus unsurprising that a majority of research in this area explores increasing the process capability of industrial robotics through dimensional compensation for the passive compliance of the tool-robot system \cite{StiffnessTrajectoryOptimisationMillingRobot, PostureOptimisationIndustrialMachiningRobot, ModelBasedPlanningMachiningRobot} and compensation for chatter instability \cite{RoboticMillingChatterStability}. Few works, such as \cite{RoboticMilling3DPointCloud} consider handling uncertainty in robotic milling applications. In a disassembly setting, this uncertainty is a considerable challenge, however, dimensional accuracy of such processes takes a lower priority, with performance metrics shifting towards productivity, lower energy consumption and tool wear. Coincidentally, disassembly workstation concepts are incorporating human-robot collaboration \cite{RoboticDisassemblyEVHRC}, which implies a shift towards lower payload robots equipped with force and torque sensing capabilities, or ``collaborative'' robots, where the torque capabilities of the robot and safety are more significant for selection of milling strategy \cite{FeasabilityMachiningLowPayloadRobot}.

Definition of a successful milling task is further dependent on selection of appropriate process parameters, however, is complicated by such uncertainty. Most works consider parameter selection in CNC -- as opposed to robotic -- applications. In \cite{OptimisationMillingAutomatic} an automatic approach for offline milling parameter global optimisation is presented. Due to the nature of the optimisation approach, prior knowledge of the material in the form of computer aided design (CAD) models and materials datasets are required, limiting the effectiveness of the approach for disassembly applications, in addition to computational overhead from the optimisation process. In \cite{TommasoCuttingPath, TommasoPathPlanningNonHolonomic} the problem of motion planning for robotic cutting in dismantling operations is addressed (with an application to nuclear decommissioning) -- however, they do not address the separate problem of online control of the manipulator against perturbations, caused by forceful interactions between the robot's end-effector (EE) tool and the workpiece. In this paper, we address this problem by enabling the robot to adaptively select key parameters of the cutting process, online during cutting. Separately, the problem of online process parameter selection has inspired learning-based approaches. For example, a meta-reinforcement-learning (RL) approach is proposed in \cite{MetaRLMachiningTurning} incorporating multiple objectives and safety constraints. However, some level of prior knowledge is still required due to the need for predictive models for tool wear and power profiles of the process. In \cite{ContourErrorCNCLearningRL} a combination of RL and learned contour error prediction model was employed for reducing contouring errors during a CNC milling process. Beyond learning for control of milling processes, mechanistic modelling approaches have been employed as a means of efficient data collection for training predictive force models \cite{CuttingModelMachineLearning}. Such modelling approaches have been proposed in a robotic context in \cite{CuttingModelRoboticMilling}. In \cite{ModelBasedPlanningMachiningRobot}, this is applied directly to an industrial robot to guide the optimal workpiece placement for dimensional compensation of a milling task. Similarly, \cite{RoboticMillingGPUVoxel} employs a novel voxel-based simulation approach for dimensional compensation. However, all of these methods still require CAD models of the desired workpiece. In the following section, we describe how such simulations can be employed using a learning-based approach to optimise a cutting task over a generalised domain of materials.

\subsection{Learning \& Interaction Control For Contact-Rich Tasks}
In simulation, complex tasks can be broken down into simpler tasks allowing them to be learned in stages. Furthermore, the cost of data collection in simulation is relatively small - as compared to learning from many real cutting experiments, which may be prohibitive. Improving domain generalisation is therefore crucial for tasks with complex dynamics, dependent on a wide range of process parameters, or tasks that are prohibitive to learn directly in the target domain due to their challenging or destructive nature. Many recent works aim to resolve this problem using domain randomisation (DR) approaches. DR is a well-known domain generalisation technique that aims to achieve zero-shot transfer to a target domain by continually varying the parameters of the source domain. Therefore, once transferred to a target domain, the agent is robust to the differing task dynamics in the real world, assuming the approximation of the task dynamics in the source domain is sufficiently accurate. However, successful applications of DR thus far have been demonstrated mainly for non-destructive tasks \cite{SimToRealDomainRandomisationPushingTask, ComplianceControlPegInHoleRL}.

Learning-based approaches to contact-rich tasks are common in literature due to their applicability to a wide range of problems. A deep RL-based approach to hybrid position-force control for robotic assembly tasks was proposed in \cite{ComplianceControlPegInHoleRL}, and similarly in \cite{HybridTrajectoryForceLearningAssembly}. Furthermore, previous work \cite{OrtenziProjectedOpSpace, OrtenziInverseDynamicsContact, OrteziHybridPostionForceReview} has explored hybrid force-position control in the robot's operation space. This has application to simultaneously controlling e.g. the path of a cutting tool across a surface, and the contact force applied against the surface. However, this requires accurate task modelling due to partitioning of the control space into position and force controlled directions. Later work \cite{OrtenziVisionConstraintEst, OrtenziProprioceptionConstraintEst} explored the use of computer vision and proprioception to obtain information about the robot's configuration constraints and the surface which the robot is contacting. 

Alternatively, \cite{VariableImpedanceActionSpaceRL} explore the problem of learning interaction control for a range of manipulation tasks through a variable impedance control (VIC)-based action space. VIC operates by imposing a desired dynamic behaviour on the robot which is assumed during interaction with unknown environments. However, guaranteeing the stable interaction of the robot with the environment is a challenging topic for learning-based control. Works specific to reinforcement learning include \cite{ActorCriticRLStabilityGuarantee, StabilityCertifiedRLControlTheory}. Notably, these proposed approaches require incorporation directly into the learning process. Outside of RL, this is addressed in \cite{EnergyTankVaribleImpedanceControl, EnergyTankVICTeleoperation} using the concept of energy tanks (ETs). In \cite{PassivityDSLearningCutting}, this is applied to guarantee stability for a cutting task, leveraging a passivity-based DS (dynamical systems) controller which allows temporary violation of passivity conditions without compromising the overall closed-loop system stability. In \cite{PassivityBasedVICRedundantRobot} an energy-tank-based VIC for redundant manipulators was proposed, allowing implementation of desired impedance behaviour in both operational space and null space.

Related to VIC, a key advantage of the operational space control paradigm \cite{KhatibOperationalSpaceControl} employed in an RL context \cite{VariableImpedanceActionSpaceRL} is that the operational space and null space dynamics are decoupled using the concept of the inertia-weighted psuedoinverse \cite{OperationalSpaceControlComparison}. This decoupling allows the null space to be exploited for purposes such as postural adjustment and maximising manipulability (controllability) of the robot along a desired path without impeding the operational space objectives (such as following the path itself). By leveraging the applicability of the energy tank approach to learning-based control, the problem of guaranteeing stability of the learned policy can be addressed with modifications to the \emph{operational space} interaction control strategy in a manner that is not only applicable during the learning process, but also pre-trained policies. Furthermore, the ET-based formalism can be related to the concept of energy budgets for human-robot interaction \cite{EnergyBudgetSafeHRI}, which provides a natural framework within which robot and coordinate-agnostic safety constraints can be expressed.

\section{Methodology\label{sec:Method}}
In this section, we show how the contemporary mechanistic cutting force model can be embedded in a simulation environment. We introduce the proposed energy-tank based operational space controller (ET-OSC) and demonstrate application of the modified OSC law results in a passive closed-loop system, which provides a stability guarantee for interaction with any passive environment. Finally, we demonstrate how this is employed in an RL context to provide a framework to learn a joint process parameter and interaction control strategy for cutting tasks.
\subsection{Milling simulation\label{sec:Method-Mechanistic-Model}}

The mechanistic milling force model \cite{TheMachiningOfMetals} is a semi-empirical model that aims to relate forces acting on a milling tool to the cross-sectional area of undeformed material removed over each revolution of the tool spindle through empirically determined mechanistic constants (labelled $\vec{K}_{c}, \vec{K}_{e}$). It assumes a homogeneous workpiece with isotropic mechanical properties. It has a number of properties that provide the basis for a useful learning environment:
\begin{itemize}
	\item Low computational complexity (allowing faster than real-time simulation and hence expeditious data collection).
	\item The dynamic behaviour of the environment is controlled by a small set of well-defined parameters.
	\item The accuracy of the model is well-validated within the model assumptions.
\end{itemize}
Typically a generalised tool model is discretised to a series of $N_f$ flutes and $N_d$ discs. We furthermore adopt the assumption from \cite{CuttingForceModelArbitraryFeedEngagement} that the values of $\vec{K}_{c}, \vec{K}_{e}$ are constant over all elements despite the varying oblique \& rake angles. Nonetheless, differing values of $\vec{K}_{c}, \vec{K}_{e}$ encompass wear of the milling tool and changes in the material mechanical properties.

\begin{figure}
	\centering
    \begin{subfigure}[t]{0.49\columnwidth}
        \centering
		\includegraphics[width=\textwidth]{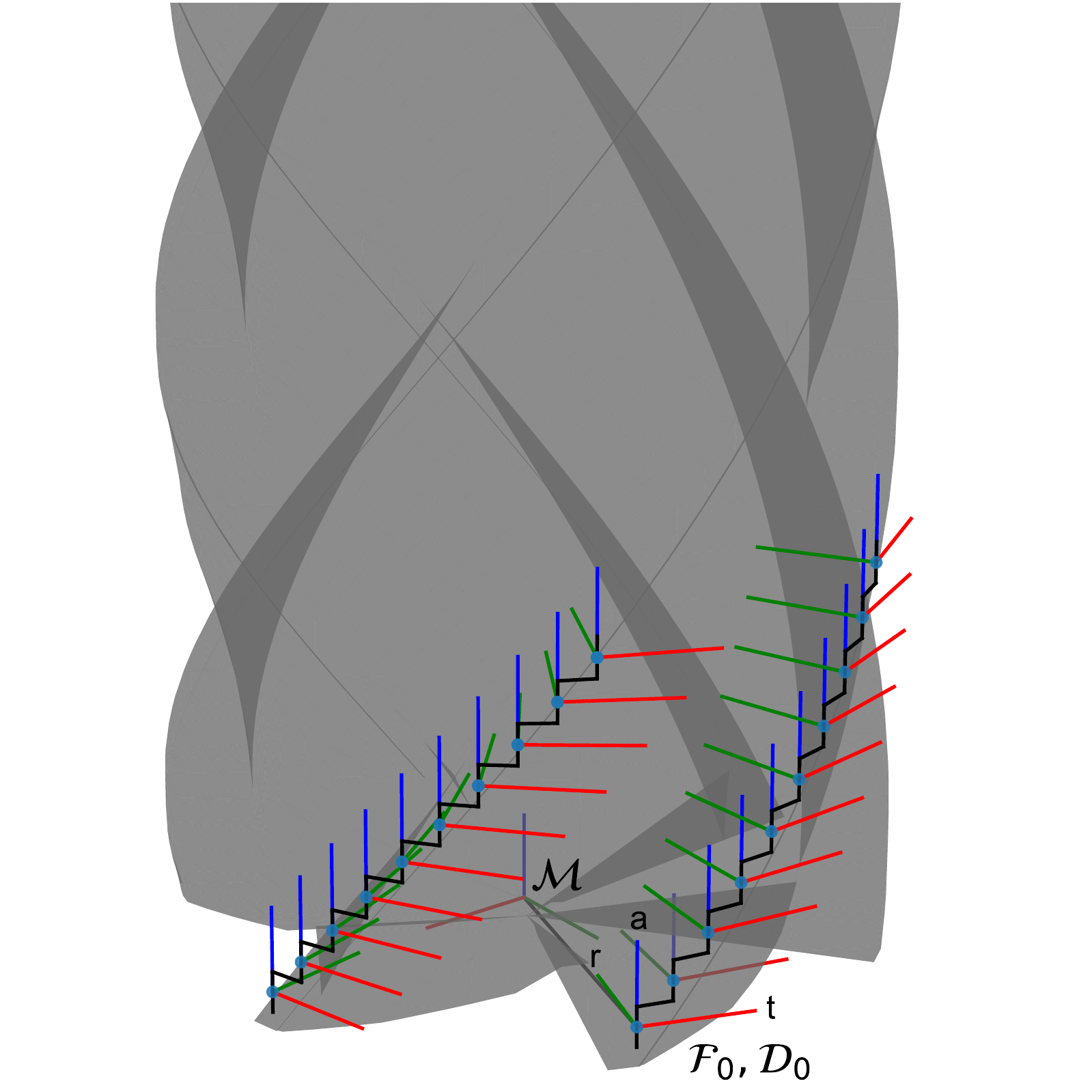}
        \caption{Six-flute end mill}
		\label{fig:casestudy0}
    \end{subfigure}
    \begin{subfigure}[t]{0.49\columnwidth}
        \centering
		\includegraphics[width=\textwidth]{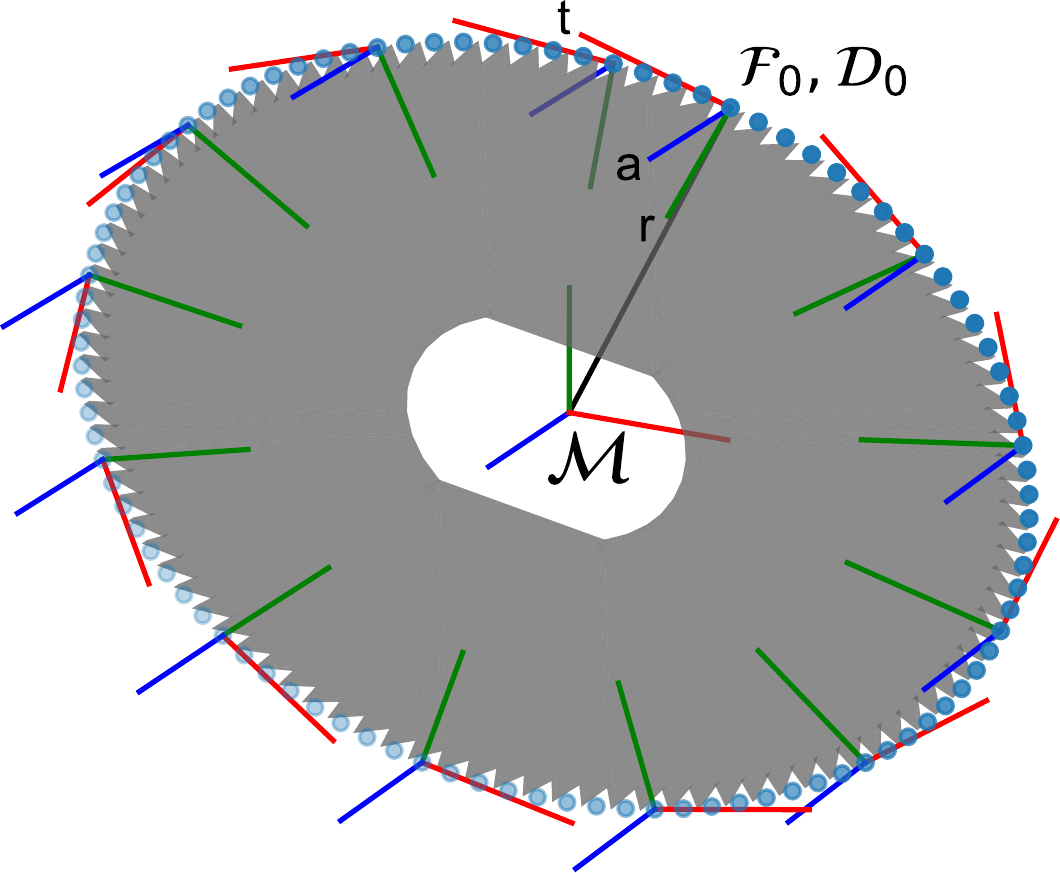}
        \caption{Slitting saw}
		\label{fig:casestudy1}
    \end{subfigure}
	\caption{Discretisation scheme of tool geometry showing model and flute coordinate systems associated with discrete disk and flute elements for end mill (left) and slitting saw tool (right). For clarity, not all flute frames are visualised.}
	\label{fig:Method-EndMillFrames}
\end{figure}

The mechanistic model defines a frame of reference ($\refframe{M}$) situated at the lower axial face of the tool, which is shown along with the discretisation scheme in Figure \ref{fig:Method-EndMillFrames}. Based on the feed rate of the tool in the world frame ($\refframe{W}$) $\vecframe{v}{W}$ and world-to-model transform $\rotation{W}{M}$, the model frame feed rate is
\begin{equation}
	\vecframe{v}{M}=\rotation{W}{M}\vecframe{v}{W}
\end{equation}
hence, by definition of the material feed per tooth $\vecframe{f}{M}$:
\begin{equation}
	\vecframe{f}{M}=\frac{1}{N_{f}\omega}\vecframe{v}{M}
\end{equation}
where $\omega$ is the spindle speed in s$^{-1}$. Adopting the approach from \cite{CuttingForceModelArbitraryFeedEngagement}, for each flute and disc element, indexed by $f$, $d$ respectively, the thickness of undeformed chip removed $h^\refframe{F,D}$ can be computed as the feed per tooth projected along the radial direction of each flute with associated frame $\refframe{F,D}$:
\begin{equation}
	h^{\mathcal{F},\mathcal{D}}=\begin{bmatrix}0 & -1 & 0\end{bmatrix} \rotation{M}{F,D}\vecframe{f}{M}
\end{equation}
This is a generalisation of the simpler circular tool path approximation approach, the latter assuming uniaxial feed of the tool along the positive $\unitvec{x}$ direction in $\refframe{M}$. The transformation associated with each cutting element is related to the element angle $\theta_{f,d}^{\refframe{M}}$ (defined clockwise from $\unitvec{y}$ in $\refframe{M}$ about $\unitvec{z}$):
\begin{equation}
\mathbf{R}_{\mathcal{F},\mathcal{D}}^{\mathcal{M}}=\begin{bmatrix}-\cos\theta_{f,d}^{\refframe{M}} & -\sin\theta_{f,d}^{\refframe{M}} & 0\\
\sin\theta_{f,d}^{\refframe{M}} & -\cos\theta_{f,d}^{\refframe{M}} & 0\\
0 & 0 & 1
\end{bmatrix}
\end{equation}
The geometry of a general fluted end-mill or slitting saw can be constructed as \cite{CuttingModelMachineLearning}:
\begin{equation}
    \theta_{f,d}^{\refframe{M}}=\Theta+f\phi-\left(d+\frac{1}{2}\right)\frac{\tan(\varphi)b_{f,d}}{R}
\end{equation}
where $\Theta$ is the tool rotation angle in $\refframe{W}$, $\phi$ is the pitch angle, $\varphi$ is the helix angle, $R$ is the tool radius, and $b_{f,d}$ is the length of cutting edge. For slitting saw and end-mill tools this coincides with the height of each discrete disc element along $\hat{\vec{z}}$ in $\mathcal{M}$. The overall cutting force on each flute $\vecframe{F}{F,D}$ can then be considered as the sum of chip cross-sectional area and edge-dependent components:
\begin{equation}
	\boldsymbol{F}^{\mathcal{F},\mathcal{D}}=b_{f,d}\begin{bmatrix}K_{e,t}\\
		K_{e,r}\\
		K_{e,a}
	\end{bmatrix}+b_{f,d}\begin{bmatrix}K_{c,t}\\
		K_{c,r}\\
		K_{c,a}
	\end{bmatrix}h^{\mathcal{F},\mathcal{D}}
	\label{eq:Method-Flute-Force}
\end{equation}
with $t, r, a$ denoting the components of $\vec{K}_{c}$, $\vec{K}_{e}$ over radial, tangential and axial directions respectively. Hence, the total model frame cutting force is the sum over all engaged cutting elements as
\begin{equation}
	\boldsymbol{F}^{\mathcal{M}}=\sum_{f}^{N_{f}}\sum_{d}^{N_{d}} G_{f,d}\mathbf{R}_{\mathcal{F},\mathcal{D}}^{\mathcal{M}}\boldsymbol{F}^{\mathcal{F},\mathcal{D}}
	\label{eq:Method-Mechanistic-Total-Force}
\end{equation}
$\mat{G}\in\mathbb{B}^{N_{f}\times N_{d}}$ is a Boolean matrix depending on if the element $f,d$ is engaged with the workpiece. Intersection is computed using a Boolean geometry workpiece model, allowing efficient query of the intersection state of a given flute. 

\subsection{Passive operational space control}

Consider the Euler-Lagrange equation for the dynamics of a rigid N-degree-of-freedom (DoF) manipulator in operational space with joint positions $\vec{q}$:
\begin{equation}
	\mat{\Lambda}(\vec{q})\ddot{\vec{x}}+\mat{\Gamma}(\vec{q},\dot{\vec{q}})+\vec{\mu}(\vec{q}) = \vec{F}_{c}+\vec{F}_{e}
	\label{eq:Method-Euler-Lagrange-Dynamics}
\end{equation}
where $\mat{\Lambda}$, $\mat{\Gamma}$, $\vec{\mu}$ are the operational space analogues to the inertia matrix, Coriolis and centrifugal force matrix and gravitational force vector respectively. $\vec{F}_{c}$ and $\vec{F}_{e}$ are the applied control wrench and external wrench apparent at the end-effector (EE) expressed in $\refframe{W}$. The control law for operational space control (OSC) with time-varying stiffness $\mat{K}_{p}(t)$ and damping $\mat{K}_{d}(t)$ (with difference between current and desired EE pose $\vec{e}=\vec{x}-\vec{x}_{d}$) is expressed as:
\begin{equation}
	\vec{F}_{c} = \mat{\Lambda}(\vec{q})\left[\ddot{\vec{x}}_{d} + \mat{K}_{d}(t)\dot{\vec{e}} + \mat{K}_{p}(t)\vec{e}\right] + \mat{\Gamma}+\vec{\mu}
\end{equation}
Hence the closed loop dynamics of the system:
\begin{equation}
	\ddot{\vec{e}}+\mat{K}_{d}(t)\dot{\vec{e}}+\mat{K}_{p}(t)\vec{e}=\mat{\Lambda}^{-1}\vec{F}_{e}
	\label{eq:Method-Closed-Loop-OSC}
\end{equation}

Under RL-based variable OSC, the time-varying gains are set by the policy and are in general discontinuous. The issue with this setup arises when considering the energy storage function for the system:
\begin{equation}
	V(\vec{e},\dot{\vec{e}})=\frac{1}{2}\dot{\vec{e}}^{\tpose}\mat{\Lambda}\dot{\vec{e}}+\frac{1}{2}\vec{e}^{\tpose}\mat{\Lambda} \mat{K}_{p}\vec{e}
	\label{eq:Method-Energy-Storage-Func-OSC}
\end{equation}
The system's passivity with respect to the input-output pair $\left(\dot{\vec{e}}, \vec{F}_{e}\right)$ is determined by the condition:
\begin{equation}
	\dot{V}(\vec{e},\dot{\vec{e}}) \leq \dot{\vec{e}}^{\tpose}\vec{F}_{e}
	\label{eq:Method-Passivity-Condition}
\end{equation}
From \eqref{eq:Method-Energy-Storage-Func-OSC}, substituting $\ddot{\vec{e}}$ from \eqref{eq:Method-Closed-Loop-OSC}, the power $\dot{V}$ of the system is
\begin{multline}
	\dot{V}(\vec{e},\dot{\vec{e}})=\dot{\vec{e}}^{\tpose}\vec{F}_{e}-\dot{\vec{e}}^{\tpose}\mat{\Lambda}\mat{K}_{d}\dot{\vec{e}}\\ +\frac{1}{2}\dot{\vec{e}}^{\tpose}\dot{\mat{\Lambda}}\dot{\vec{e}}+\frac{1}{2}\vec{e}^{\tpose}\dot{\mat{\Lambda}}\mat{K}_{p}\vec{e}+\frac{1}{2}\vec{e}^{\tpose}\mat{\Lambda}\dot{\mat{K}}_{p}\vec{e}
	\label{eq:Method-Variable-Gain-OSC-Power}
\end{multline}
and it is clear that the latter 3 terms add energy to the system that violates \eqref{eq:Method-Passivity-Condition}.

In general, a loss of passivity affects the stability of the interaction between the robot and environment, and convergence to zero tracking error is no longer guaranteed in free space \cite{EnergyTankVaribleImpedanceControl}. For RL, this is an issue as without constraints on the policy action space, it cannot be guaranteed that $\dot{\mat{K}}_{p}(t)$, $\dot{\mat{\Lambda}}(t)$ meets \eqref{eq:Method-Passivity-Condition}. One approach could be to leverage action space design to restrict the magnitude of $\dot{\mat{K}}_{p}(t)$, however in practice from our preliminary evaluations using action spaces incorporating $\dot{\mat{K}}_{p}(t)$, this resulted in poor training performance.

The philosophy of energy tank (ET)-based control is that the energy dissipated through the system damping acts as a \emph{passivity margin} within which a desired stiffness policy can be implemented. The excess dissipated energy is stored in a virtual, finite energy reservoir. When the desired policy violates \eqref{eq:Method-Passivity-Condition}, the extra energy is instead drained from the tank to implement non-passive control actions, until the tank is depleted. Thus, the overall energy of the closed-loop system remains bounded. In fact, the stability of the system is guaranteed when interacting with any unknown environment, as long as the environment is also \emph{passive}.

Since ET control operates based on the exchange of energy between tank and system, the port-Hamiltonian (PH) approach \cite{PortHamiltonianFirstPaper} is a natural modelling framework for such systems. This has been employed in \cite{RobustAdaptivePassivityPHModelWithRL} and \cite{Michel20215429}, which we extend to OSC. The generic representation of a PH system with state $\vec{x}$, input-output pair $\vec{u}$, $\vec{y}$ and Hamiltonian $H$ coupled to an energy tank with counterpart $x_{t}$, $u_{t}$, $y_{t}$, $H_{t}$ is expressed as follows:
\begin{equation}
	\begin{cases}
		\dot{\vec{x}}=\left[\mat{J}(\vec{x})-\mat{R}(\vec{x})\right]\frac{\partial H}{\partial \vec{x}}+\vec{g}(\vec{x})\vec{u} & \,\\
		\dot{x}_{t}=\frac{\sigma}{x_{t}}D(\vec{x})+\frac{1}{x_{t}}\left(\sigma P_{\mathrm{in}}-P_{\mathrm{out}}\right)+u_{t} & \,\\
		\vec{y}=\vec{g}^{\tpose}(\vec{x})\frac{\partial H}{\partial \vec{x}} & \,\\
		y_{t}=\frac{\partial H_{t}}{\partial x_{t}} & \,
	\end{cases}
	\label{eq:Method-Generic-PH-System}
\end{equation}
where $\mat{J}(\vec{x})$, $\mat{R}(\vec{x})$ are matrices describing the power transfer and dissipation within the system, $\vec{g}(\vec{x})$ the input matrix. $\sigma$ is a gate function controlling the power transferred to the tank through the dissipation $D(\vec{x})$, while $P_{\mathrm{in}}$, $P_{\mathrm{out}}$ describe external inward and outward power flow. To derive the PH representation of the variable-gain system, we adopt the approach from \cite{EnergyTankVaribleImpedanceControl,EnergyTankVICTeleoperation} by defining the desired gains as a sum of constant and time-varying components:
\begin{equation}
	\mat{\Lambda}(t) = \mat{\Lambda}_{c} + \mat{\Lambda}_{v}(t)
\end{equation}
\begin{equation}
	\mat{K}_{p}(t) = \mat{K}_{c} + \mat{K}_{v}(t)
\end{equation}
then the energy storage function (Hamiltonian) of the system with constant and time-varying gains is as follows:
\begin{multline}
V=\frac{1}{2}\vec{p}^{\tpose}\mat{\Lambda}^{-1}_{c}\vec{p}+\frac{1}{2}\dot{\vec{e}}^{\tpose}\mat{\Lambda}_{v}(t)\dot{\vec{e}}\\+\frac{1}{2}\vec{e}^{\tpose}\left(\mat{\Lambda}_{c}+\mat{\Lambda}_{v}(t)\right)\left(\mat{K}_{c}+\mat{K}_{v}(t)\right)\vec{e}
\end{multline}
where $\vec{p}=\mat{\Lambda}_{c}\vec{\dot{e}}$ is the generalised momentum of the system. $\vec{x}=\begin{bmatrix}\vec{e} & \vec{p}\end{bmatrix}^{\tpose}$. Using \eqref{eq:Method-Closed-Loop-OSC}:
\begin{multline}
	\dot{\vec{p}}=\vec{F}_{e}-\mat{\Lambda}(t)\mat{K}_{d}\dot{\vec{e}}-\mat{\Lambda}_{c}\mat{K}_{c}\vec{e}-\mat{\Lambda}_{v}(t)\mat{K}_{c}\vec{e}\\-\mat{\Lambda}(t)\mat{K}_{v}(t)\vec{e}-\mat{\Lambda}_{v}(t)\ddot{\vec{e}}
\end{multline}
the PH representation of the closed loop system is thus:
\begin{equation}
	\begin{cases}
		\begin{bmatrix}
			\dot{\vec{e}} \\
			\dot{\vec{p}}
		\end{bmatrix}=\begin{bmatrix}
			\mat{0} & \mat{I} \\
			-\mat{I} & -\mat{\Lambda}(t)\mat{K}_{d}
		\end{bmatrix}\begin{bmatrix}
			\mat{\Lambda}_{c}\mat{K}_{c}\vec{e} \\
			\mat{\Lambda}_{c}^{-1}\vec{p}
		\end{bmatrix} + \begin{bmatrix} 
			\mat{0} \\
			\mat{I}
		\end{bmatrix}
		\left(\vec{F}_{e}+\vec{w}(t)\right) & \,\\
		\vec{y}=\begin{bmatrix}\mat{0} & \mat{I}\end{bmatrix}\begin{bmatrix}\mat{\Lambda}_{c}\mat{K}_{c}\vec{e}\\
			\mat{\Lambda}_{c}^{-1}\vec{p}
		\end{bmatrix}=\dot{\vec{e}} & \,
	\end{cases}
	\label{eq:Method-Port-Hamiltonian-ET-OSC}
\end{equation}
where
\begin{equation}
	\vec{w}(t) = -\mat{\Lambda}_{v}(t)\mat{K}_{c}\vec{e}-\mat{\Lambda}(t)\mat{K}_{v}(t)\vec{e}-\mat{\Lambda}_{v}(t)\ddot{\vec{e}}
\end{equation}
The energy added to the tank through dissipation can be computed \cite{EnergyTankVICTeleoperation} as
\begin{equation}
    D(x) = \frac{\partial^{\tpose}H}{\partial \vec{x}}\mat{R}(x)\frac{\partial H}{\partial \vec{x}} = \dot{\vec{e}}^{\tpose}\mat{\Lambda}(t)\mat{K}_{d}\dot{\vec{e}}
    \label{eq:Method-Tank-Dissipation}
\end{equation}

A power balance between power port of tank $(u_{t}, y_{t})$ and closed-loop dynamic system $(\vec{u}, \vec{y})$ implies the following equality:
\begin{equation}
	u_{t}^{\tpose}=-\frac{1}{x_{t}}\vec{w}^{\tpose}\vec{y}
	\label{eq:Method-Tank-System-Power-Balance}
\end{equation}
Therefore, from \eqref{eq:Method-Generic-PH-System} with \eqref{eq:Method-Tank-Dissipation} and \eqref{eq:Method-Tank-System-Power-Balance}, the tank dynamics are
\begin{equation}
	\begin{cases}
		\dot{x}_{t}=\frac{\sigma}{x_{t}}\dot{\vec{e}}^{\tpose}\mat{\Lambda}(t)\mat{K}_{d}\dot{\vec{e}}-\frac{1}{x_{t}}\vec{w}^{\tpose}(t)\dot{\vec{e}} & \,\\
		y_{t}=x_{t} & \,
	\end{cases}
	\label{eq:Method-Energy-Tank-Dynamics}
\end{equation}
Setting $\ddot{\vec{x}}_{d}=\vec{0}$, from \eqref{eq:Method-Euler-Lagrange-Dynamics} and \eqref{eq:Method-Port-Hamiltonian-ET-OSC} the energy-tank OSC law (excluding the feedforward terms $\mat{\Gamma}$, $\vec{\mu}$) is thus:
\begin{equation}
	\vec{F}_{c}=-\mat{\Lambda}_{c}\mat{K}_{c}\vec{e}-\mat{\Lambda}(t)\mat{K}_{d}\dot{\vec{e}}+\vec{w}(t)+\mat{\Lambda}_{v}(t)\ddot{\vec{e}}
\end{equation}
However, when $\vec{w}(t)=\vec{0}$, the above control law is inadequate due to estimation noise associated with the $\mat{\Lambda}_{v}(t)\ddot{\vec{e}}$ feedback term. In this case, an alternative control law may be derived:
\begin{equation}
	\vec{F}_{c}=\mat{\Lambda}(t)\left(-\mat{\Lambda}_{c}^{-1}\mat{\Lambda}(t)\mat{K}_{d}\dot{\vec{e}}-\mat{K}_{c}\vec{e}\right)+\mat{\Lambda}_{v}(t)\mat{\Lambda}_{c}^{-1}\vec{F}_{e}
\end{equation}
which recovers the desired closed loop dynamics in the case $\vec{w}(t) = \vec{0}$:
\begin{equation}
	\mat{\Lambda}_{c}\ddot{\vec{x}}+\mat{\Lambda}(t)\mat{K}_{d}\dot{\vec{e}}+\mat{\Lambda}_{c}\mat{K}_{c}\vec{e}=\vec{F}_{e}
\end{equation}
Note the $\mat{\Lambda}_{v}(t)\mat{\Lambda}_{c}^{-1}\vec{F}_{e}$ term implies external force feedback is still required to shape the inertial characteristics of the closed-loop system.

\subsubsection{Passivity of Energy-Tank-Based OSC}
The proof of passivity of the ET-based OSC is now broadly similar to the impedance controller presented in \cite{EnergyTankVICTeleoperation}. Consider the Hamiltonian of the combined tank-robot system:
\begin{align}
	W(\vec{e},\dot{\vec{e}},x_{t}) &= H_{c}(\vec{e},\dot{\vec{e}}) + H_{t}(x_{t})\\&= \frac{1}{2}\dot{\vec{e}}^{\tpose}\mat{\Lambda_c}\dot{\vec{e}} + \frac{1}{2}\vec{e}^{\tpose}\mat{\Lambda}_{c}\mat{K}_{c}\vec{e} + \frac{1}{2}x^{2}_{t}
\end{align}
where $H_{c}$ is the Hamiltonian of the constant gain system from \eqref{eq:Method-Port-Hamiltonian-ET-OSC}. Then, by the same procedure as for \eqref{eq:Method-Variable-Gain-OSC-Power}:
\begin{equation}
	\dot{H}_{c}(\vec{e},\dot{\vec{e}})=-\dot{\vec{e}}^{\tpose}\mat{\Lambda}(t)\mat{K}_{d}\dot{\vec{e}}+\dot{\vec{e}}^{\tpose}\vec{F}_{e}+\dot{\vec{e}}^{\tpose}\vec{w}(t)
\end{equation}
similarly using \eqref{eq:Method-Energy-Tank-Dynamics}:
\begin{equation}
	\dot{H}_{t}(x_{t}) = \dot{x}_{t}x_{t} = \sigma\dot{\vec{e}}^{\tpose}\mat{\Lambda}(t)\mat{K}_{d}\dot{\vec{e}}-\vec{w}^{\tpose}(t)\dot{\vec{e}}
\end{equation}
thus
\begin{equation}
	\dot{W}(\vec{e},\dot{\vec{e}},x_{t})=\left(\sigma-1\right)\dot{\vec{e}}^{\tpose}\mat{\Lambda}(t)\mat{K}_{d}\dot{\vec{e}}+\dot{\vec{e}}^{\tpose}\vec{F}_{e}
\end{equation}
which, noting the design $0 \leq \sigma \leq 1$, satisfies \eqref{eq:Method-Passivity-Condition}.

\subsection{Reinforcement learning\label{sec:Method-Training-Env}}
\begin{figure}
    \centering
    \begin{subfigure}[t]{0.45\columnwidth}
        \centering\includegraphics[width=\textwidth]{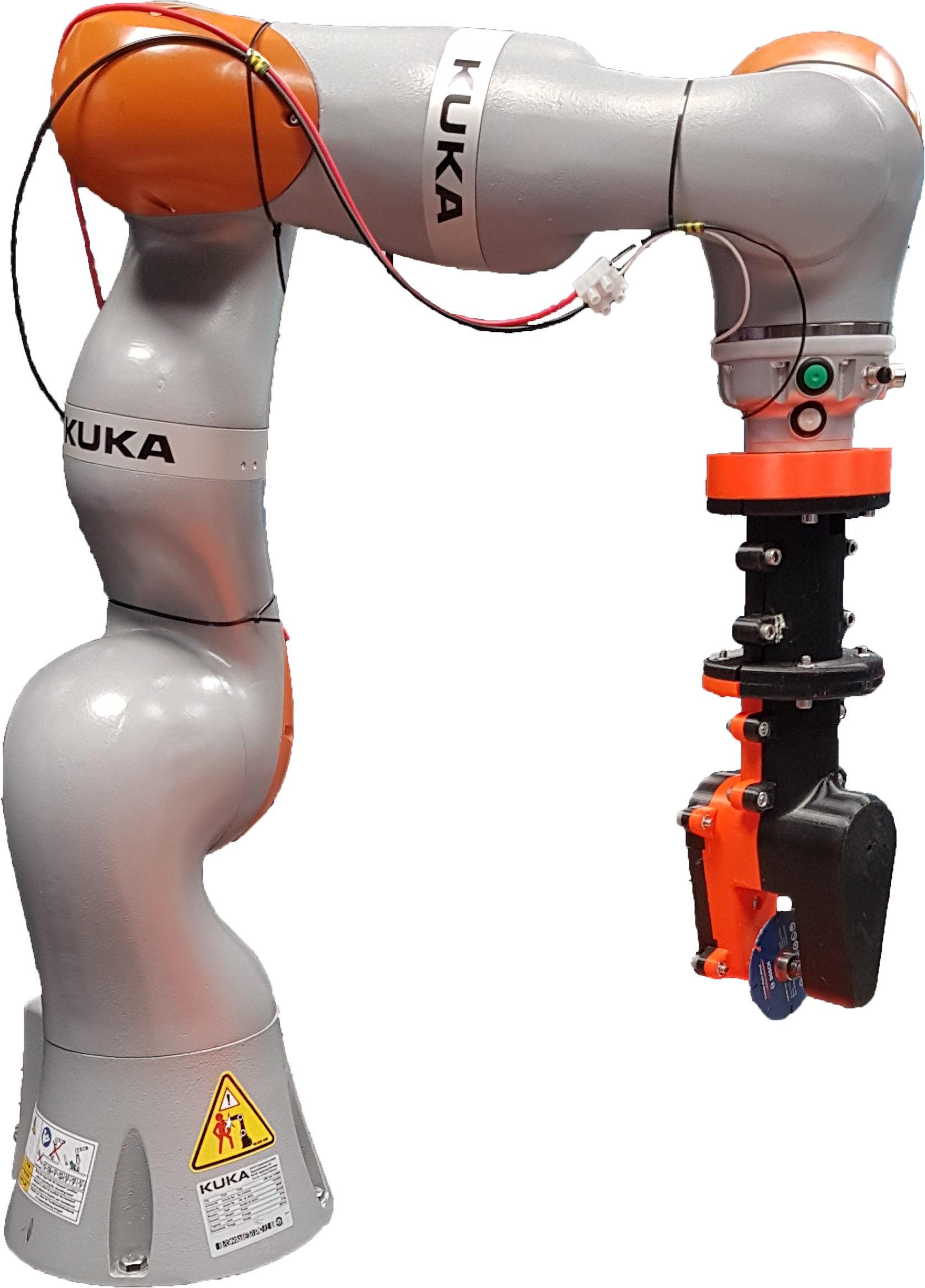}
        \caption{KUKA LBR iiwa R820 with slitting saw tool.}
    \end{subfigure}
    \hfill
    \begin{subfigure}[t]{0.45\columnwidth}
        \centering\includegraphics[width=\textwidth]{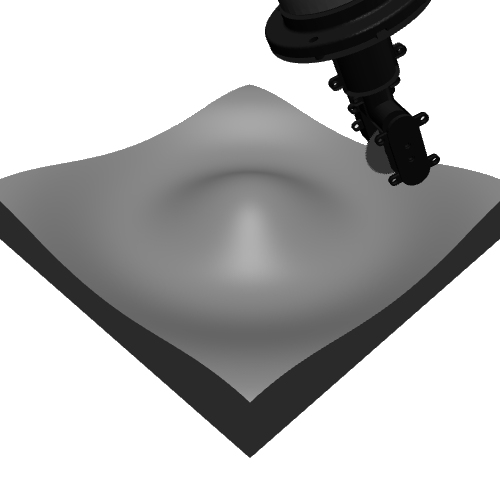}
        \caption{Mujoco simulation with heightfield workpiece.}
    \end{subfigure}
	\caption{Experimental and simulated robotic cutting setup consisting of KUKA LBR iiwa R820 14kg collaborative robot with wrist-mounted motorised slitting saw tool.}
    \label{fig:Method-Experimental-Setup}
\end{figure}

We define a simulation environment using the Mujoco \cite{Mujoco} package based on the modelling framework in section \ref{sec:Method-Mechanistic-Model}, replicating the experimental setup as shown in Figure \ref{fig:Method-Experimental-Setup}. The tool parameters were determined as $R=25$\,mm, $\phi=0.1257$\,rad, $\varphi=0.0$\,rad, $N_f=50$, $N_d=1$, $b_{f,d}=0.5$\,mm, and $\omega=1000$\,rpm. The workpiece was modelled as a heightfield data array interpolated across a bivariate cubic spline surface. Such an approach is chosen over methods with greater complexity and accuracy to minimise computational overhead, as such approaches are prevalent in literature \cite{OptimisationMillingAutomatic, RoboticMillingGPUVoxel}, and the focus of the current work is on learning a generalised control policy over a domain of materials, not on modelling the resulting workpiece geometry / tolerances. Tool paths were generated as NURBS (Non-Uniform Rational B-Spline) curves $\vec{c(t)}$.
The action space for the controller is defined as
\begin{equation}
	\begin{bmatrix}\diag^{-1}\left({\mat{K}_{p}}\right) & \dot{t}_{\Delta} & \dot{n}_{\Delta} \end{bmatrix}^{\tpose}
\end{equation}
which relates to the controller stiffness, and setpoint position $\vec{x}_{d}$, which is adjusted according to the planned path, time and normal offsets ($t_{\Delta}$, $n_{\Delta}$) as:
\begin{equation}
	\vec{x}_{d} = \vec{c}(t+t_\Delta) + n_{\Delta}\hat{\vec{n}}
\end{equation}
where $\hat{\vec{n}}$ is the path normal vector. The observations provided to the agent are
\begin{equation}
	\vec{\xi} = \begin{bmatrix}\dot{\vec{c}}(t)\dot{\vec{x}}^{\tpose} & \vec{e} & \dot{\vec{x}} & \vec{F}_{e} & t_{\Delta} & n_{\Delta} & \diag^{-1}\left({\mat{K}_{p}}\right) \end{bmatrix}^{\tpose}
\end{equation}
while for the ET-based controller, the observation vector is augmented as:
\begin{equation}
	\vec{\xi}_{\mathrm{aug}} = \begin{bmatrix}\vec{\xi} & H_{t}(x_{t}) & \dot{H_{t}}(x_{t})\end{bmatrix}^{\tpose}
\end{equation}

Note that while the material geometry and properties may vary, this information -- including values of $\vec{K}_{c}$, $\vec{K}_{e}$ -- is not provided to the agent at runtime; only the desired reference path is known, which may be adjusted by the agent at runtime. A notable contrast with related works \cite{OptimisationMillingAutomatic,MetaRLMachiningTurning} is that we consider the optimisation of a single milling process in isolation. Typical cognitive robotic approaches to disassembly incorporate trial and error processes of exploration to determine the required disassembly process plan \cite{CognitiveRoboticsBasic, CognitiveRoboticsRevision}, therefore, in general the required processes (e.g. to separate the casing of a component) are not known in advance. Hence, we define the reward function as follows:

\begin{equation}
r = Q_{\textrm{MRV}}\cdot\mathrm{MRV} - Q_{\mathrm{cut}}t_{\mathrm{cut}} - \vec{e}\mat{Q}_{d}\vec{e}^\tpose - \vec{F}_{e}\mat{Q}_{f}\vec{F}_{e}^\tpose
\end{equation}
where the first two reward components are related to productivity of the cutting task, based on: 
\begin{itemize}
	\item Material removed volume $\mathrm{MRV}$ -- computed from the uncut chip thickness, tool engagement and rotational speed of the cutter
	\item Time elapsed per operation $t_{\mathrm{cut}}$
\end{itemize}
These components are weighted by $Q_{\textrm{MRV}}$, the average cost of disassembled product per unit volume of material removed, and the cost of machining time $Q_{\mathrm{cut}}$. For simplicity, we consider only the time elapsed during the machining process and neglect tool changing, setup time and downtime.

The second two reward components are additional terms used to guide the control policy to avoid dangerous or unrealistic actions:
\begin{itemize}
	\item $\vec{e}\mat{Q}_{d}\vec{e}^\tpose$, the weighted sum of the deviation from the desired path setpoint
	\item $\vec{F}_{e}\mat{Q}_{f}\vec{F}_{e}^\tpose$, similarly, for the end-effector wrench acting on the tool
\end{itemize}
$\mat{Q}_{f}$ is a cost weighting selected such that the reduction in cost from increasing MRV is balanced by increasing load on the tool at the maximum permissible process force.
\begin{table}
    \centering
	\caption{Model hyperparameters used with Proximal Policy Optimisation (PPO) algorithm for training the variable OSC policy for a cutting task. ``MLP'' refers to a multi-layer perceptron. Network architecture is represented as a list of hidden layer sizes.}
	\begin{tabular}{|l|l|l|}
		\hline
		Hyperparameter & Value & Search space \\
		\hline
		Learning rate (LR) & 3$\times$10$^{-4}$ & $10^{-5}$--$10^{-3}$ \\
		LR half-life (as ratio of total timesteps) & 0.25 & 0.125--0.5 \\
		Batch size & 1024 & 256--2048 \\
		Discount factor & 0.99 & 0.9--0.99 \\
		Actor/critic network & MLP & --- \\
		Actor/critic network architecture & [64, 64] & --- \\
		\hline
	\end{tabular}
	\label{tab:Method-Hyperparameters}
\end{table}

Policies were learned using the Proximal Policy Optimisation (PPO) learning algorithm \cite{PPOPaper}. PPO is an on-policy actor-critic policy gradient algorithm which employs a clipped objective function to constrain the magnitude of policy parameter updates. It features improvements over other policy gradient algorithms such as Deep Deterministic Policy Gradient (DDPG), as it is resistant to the so-called ``catastrophic forgetting'' problem. Compared to state-of-the-art off-policy algorithms such as Twin-Delayed DDPG (TD3) or Soft Actor-Critic (SAC) it exhibits faster convergence for low-dimensional problems, however is comparatively less sample efficient---in the presented problem formulation, this is an acceptable trade-off due to the low computational complexity of the simulation. The training hyperparameters were informed by manual search summarised in Table \ref{tab:Method-Hyperparameters}. To improve training performance, reward normalisation and rolling average observation normalisation were employed. Additionally, domain randomisation is applied to make the trained agent robust to variations in the tool and workpiece properties, which further aids \emph{domain generalisation} when aiming to transfer the developed policy to new domains, such as the real world, or different selections of milling tool. The workpiece geometry is regenerated at the beginning of each training episode, providing a wide range of surface geometries with different tool-workpiece engagement profiles. The mechanistic constants are sampled from a random uniform distribution informed from values obtained in literature and preliminary experiments.

\section{Results \& Discussion \label{sec:Results}}
In this section we validate the proposed environment for learning cutting tasks based on collection of real world cutting data. Then, the effectiveness of the proposed ET-OSC vs. traditional OSC is compared while carrying out a cutting task. Finally, the performance of the proposed framework in simulation is evaluated and compared with a state-of-the-art efficient global optimisation (EGO) strategy.

\subsection{Real world model validation}
\begin{figure}
    \centering\includegraphics[width=0.95\columnwidth]{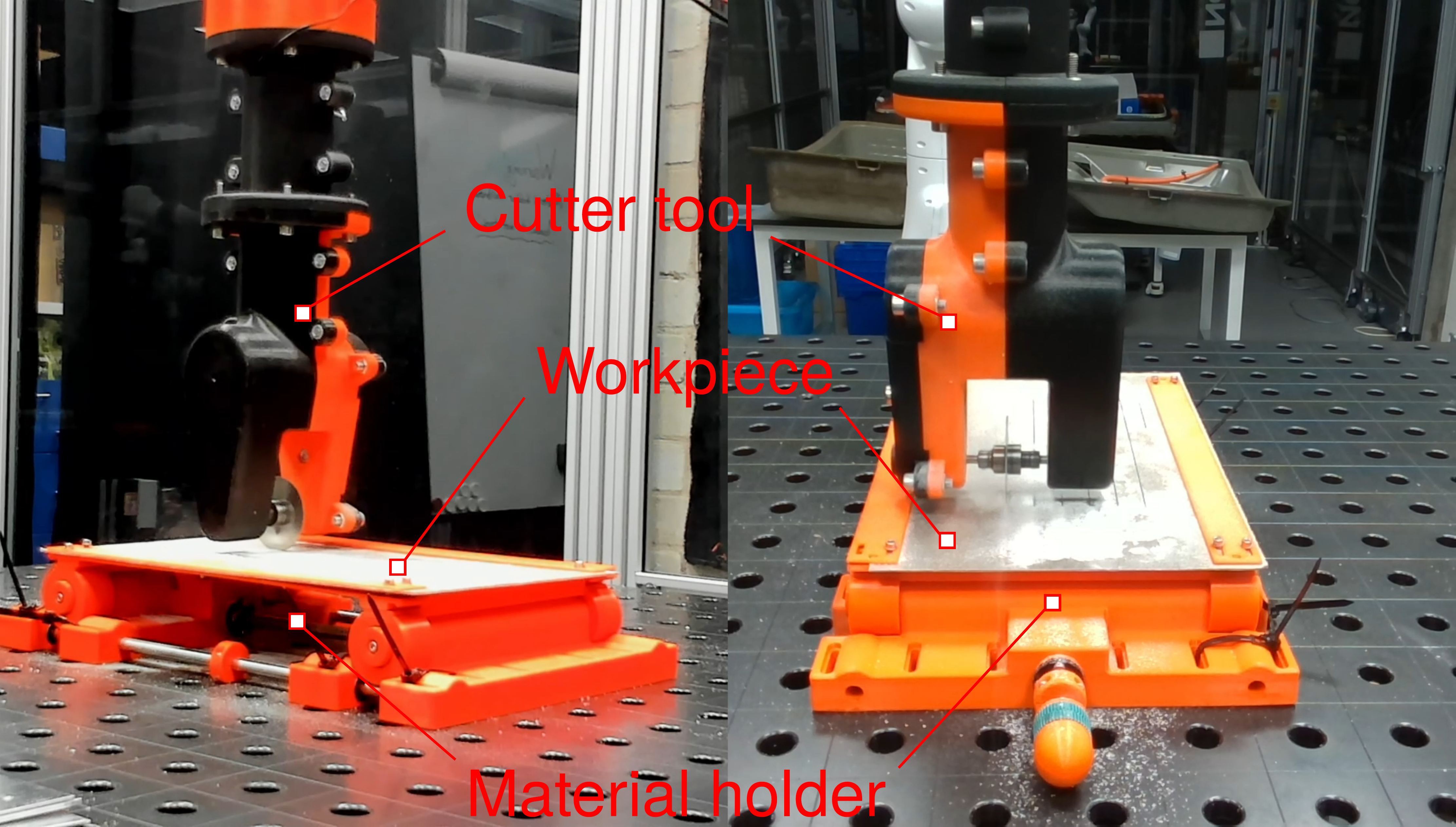}
    \caption{Overview of experimental setup during real world cutting task on mica sheet.}
    \label{fig:Results-Cutting-Example}
\end{figure}

\begin{figure}
    \centering
    \begin{subfigure}[t]{0.49\columnwidth}
        \centering\includegraphics[width=0.4\textwidth]{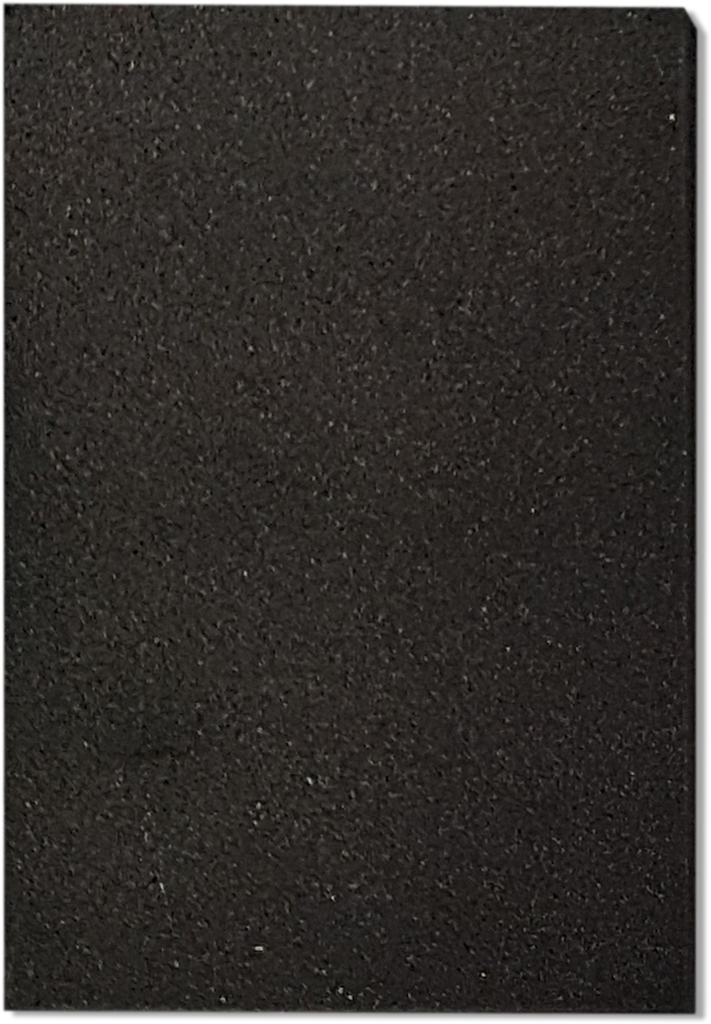}
        \caption{Polyurethane (PU) foam}
        \label{Results-Materials-Foam}
    \end{subfigure}
    \begin{subfigure}[t]{0.49\columnwidth}
        \centering\includegraphics[width=0.4\textwidth]{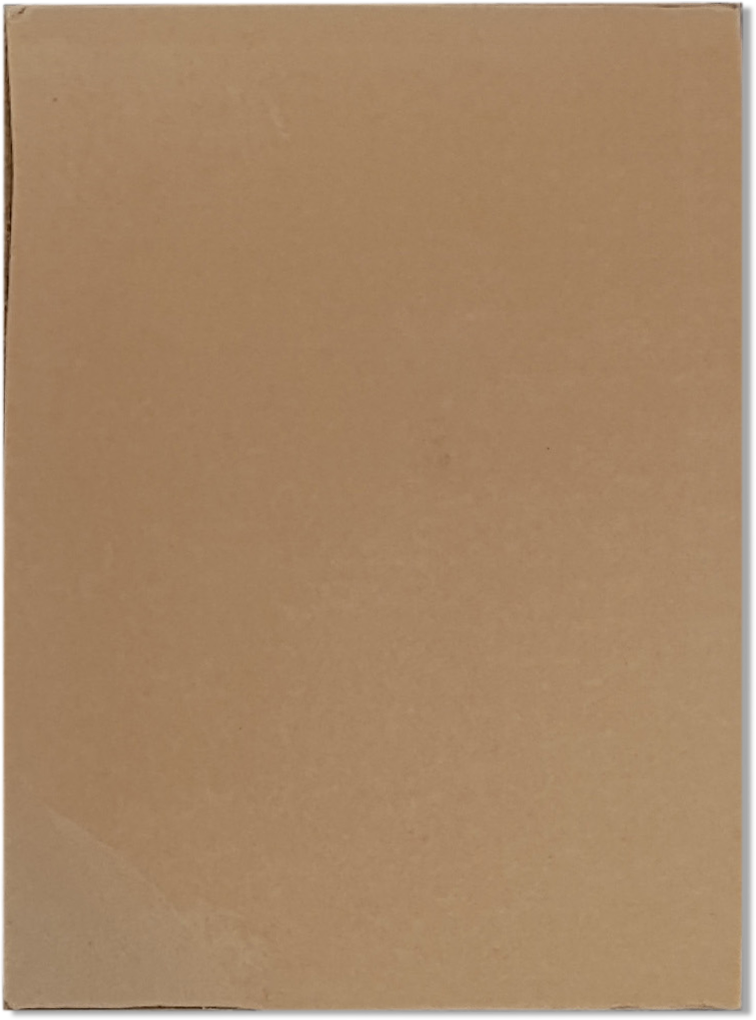}
        \caption{Cardboard}
        \label{Results-Materials-Cardboard}
    \end{subfigure}\\
    \begin{subfigure}[t]{0.49\columnwidth}
        \centering\includegraphics[width=0.4\textwidth]{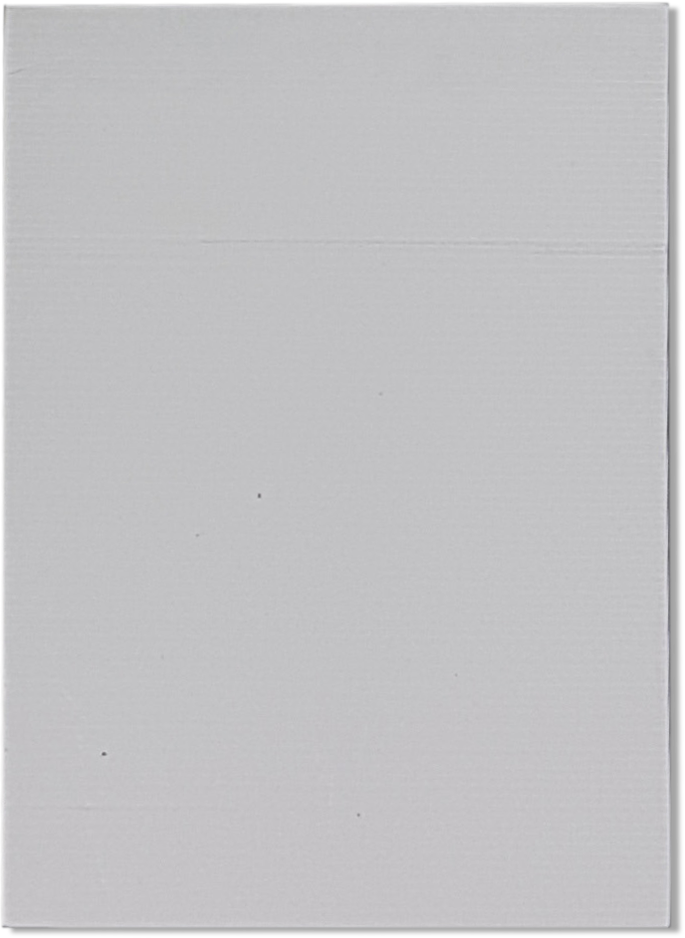}
        \caption{Corrugated plastic}
        \label{Results-Materials-CorrugatedPlastic}
    \end{subfigure}
    \begin{subfigure}[t]{0.49\columnwidth}
        \centering\includegraphics[width=0.4\textwidth]{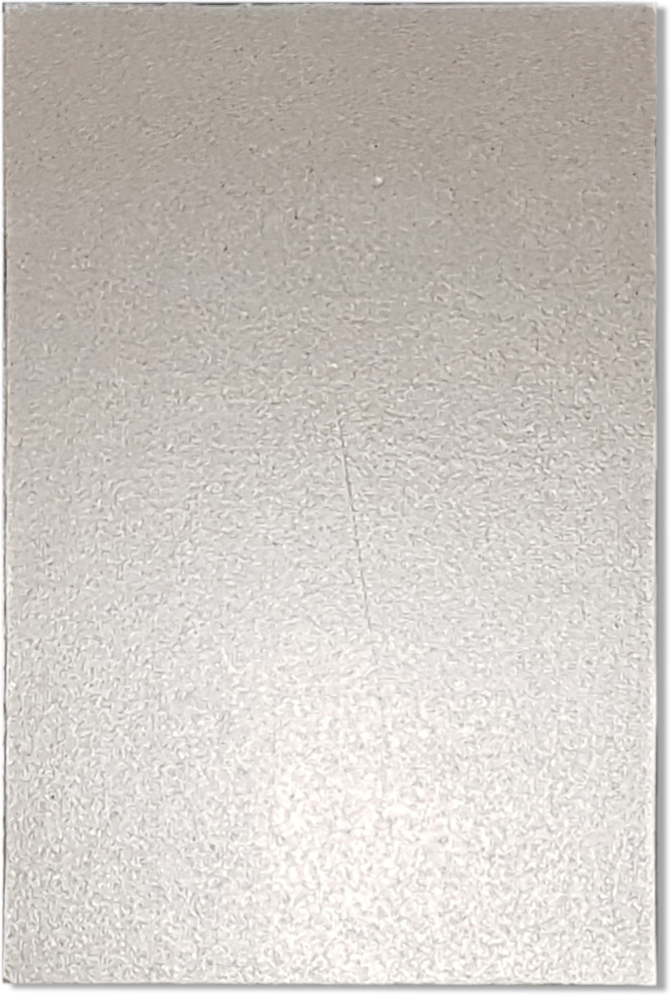}
        \caption{Mica sheet}
        \label{Results-Materials-Mica}
    \end{subfigure}
    \caption{Selected materials for model validation}
    \label{fig:Results-Materials}
\end{figure}

The bulk of experimental validation for the mechanistic modelling approach has been carried out for common metals using high-precision measurement equipment which is impractical to employ in a disassembly scenario. To be broadly applicable in these scenarios, the replicability of force measurements should be demonstrated on a real robot setup from onboard sensors over a range of different materials. For proof of principle we tried this using a KUKA LBR cobot equipped with joint torque sensors as shown in Figure \ref{fig:Results-Cutting-Example}. We selected experimental materials, shown in Figure \ref{fig:Results-Materials} to match the payload capabilities of this robot. However, our method should also be workable on larger industrial robots equipped with e.g. a 6-axis force torque sensor at the wrist. Cutting experiments were carried out at varying feed rates within a range selected for each material, keeping radial depth of cut (RDOC) and tool speed $\omega$ fixed. These materials possess dissimilar mechanical properties, varying degrees of structural homogeneity and thickness, ensuring differing force profiles are generated in each experiment. To correct for sensor bias in the collected force measurements, the observed end-effector forces were recorded prior to engagement with the material and the average added as a measurement offset during cutting. The experimental setup was replicated in the simulator and the average measured force fitted with the average model force using the Levenberg-Marquardt algorithm.

\begin{figure*}
	\centering
    \begin{subfigure}[t]{0.24\textwidth}
        \centering\includegraphics[width=\textwidth]{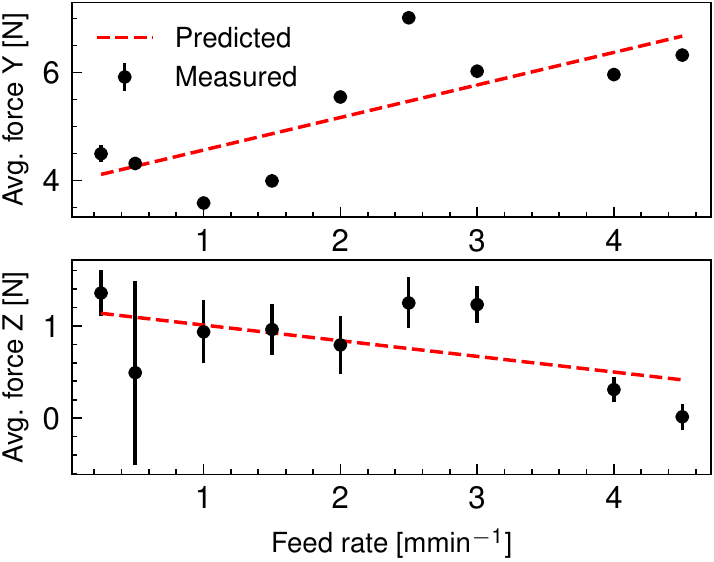}
        \caption{Foam, down milling RDOC 13.72mm}
        \label{fig:Results-Experimental-Foam-Fitted}
    \end{subfigure}
    \hfill
    \begin{subfigure}[t]{0.24\textwidth}
        \centering\includegraphics[width=\textwidth]{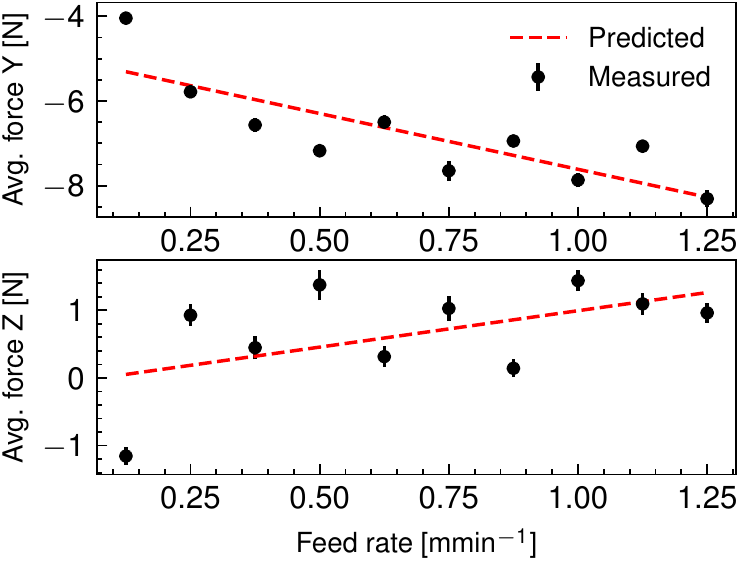}
        \caption{Cardboard, up milling, RDOC 6.34mm}
		\label{fig:Results-Experimental-Cardboard-Fitted}
    \end{subfigure}
    \hfill
    \begin{subfigure}[t]{0.24\textwidth}
        \centering\includegraphics[width=\textwidth]{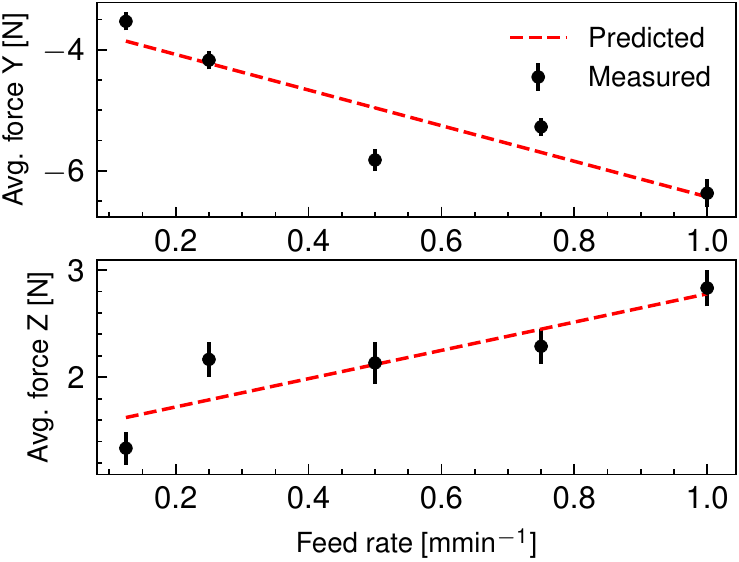}
        \caption{Corrugated plastic, up milling, RDOC 1.95mm}
		\label{fig:Results-Experimental-Plastic-Fitted}
    \end{subfigure}
    \hfill
    \begin{subfigure}[t]{0.24\textwidth}
        \centering\includegraphics[width=\textwidth]{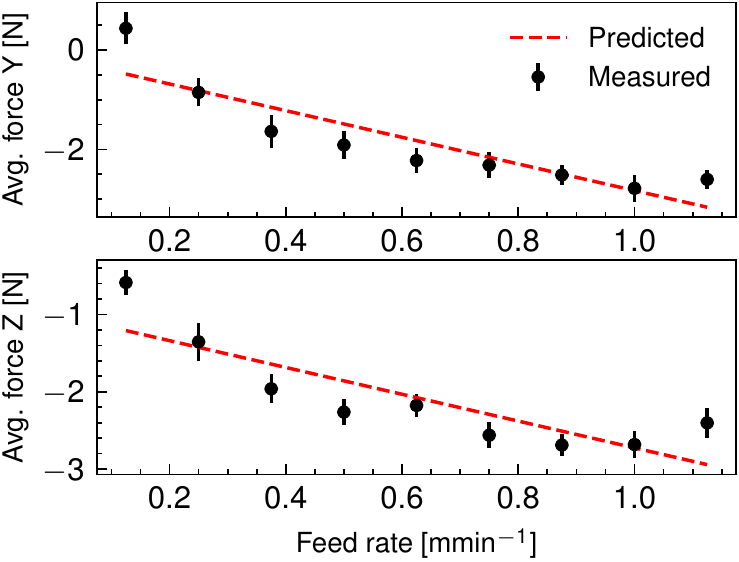}
        \caption{Mica, down milling, RDOC 1.00mm}
		\label{fig:Results-Experimental-Mica-Fitted}
    \end{subfigure}
	\caption{Average cutting forces measured from onboard sensors for selected materials taken with slitting saw tool with $N_{f}=100$, $\omega=500$rpm, $R=0.025$ at varying feed rates, overlaid with mechanistic model predicted average forces. RDOC refers to the experiment radial depth of cut. Forces transverse to the feed direction (+Y for up milling, -Y for down milling) are omitted.}
	\label{fig:Results-Experimental-Fitted}
\end{figure*}

Figure \ref{fig:Results-Experimental-Fitted} shows the average fitted model force for polyurethane (PU) foam, corrugated plastic and mica sheet respectively. For simplicity, the force components transverse to the feed direction ($\pm$Y) are neglected, as the influence of the milling force contribution from these directions is relatively marginal. For the latter three case studies of cardboard, plastic and mica, there is good correspondence of the average force between the model predictions and the measured forces with overall RMSE (root mean square error) of 0.634N, 0.700N, 0.396N respectively, despite their differing structure from common engineering materials, such as steels. For mica, the strongest relationship is observed, chiefly due to its greater degree of structural homogeneity and higher mechanical strength. The weakest relationship is observed for foam, which also exhibits behaviour in the feed direction contrary to expectation for a down-milling configuration. This is posited to be due to the high structural porosity and low mechanical strength, contributing to low observed cutting force, apparent in the surface normal direction (+Z), in combination with viscous friction effects in opposition to the feed direction. Note, however, that approximate modelling of the interaction forces is still possible without modifications to the model even in spite of this condition, with an overall RMSE of 0.574N, albeit suffering from a higher RMSE of 0.727N parallel to the feed. In practice, for instantaneous force modelling, an RMSE of $\sim$1N is observed even for higher accuracy modelling approaches based on machine learning \cite{CuttingModelMachineLearning}. While the relative error is much lower, due to the much higher forces involved, such models would need to be adapted to the range of materials considered.
Chiefly, it should be mentioned the goal of the proposed modelling and simulation approach is not necessarily accurate reproduction of the instantaneous forces, but rather a capability to replicate approximately the distribution of observed forces apparent at the end-effector, as these will have the greatest impact on the sampled state space and overall reward -- and hence the policy -- during the training process.

\subsection{Comparison of OSC with ET-based OSC}
To compare ET-OSC with traditional OSC in the case of a variable stiffness policy, we establish a case study for a cutting task over a planar surface with fixed, random material properties, using a variable stiffness policy trained using the procedure described in section \ref{sec:Method-Training-Env}. To demonstrate the applicability of ET-OSC even for pre-trained policies, the policy was trained with only traditional OSC at a critically damped configuration. In the first case, the trained policy is deployed directly using OSC without modification, while in the second case the policy is deployed using ET-OSC. To evaluate the performance of ET-OSC, 10 repeat evaluations are performed, reducing the damping ratio of the controller from 1.0 to 0.1. Note the evaluation policy is trained only for a damping ratio of 1.0, however, in all expected scenarios, completion of the task is expected without violating safety constraints imposed upon the manipulator, such as joint limits.

\begin{figure*}
	\centering
    \begin{subfigure}[t]{0.32\textwidth}
        \centering\includegraphics[width=\textwidth]{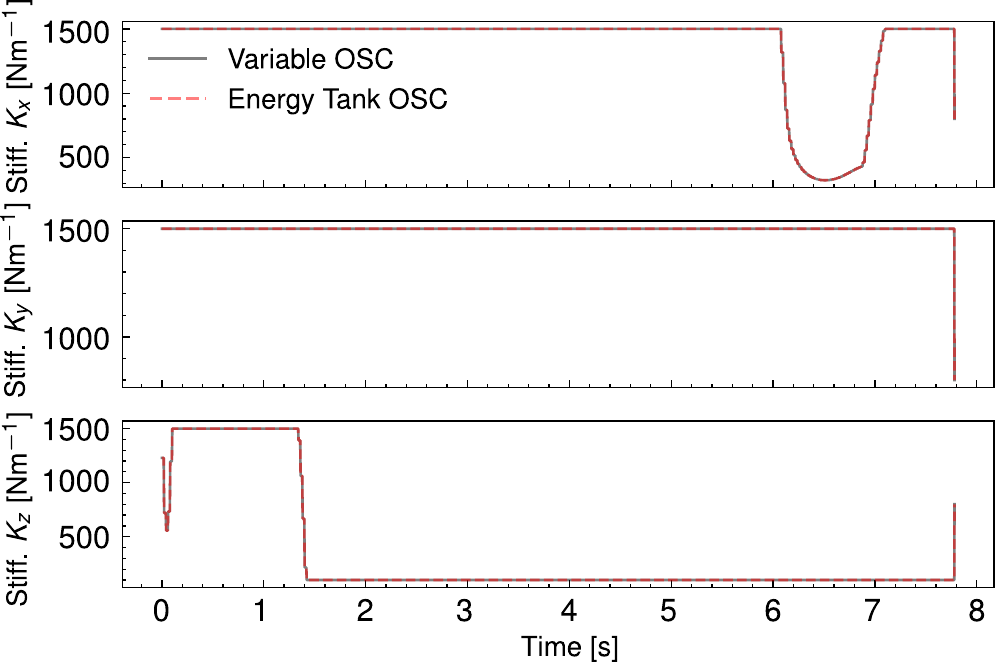}
        \caption{Damping ratio 1.0}
        \label{fig:Results-Damping-Ratio-Action-10}
    \end{subfigure}
    \begin{subfigure}[t]{0.32\textwidth}
        \centering\includegraphics[width=\textwidth]{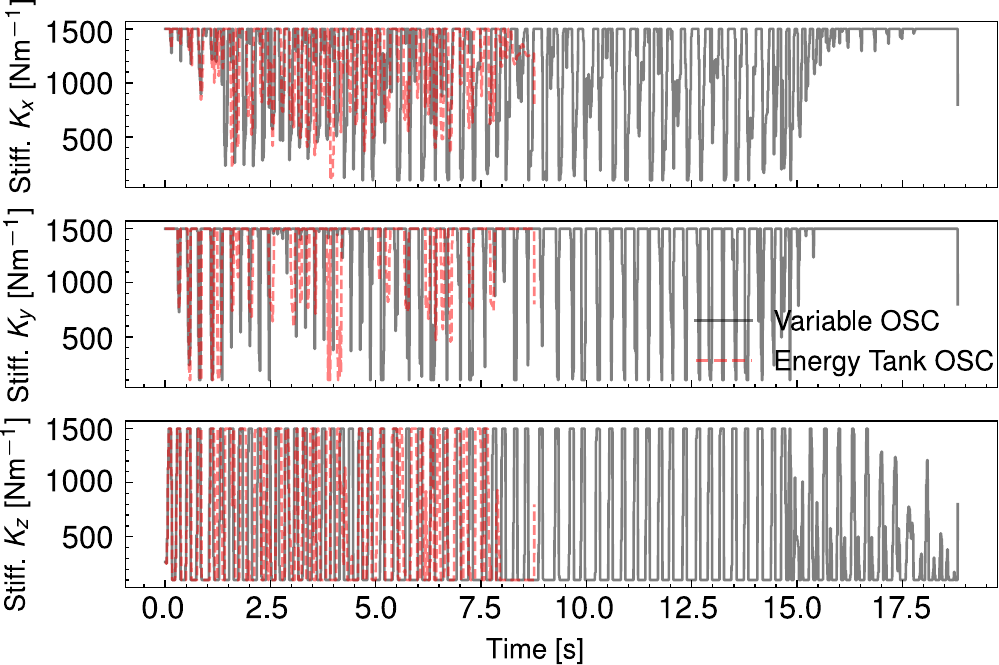}
        \caption{Damping ratio 0.1}
        \label{fig:Results-Damping-Ratio-Action-01}
    \end{subfigure}\\
        \begin{subfigure}[t]{0.32\textwidth}
        \centering\includegraphics[width=\textwidth]{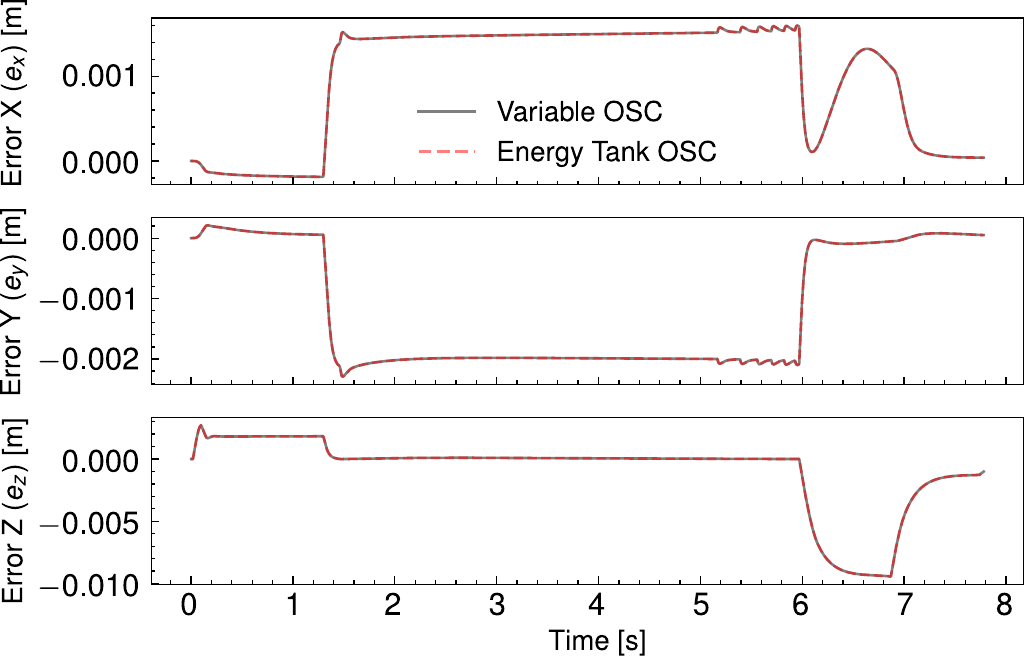}
        \caption{Damping ratio 1.0}
        \label{fig:Results-Damping-Ratio-Comparison-10}
    \end{subfigure}
    \begin{subfigure}[t]{0.32\textwidth}
        \centering\includegraphics[width=\textwidth]{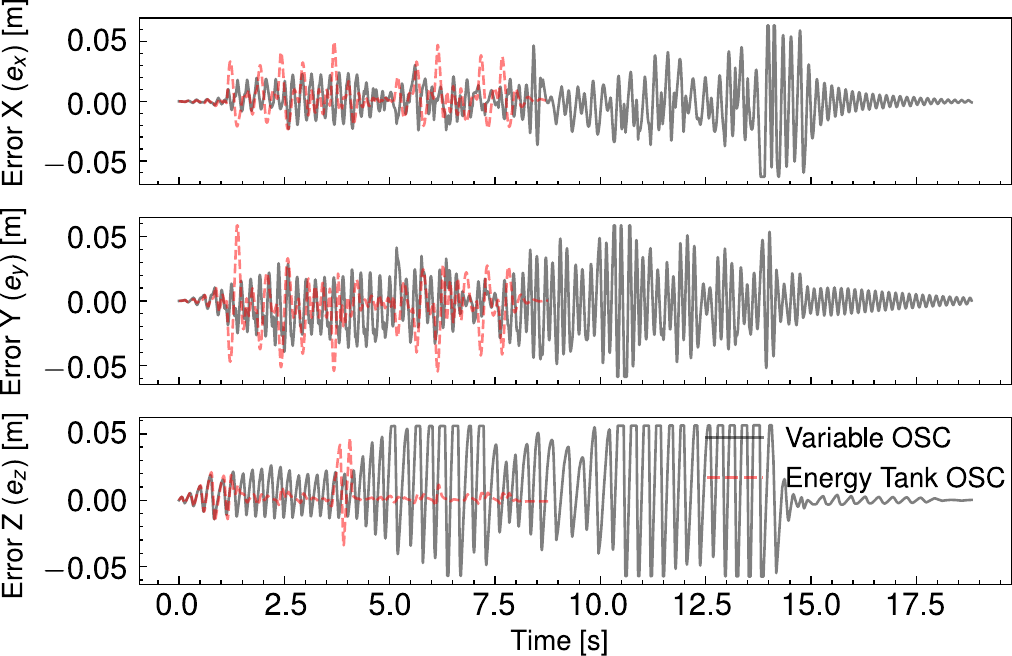}
        \caption{Damping ratio 0.1}
        \label{fig:Results-Damping-Ratio-Comparison-01}
    \end{subfigure}
	\caption{Comparison of stiffness of energy-tank-based OSC with traditional variable OSC with different settings of controller damping ratio (relative to critically damped configuration). With a gain-based policy action space, it is possible for strategies to be implemented that result in rapid variation of the controller stiffness over time. With insufficient damping, instability results in divergence of the traditional OSC and early termination due to violation of safety constraints.}
	\label{fig:Results-Damping-Ratio}
\end{figure*}

An overview of the agent variable stiffness policy outputs is shown in Figure \ref{fig:Results-Damping-Ratio-Action-10}, \ref{fig:Results-Damping-Ratio-Action-01}, which demonstrates the standard stiffness variation profile for the critically damped configuration and aggressive variations in the stiffness for the lowest damping ratios. Along the direction of cut, in the positive x direction, the agent adopts a consistently high stiffness. Figure \ref{fig:Results-Damping-Ratio-Comparison-10}, \ref{fig:Results-Damping-Ratio-Comparison-01} shows the deviation of the policy from the setpoint position, indicating the tracking performance of ET-OSC and OSC. Notable by comparing Figure \ref{fig:Results-Damping-Ratio-Action-10}--\ref{fig:Results-Damping-Ratio-Comparison-10} and Figure \ref{fig:Results-Damping-Ratio-Action-01}--\ref{fig:Results-Damping-Ratio-Comparison-01}, the policy overcompensates for the path error with the reduced damping and implements undesirable behaviour which adds energy to the system. In the case of damping ratio of 1.0, the performance of the ET-OSC and OSC are broadly similar, indicating the performance of the ET-OSC in the case dissipation is adequate to fill the tank. However, with further reduced damping ratio (Figure \ref{fig:Results-Damping-Ratio-Action-01}), the effect of stiffness variation of the policy becomes significant as the damping is insufficient to guarantee passivity, and the traditional OSC policy diverges. This effect is most pronounced in the Z direction, where the saturation of the error signal indicates the violation of safety constraints imposed on the workspace and joint limits. Furthermore, OSC fails to converge to the desired path throughout the task, which is remediated only by stabilisation of the commanded stiffness signal after $\sim$15 s, as shown in Figure \ref{fig:Results-Damping-Ratio-Action-01}. Note in both cases, the oscillation and reduced task performance of the controller is unavoidable as the system is highly under-damped. However, the ET-based OSC is capable of completing the cutting task without divergence or violation of joint safety constraints in spite of this condition. This furthermore demonstrates the ability to re-use policies even with aggressive variable stiffness characteristics without modification with the proposed ET-based OSC.

\subsection{Agent Evaluation}
\begin{figure}
	\centering
    \begin{subfigure}[t]{0.45\columnwidth}
        \centering\includegraphics[width=\textwidth]{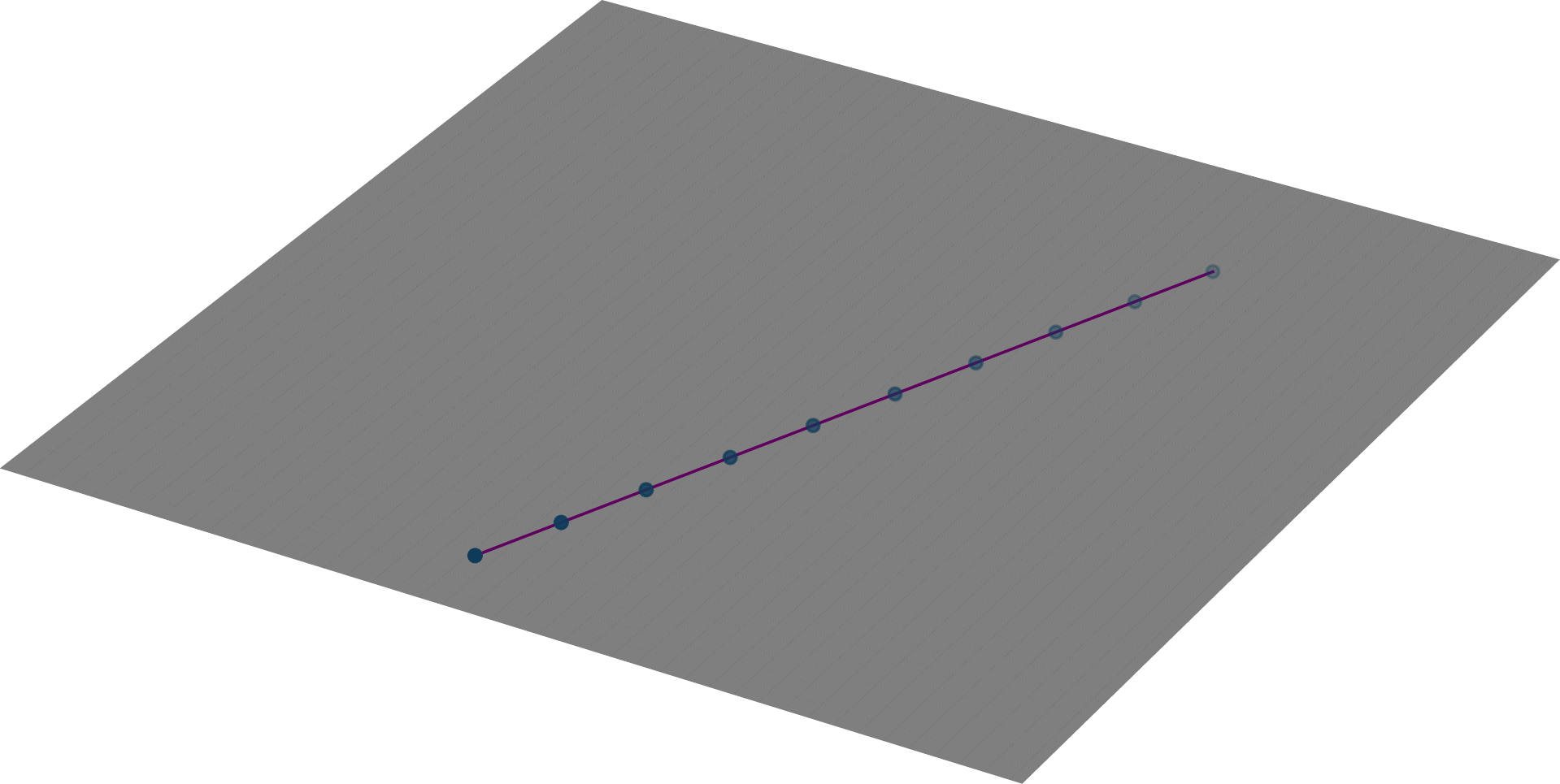}
        \caption{Flat surface}
        \label{fig:Results-Case-Studies-1}
    \end{subfigure}
    \begin{subfigure}[t]{0.45\columnwidth}
        \centering\includegraphics[width=\textwidth]{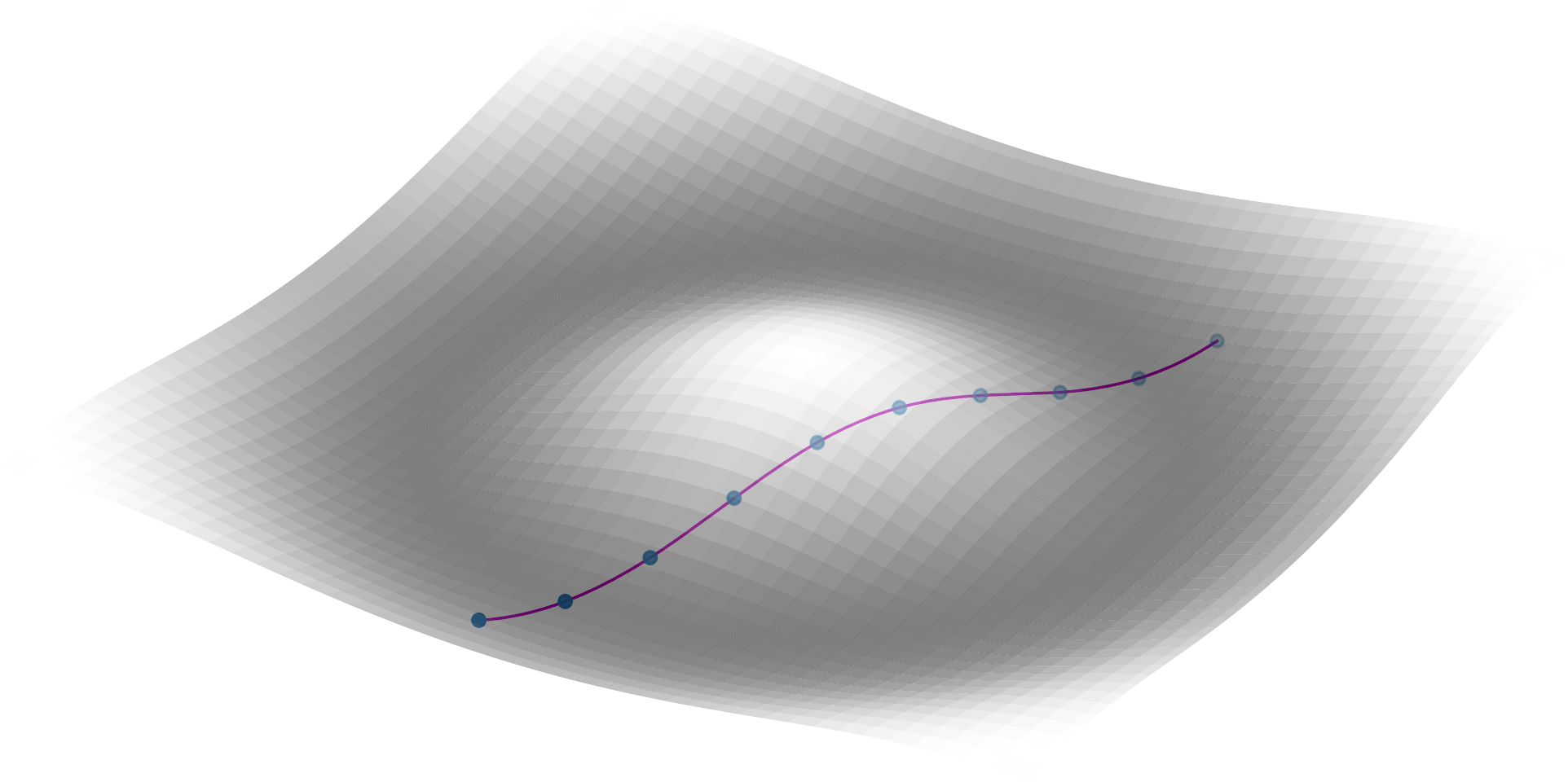}
        \caption{Sinusoidal}
        \label{fig:Results-Case-Studies-2}
    \end{subfigure}\\\smallskip
    \begin{subfigure}[t]{0.45\columnwidth}
        \centering\includegraphics[width=\textwidth]{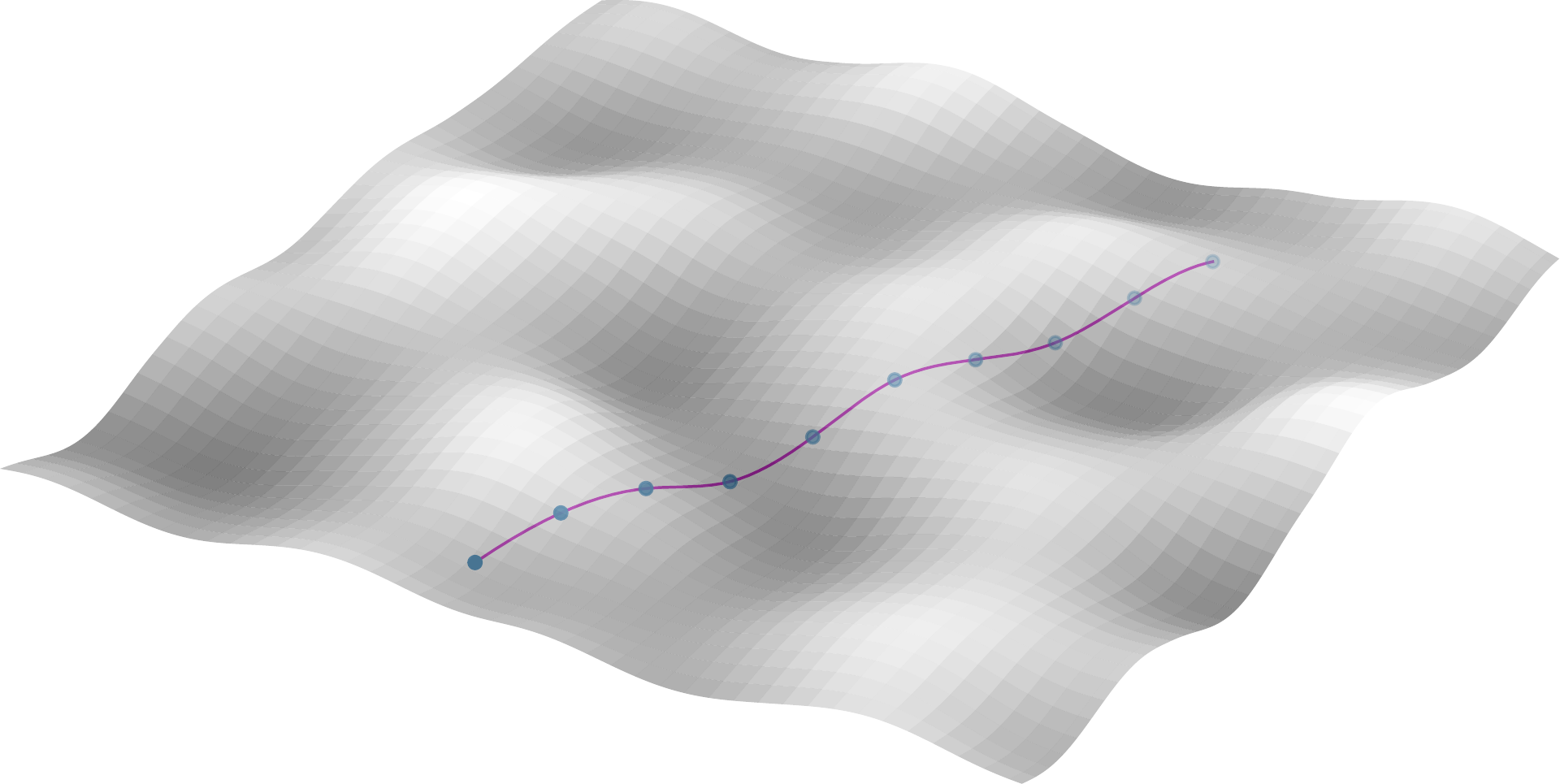}
        \caption{Perlin noise}
        \label{fig:Results-Case-Studies-3}
    \end{subfigure}
    \begin{subfigure}[t]{0.45\columnwidth}
        \centering\includegraphics[width=\textwidth]{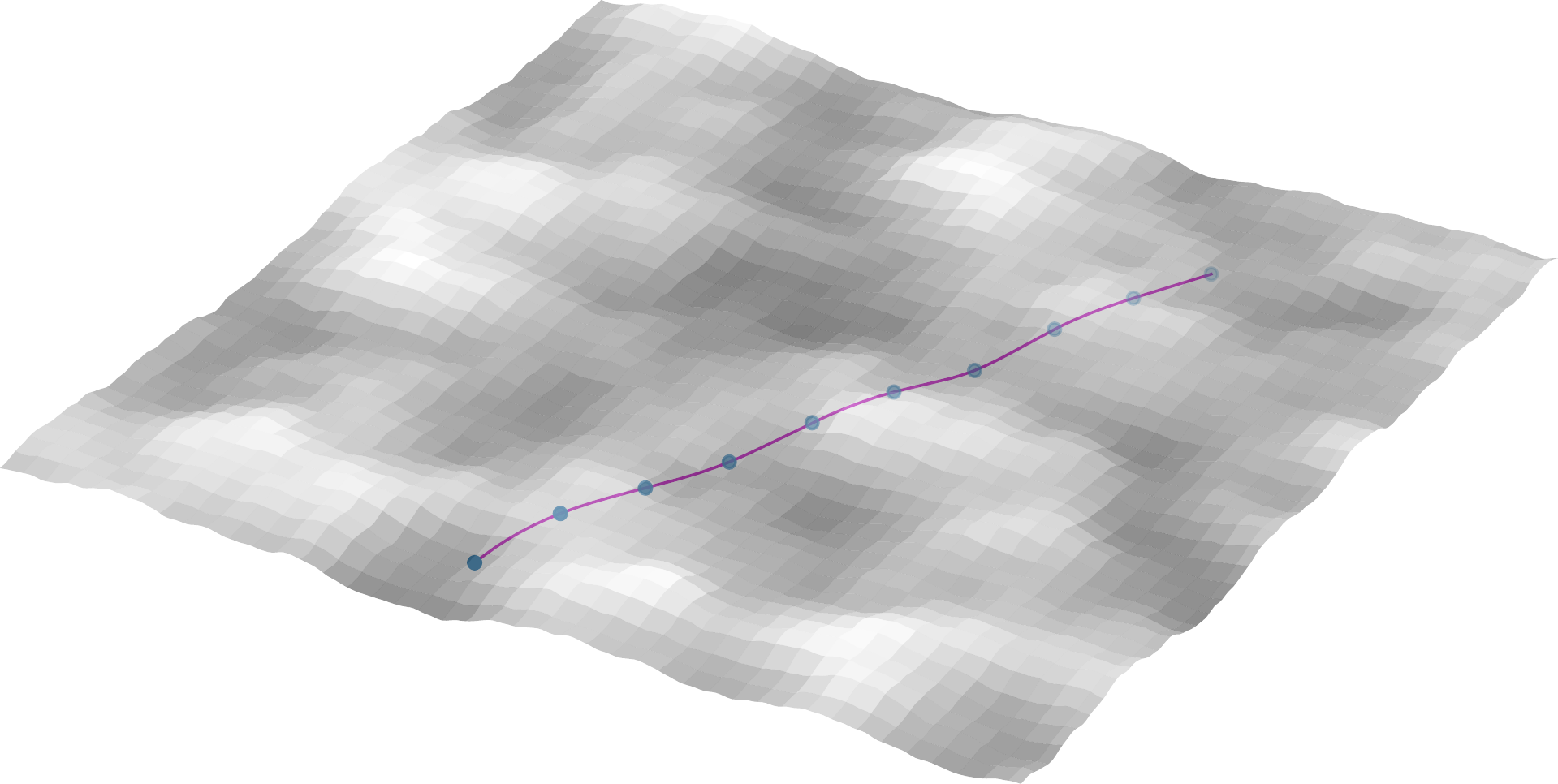}
        \caption{Fractal noise}
        \label{fig:Results-Case-Studies-4}
    \end{subfigure}
	\caption{Surface geometry case studies considered in simulation: planar material \ref{fig:Results-Case-Studies-1}, low-curvature surface \ref{fig:Results-Case-Studies-2}, high curvature or deformed material \ref{fig:Results-Case-Studies-3}, and rough, textured surface with high local surface curvature \ref{fig:Results-Case-Studies-4}. Different workpiece geometries influence the tool-workpiece engagement profile, which affects the selection of relevant process parameters (e.g. depth of cut).}
	\label{fig:Results-Case-Studies}
\end{figure}

We train a variable OSC policy based on the procedure in section \ref{sec:Method-Training-Env} and evaluate using 4 random case study environments shown in Figure \ref{fig:Results-Case-Studies}. The case studies encapsulate differing levels of local surface curvature and introduce random variation from task to task, which requires the learned policy to adapt to differing tool-workpiece engagement (TWE) and material properties, reflective of a previously unseen component. The material properties were sampled from a random uniform distribution reflecting a down-milling (climb) configuration over the range of evaluated case study materials. To evaluate the effectiveness of the proposed method, we compare over 20 trials with an efficient global optimisation (EGO) strategy as proposed in \cite{OptimisationMillingAutomatic}. This comparison accounts for the case where prior knowledge, such as CAD and material models are both known. We adopt similar conditions to those applied in \cite{OptimisationMillingAutomatic}, using a kriging model with constant mean and Gaussian process (GP) estimator with radial basis function (RBF) kernel to model the process cost, sampling the optimisation space over 115 rollouts, and the optimal set of process parameters estimated from the maxima of the GP reward surface model. A favourable comparison with the EGO approach suggests the capability of the learned policy to select process parameters online without prior knowledge of the task. As a benchmark, we compare with a basic ``baseline'' policy which selects constant process parameters as $||\vecframe{v}{W}||=1.5$mmin\textsuperscript{-1}, depth of cut DOC$=$5mm, $\mat{K}_{p}=800\mat{I_{3}}$, which reflects a conservative initial attempt for an unseen component under the trial-and-error approach. Finally, we investigate the effect of adding a radial depth of cut (RDOC) offset to the policy outputs. This offset is identical to the benchmark case, which explores the capability of the policy to be guided by operator input and robustness to this scenario by adjusting the remaining process parameters if the user selection of RDOC is inappropriate for the task.
\begin{figure}
	\centering
	\includegraphics[width=0.95\columnwidth]{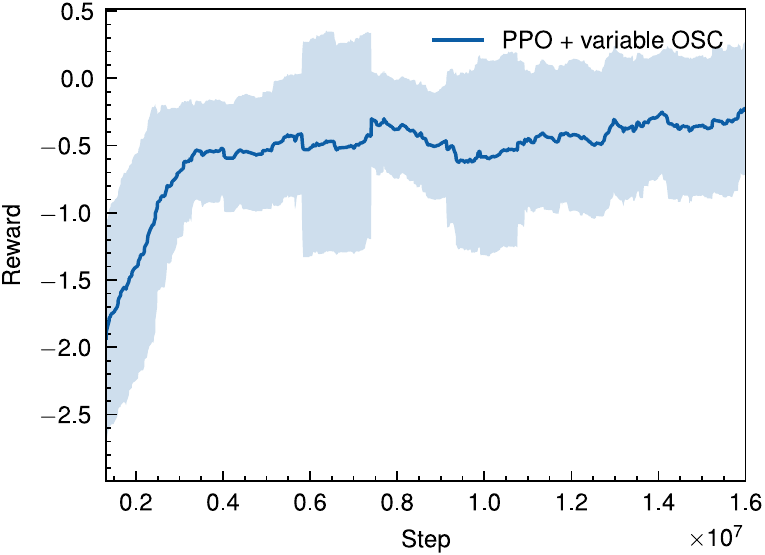}
	\caption{Training curve for cutting task with Proximal Policy Optimisation (PPO) algorithm and variable operational space control with domain randomisation of the workpiece geometry, tool path and material properties.}
	\label{fig:Results-Training-Curves}
\end{figure}
\begin{figure}
	\centering
	\includegraphics[width=0.95\columnwidth]{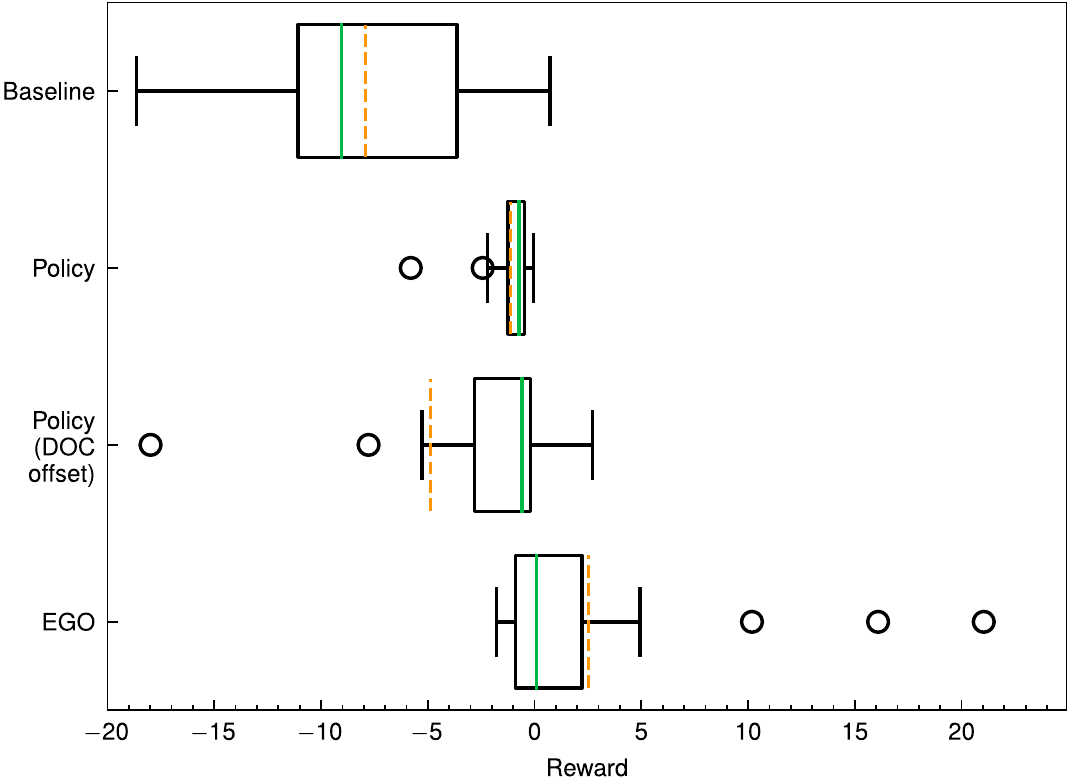}
	\caption{Box plot showing distribution of rewards from 20 simulated cutting experiments between fixed process parameter ``baseline'' policy, trained variable operational space control policy, and offline optimisation (EGO). One outlier is present at -56.53 for the policy with depth of cut (DOC) offset, which is omitted for clarity.}
	\label{fig:Results-Reward-Distribution-Comparison}
\end{figure}
\begin{table*}
	\caption{Comparison of average reward function components over 20 simulated cutting trials between the trained variable OSC policy, fixed process parameter ``baseline'' policy, and efficient global optimisation (EGO) approach, with reward breakdown of four sample trials in Figures \ref{fig:Results-Reward-Contour-Expt-1}--\ref{fig:Results-Reward-Contour-Expt-4}.\label{tab:Results-Reward-Comparison}}
    \centering
	\begin{tabular}{|l|l|>{\raggedright}p{0.1\textwidth}|>{\raggedright}p{0.1\textwidth}|>{\raggedright}p{0.1\textwidth}|>{\raggedright}p{0.1\textwidth}|>{\raggedright}p{0.1\textwidth}|}
		\hline 
		Strategy & Expt & Time & Deviation & MRV & Force & Total\tabularnewline
		\hline
		{Baseline} & {1} & {-0.6752} & {-0.04292} & {0.1487} & {-1.639} & {-2.209}\tabularnewline
		& {2} & {-1.555} & {-2.334} & {2.056} & {-12.69} & {-14.53}\tabularnewline
		& {3} & {-1.302} & {-3.387} & {7.889} & {-13.82} & {-10.62}\tabularnewline
		& {4} & {-0.9532} & {-0.3946} & {3.764} & {-1.701} & {0.7154}\tabularnewline
		\cline{2-7}
		& \emph{Avg.} & \emph{-0.9662}\par\emph{$\pm$0.06015} & \emph{-1.561}\par\emph{$\pm$0.2477} & \emph{2.426}\par\emph{$\pm$0.4446} & \emph{-7.827}\par\emph{$\pm$1.012} & \emph{-7.928}\par\emph{$\pm$1.173}\tabularnewline
		\hline
		{EGO\cite{OptimisationMillingAutomatic}} & {1} & {-0.3562} & {-0.02075} & {0.1259} & {-0.2042} & {-0.4553}\tabularnewline
		& {2} & {-0.7818} & {-1.773} & {16.31} & {-11.39} & {2.365}\tabularnewline
		& {3} & {-0.6538} & {-0.7885} & {23.65} & {-6.121} & {16.09}\tabularnewline
		& {4} & {-0.4922} & {-0.3531} & {7.554} & {-1.773} & {4.935}\tabularnewline
		\cline{2-7}
		& \emph{Avg.} & \emph{-0.4947}\par\emph{$\pm$0.02975} & \emph{-0.4525}\par\emph{$\pm$0.1054} & \emph{6.832}\par\emph{$\pm$1.936} & \emph{-3.354}\par\emph{$\pm$0.7925} & \emph{2.530}\par\emph{$\pm$1.383}\tabularnewline
		\hline
		{Ours} & {1} & {-0.4550} & {-0.03773} & {0.4675} & {-0.3418} & {-0.3670}\tabularnewline
		& {2} & {-0.9090} & {-0.03431} & {0.05685} & {-0.05936} & {-0.9458}\tabularnewline
		& {3} & {-0.8088} & {-0.2255} & {2.669} & {-3.248} & {-1.614}\tabularnewline
		& {4} & {-0.5976} & {-0.06534} & {0.6324} & {-0.2541} & {-0.2846}\tabularnewline
		\cline{2-7}
		& \emph{Avg.} & \emph{-0.6135}\par\emph{$\pm$0.02923} & \emph{-0.09865}\par\emph{$\pm$0.01781} & \emph{0.5349}\par\emph{$\pm$0.1419} & \emph{-0.9572}\par\emph{$\pm$0.3273} & \emph{-1.135}\par\emph{$\pm$0.2825}\tabularnewline
		\hline
		{Ours (DOC offset)} & {1} & {-0.4428} & {-0.03700} & {1.315} & {-1.135} & {-0.2996}\tabularnewline
		& {2} & {-0.8986} & {-0.1114} & {1.017} & {-0.5519} & {-0.5445}\tabularnewline
		& {3} & {-1.050} & {-1.137} & {8.822} & {-14.41} & {-7.774}\tabularnewline
		& {4} & {-0.5752} & {-0.1399} & {1.710} & {-0.6540} & {0.3405}\tabularnewline
		\cline{2-7}
		& \emph{Avg.} & \emph{-0.6516}\par\emph{$\pm$0.03687} & \emph{-0.7065}\par\emph{$\pm$0.4077} & \emph{1.814}\par\emph{$\pm$0.4388} & \emph{-5.336}\par\emph{$\pm$2.508} & \emph{-4.880}\par\emph{$\pm$2.886}\tabularnewline
		\hline
	\end{tabular}
\end{table*}

The evolution of the reward function over training is shown in Figure \ref{fig:Results-Training-Curves}, showing rapid improvement up to $4\times10^{6}$ samples, before convergence to maximum reward, demonstrating successful learning of the process parameter selection strategy. To identify the overall level of performance in the context of other approaches, the distribution of the rewards obtained between the range of tasks is shown in Figure \ref{fig:Results-Reward-Distribution-Comparison}, and average rewards over the presented case studies in Table \ref{tab:Results-Reward-Comparison}. The overall level of performance between each strategy demonstrates the agent performs to a similar level as an offline optimisation strategy, highlighting the effectiveness of the proposed method. However, as the optimisation strategy has access to the full material and CAD models beforehand, the expected performance of the EGO approach is higher overall. The policy performance is markedly more consistent than the baseline for the majority of trials, even in the case that a DOC offset bias is added by the user, implying the limitations of a trial-and-error parameter selection strategy. Comparing the individual reward function components suggests reductions in path tracking error by 54\% relative to the benchmark, even for the DOC adjusted policy, while process time is maintained within 25\% of the optimum obtained with EGO, rising to 31\% for the DOC offset case. 

\begin{figure*}
	\centering
    \begin{subfigure}[t]{0.95\columnwidth}
        \centering\includegraphics[width=\textwidth]{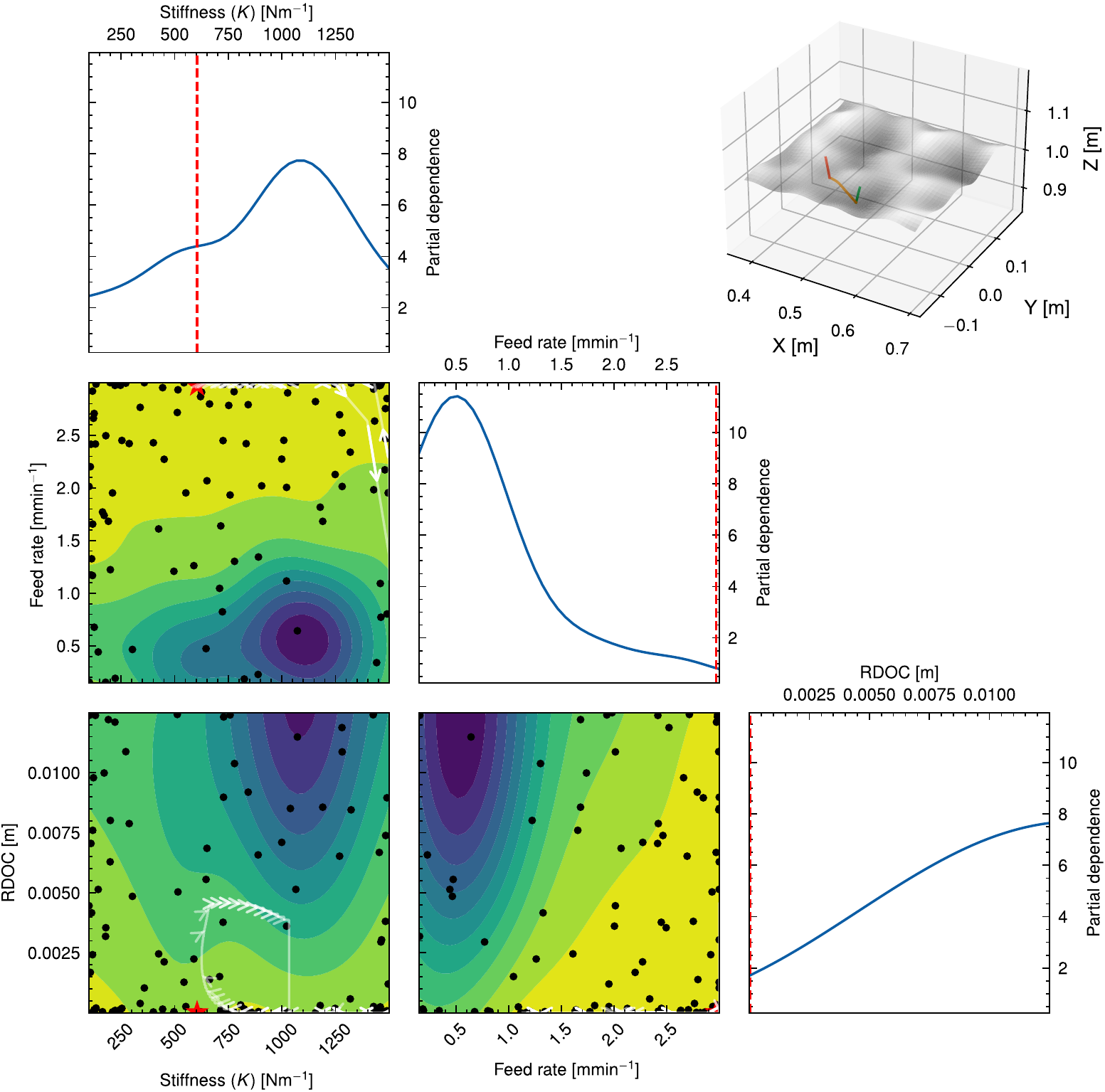}
        \caption{Case study 1 $\vec{K}_{c}=\begin{bmatrix}718.7 & 839.9 & 0.03656\end{bmatrix}^{\tpose}$Nmm$^{-2}$, $\vec{K}_{e}=\begin{bmatrix}8.337 & 0.4894 & -0.009854\end{bmatrix}^{\tpose}$ Nmm$^{-1}$}
        \label{fig:Results-Reward-Contour-Expt-1}
    \end{subfigure}
    \begin{subfigure}[t]{0.95\columnwidth}
        \centering\includegraphics[width=\textwidth]{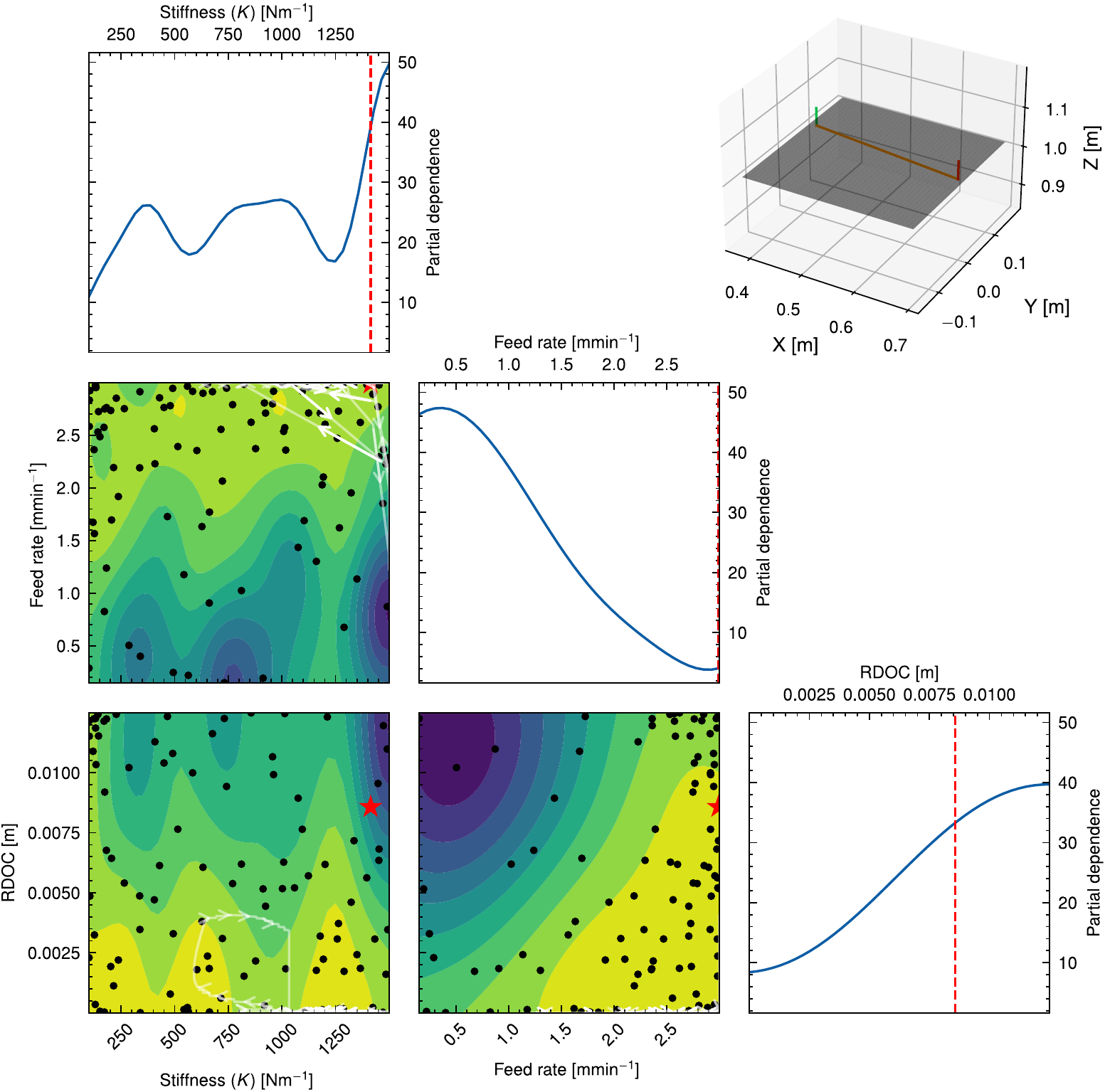}
        \caption{Case study 2 $\vec{K}_{c}=\begin{bmatrix}368.4 & 759.6 & 0.03994\end{bmatrix}^{\tpose}$Nmm$^{-2}$, $\vec{K}_{e}=\begin{bmatrix}3.306 & 3.509 & -0.007470\end{bmatrix}^{\tpose}$ Nmm$^{-1}$}
        \label{fig:Results-Reward-Contour-Expt-2}
    \end{subfigure}
    \begin{subfigure}[t]{0.95\columnwidth}
        \centering\includegraphics[width=\textwidth]{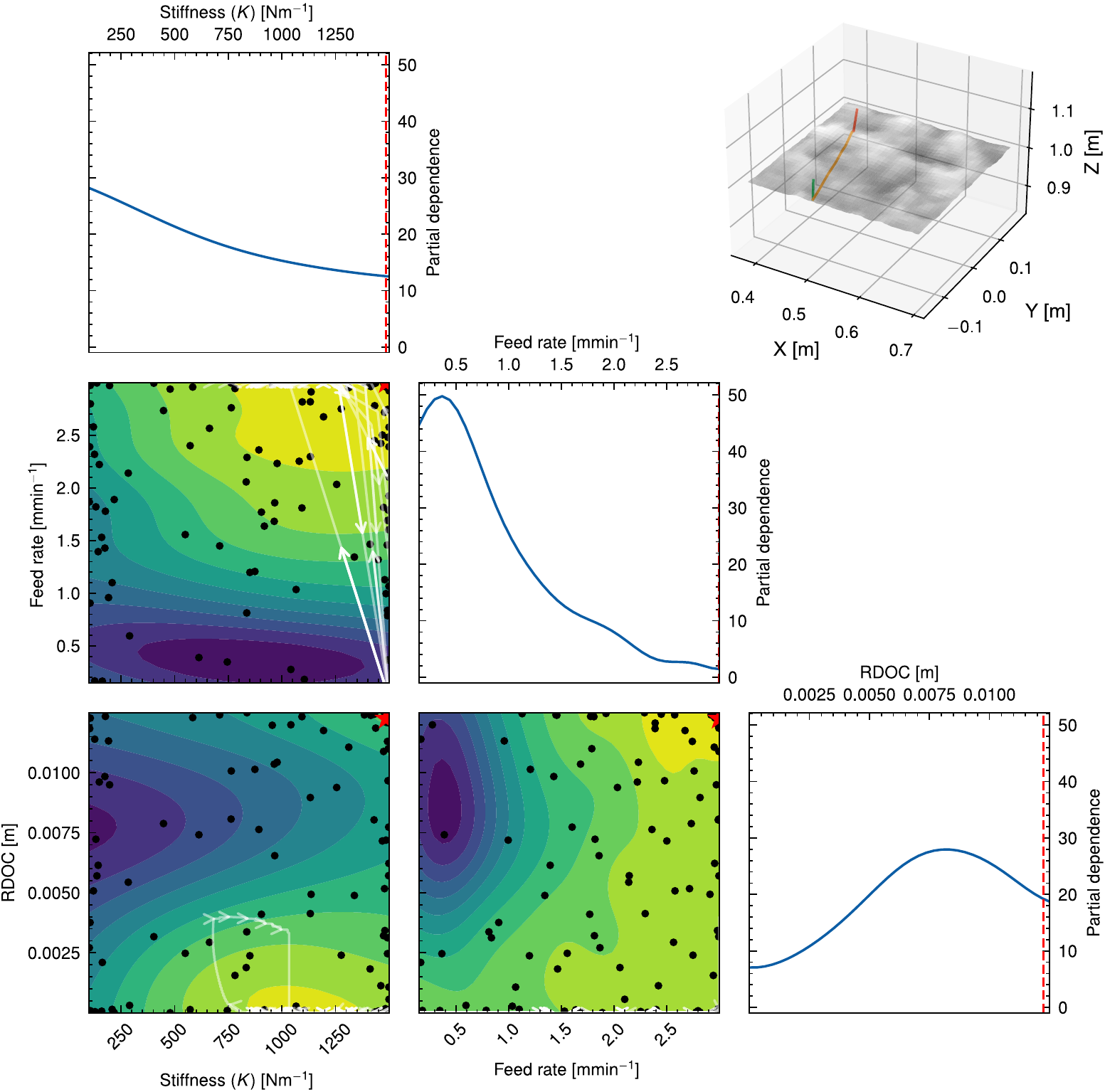}
        \caption{Case study 3 $\vec{K}_{c}=\begin{bmatrix}343.7 & 788.0 & -0.04609\end{bmatrix}^{\tpose}$Nmm$^{-2}$, $\vec{K}_{e}=\begin{bmatrix}9.203 & 3.984 & -0.0005049\end{bmatrix}^{\tpose}$ Nmm$^{-1}$}
        \label{fig:Results-Reward-Contour-Expt-3}
    \end{subfigure}
    \begin{subfigure}[t]{0.95\columnwidth}
        \centering\includegraphics[width=\textwidth]{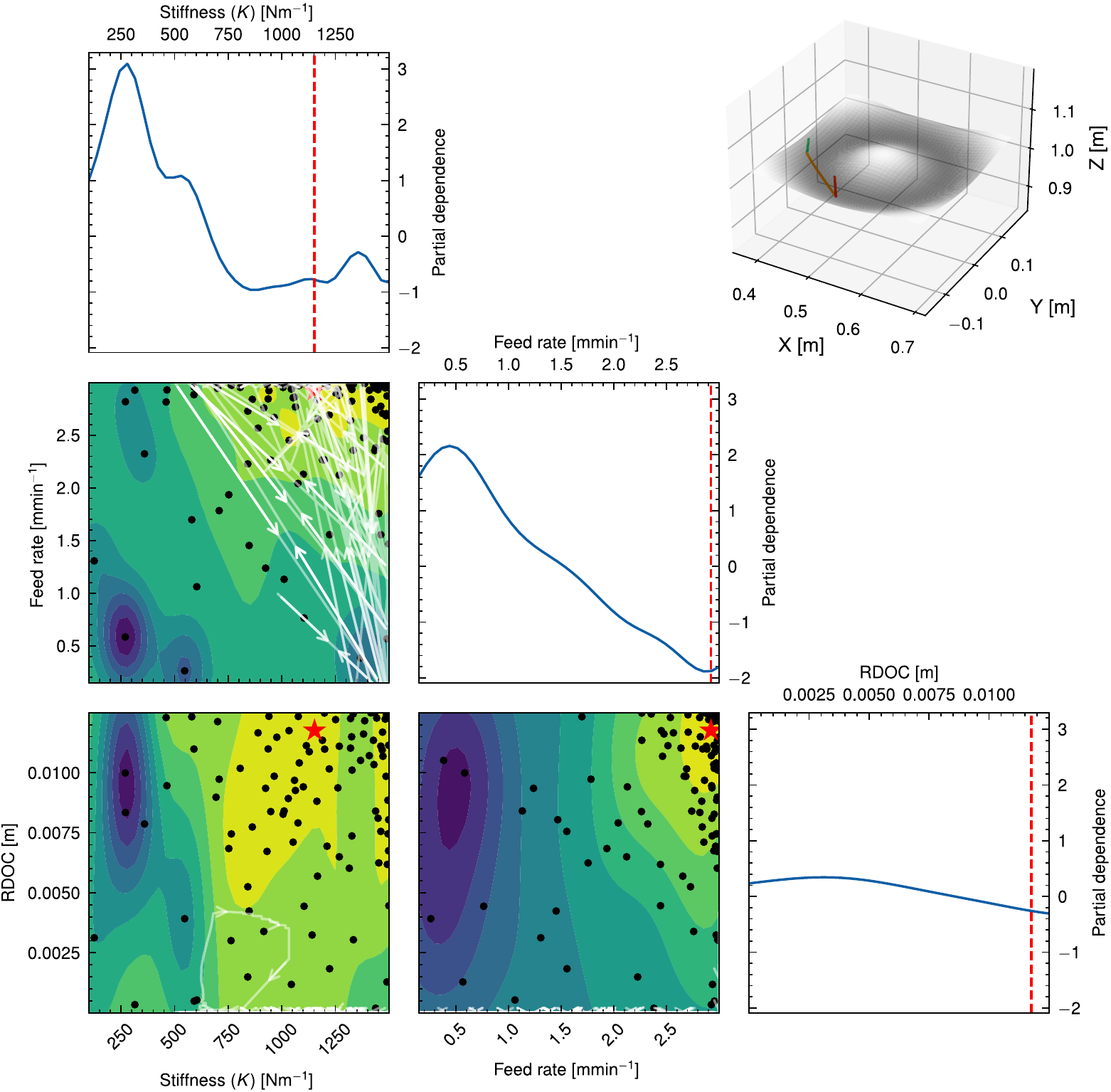}
        \caption{Case study 4 $\vec{K}_{c}=\begin{bmatrix}463.4 & 997.7 & 0.09269\end{bmatrix}^{\tpose}$Nmm$^{-2}$, $\vec{K}_{e}=\begin{bmatrix}3.253 & 6.923 & 0.0002610\end{bmatrix}^{\tpose}$ Nmm$^{-1}$}
        \label{fig:Results-Reward-Contour-Expt-4}
    \end{subfigure}
	\caption{Contour plots of estimated reward function dependence of each process parameter pair for each cutting case study using Gaussian process (GP) kriging model. The star shows the location of the found optimum. The diagonal plots show the partial dependence of the reward function with respect to each process parameter; the dotted line marks the found optimum. Top right: Plot of surface geometry and planned cutting path.}
	\label{fig:Results-Reward-Contour-Expts}
\end{figure*}
\begin{figure}
	\centering
    \begin{subfigure}[t]{0.49\columnwidth}
        \centering\includegraphics[width=\textwidth]{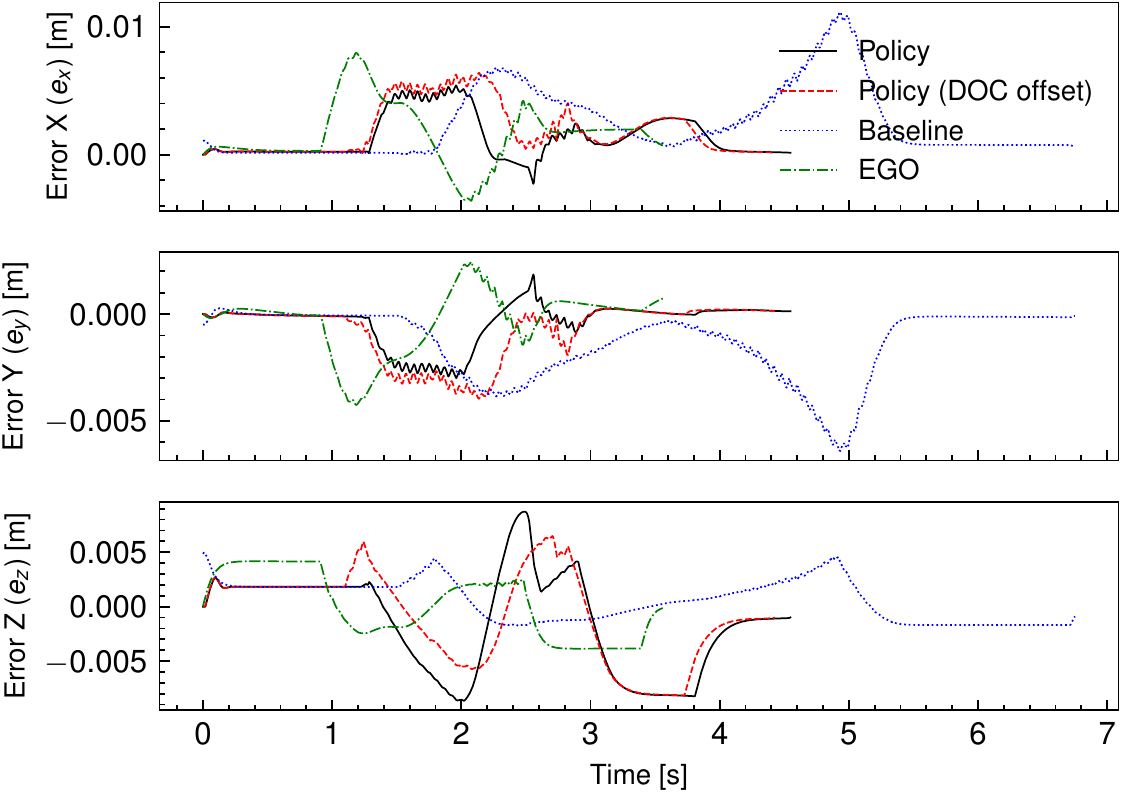}
        \caption{Path error, all}
        \label{fig:Results-Obs-Expt-1-Error}
    \end{subfigure}
    \begin{subfigure}[t]{0.49\columnwidth}
        \centering\includegraphics[width=\textwidth]{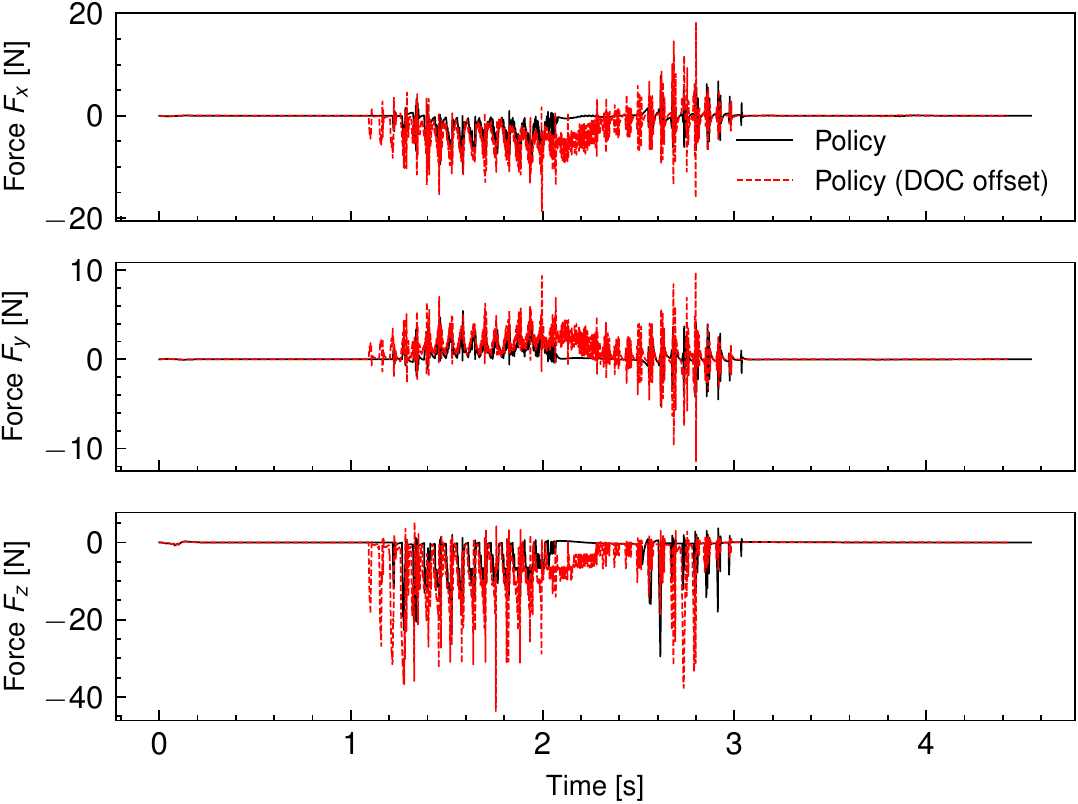}
        \caption{Force, policies}
        \label{fig:Results-Obs-Expt-1-Force-Policies}
    \end{subfigure}
    \begin{subfigure}[t]{0.49\columnwidth}
        \centering\includegraphics[width=\textwidth]{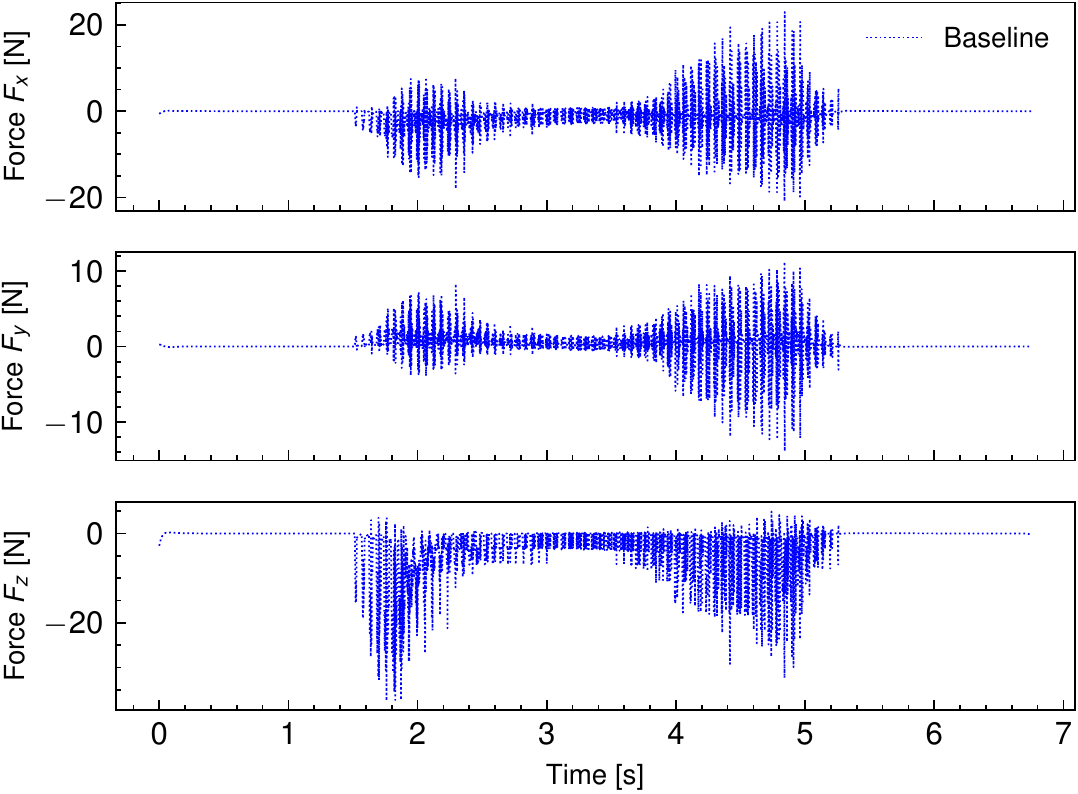}
        \caption{Force, baseline}
        \label{fig:Results-Obs-Expt-1-Force-Baseline}
    \end{subfigure}
    \begin{subfigure}[t]{0.49\columnwidth}
        \centering\includegraphics[width=\textwidth]{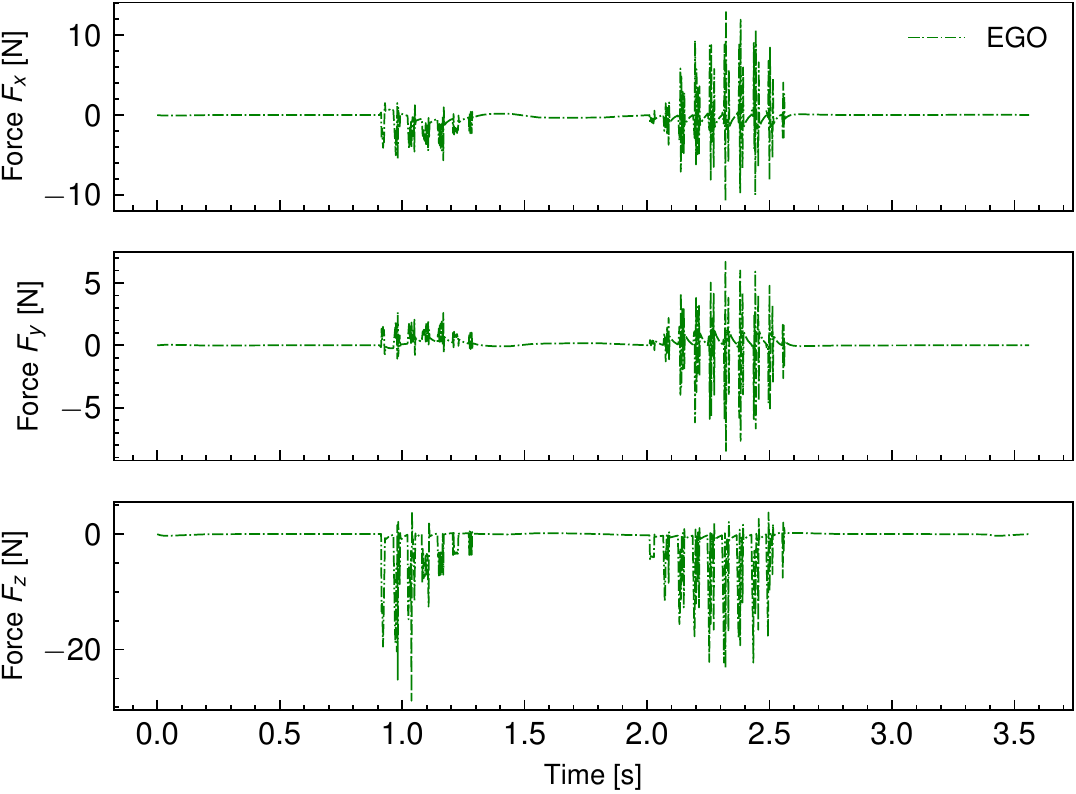}
        \caption{Force, EGO}
        \label{fig:Results-Obs-Expt-1-Force-EGO}
    \end{subfigure}
    \begin{subfigure}[t]{0.49\columnwidth}
        \centering\includegraphics[width=\textwidth]{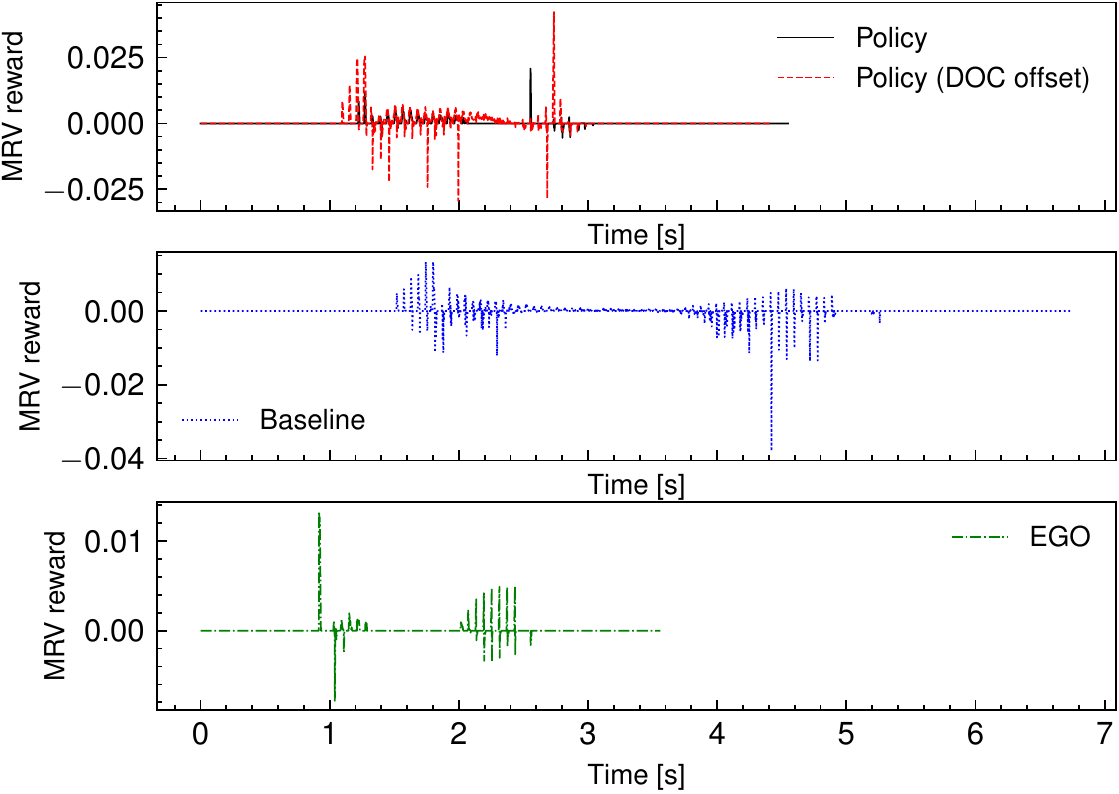}
        \caption{MRV, Top: policy, Middle: baseline, Bottom: EGO}
        \label{fig:Results-Obs-Expt-1-MRV-Policies}
    \end{subfigure}
    \begin{subfigure}[t]{0.49\columnwidth}
        \centering\includegraphics[width=\textwidth]{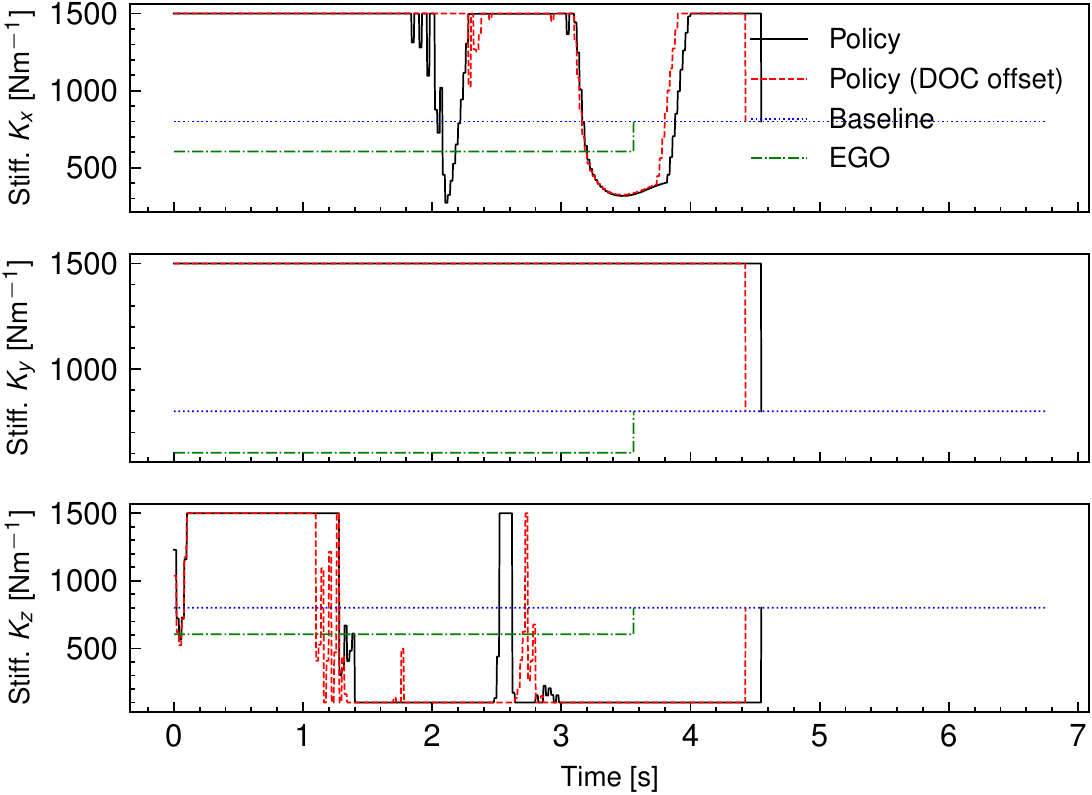}
        \caption{Implemented stiffness actions, all}
        \label{fig:Results-Stiffness-Expt-1}
    \end{subfigure}
	\caption[Observations during rollout for cutting of material (in experiment 1).]{Observations during rollout for cutting of material ($\vec{K}_{c}=\begin{bmatrix}718.7 & 839.9 & 0.03656\end{bmatrix}^{\tpose}$Nm$^{-2}$, $\vec{K}_{e}=\begin{bmatrix}8.337 & 0.4894 & -0.009854\end{bmatrix}^{\tpose}$ Nm$^{-1}$).}
	\label{fig:Results-Obs-Expt-1}
\end{figure}
\begin{figure}
	\centering
     \begin{subfigure}[t]{0.49\columnwidth}
        \centering\includegraphics[width=\textwidth]{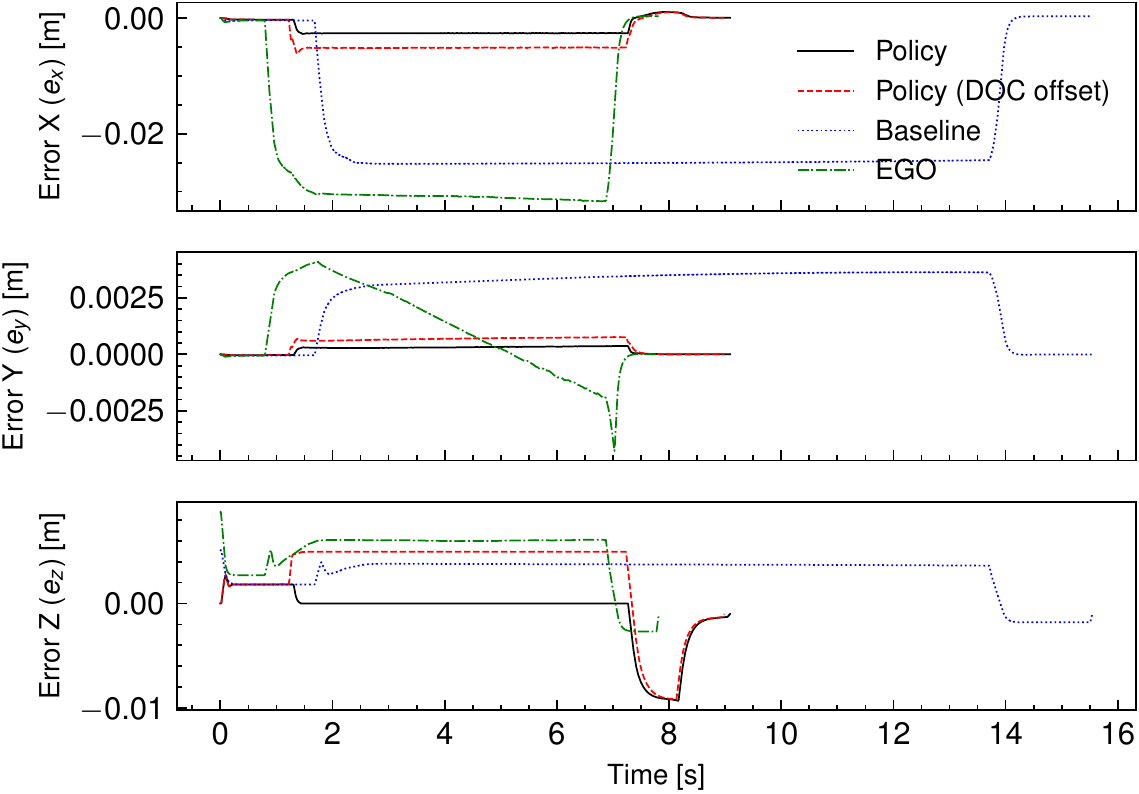}
        \caption{Path error, all}
        \label{fig:Results-Obs-Expt-2-Error}
    \end{subfigure}
    \begin{subfigure}[t]{0.49\columnwidth}
        \centering\includegraphics[width=\textwidth]{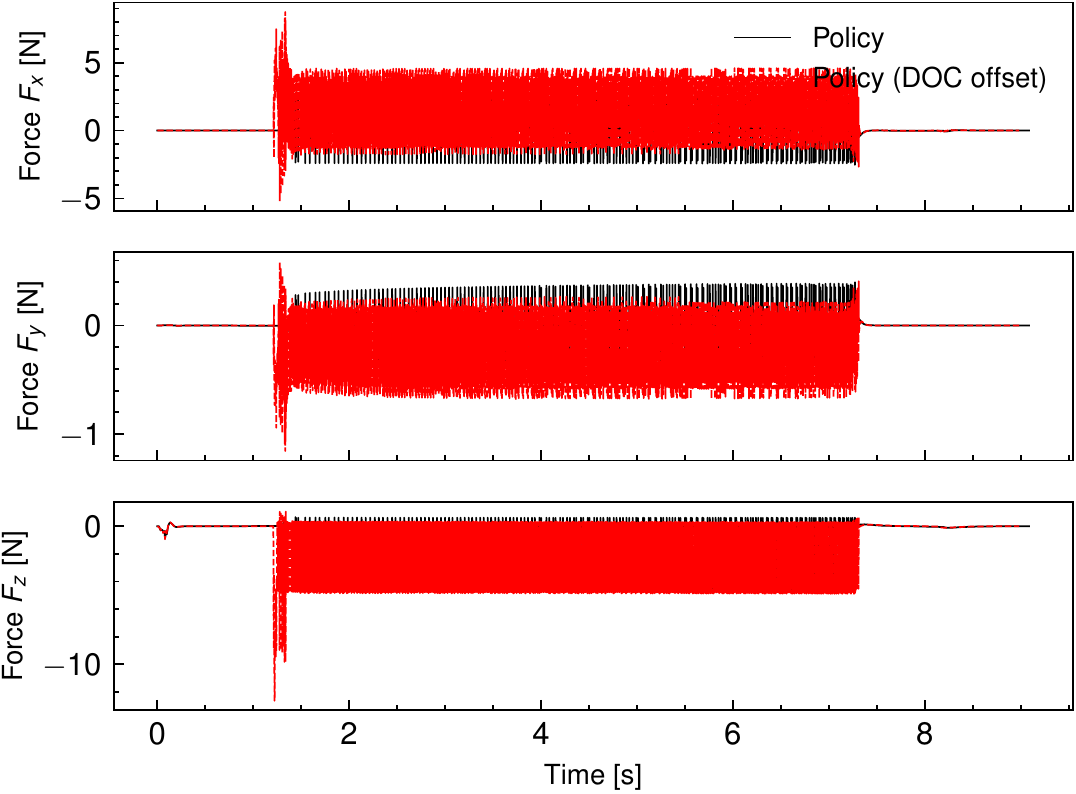}
        \caption{Force, policies}
        \label{fig:Results-Obs-Expt-2-Force-Policies}
    \end{subfigure}
    \begin{subfigure}[t]{0.49\columnwidth}
        \centering\includegraphics[width=\textwidth]{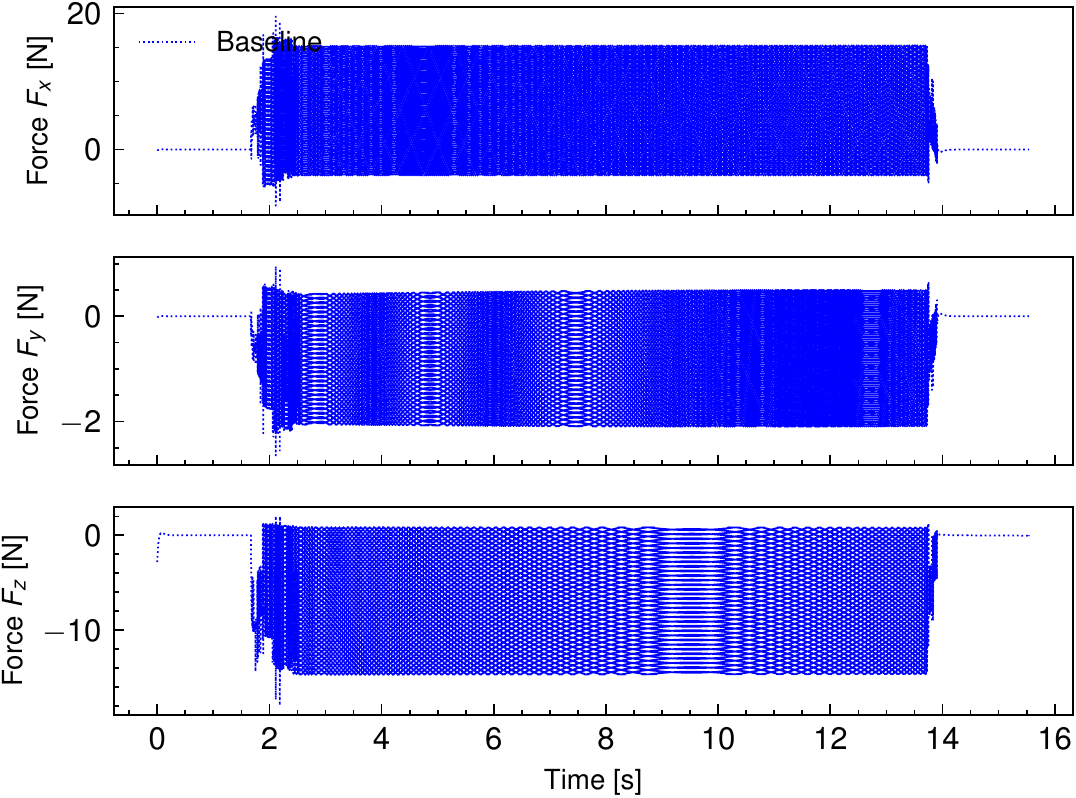}
        \caption{Force, baseline}
        \label{fig:Results-Obs-Expt-2-Force-Baseline}
    \end{subfigure}
    \begin{subfigure}[t]{0.49\columnwidth}
        \centering\includegraphics[width=\textwidth]{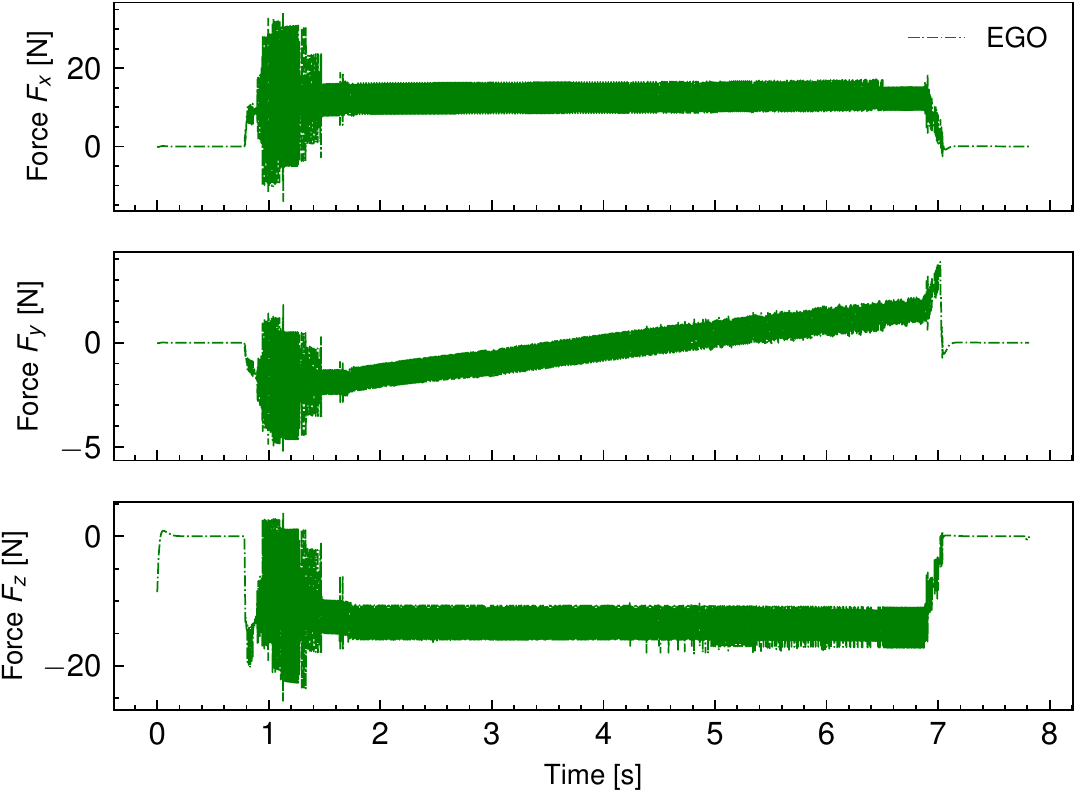}
        \caption{Force, EGO}
        \label{fig:Results-Obs-Expt-2-Force-EGO}
    \end{subfigure}
    \begin{subfigure}[t]{0.49\columnwidth}
        \centering\includegraphics[width=\textwidth]{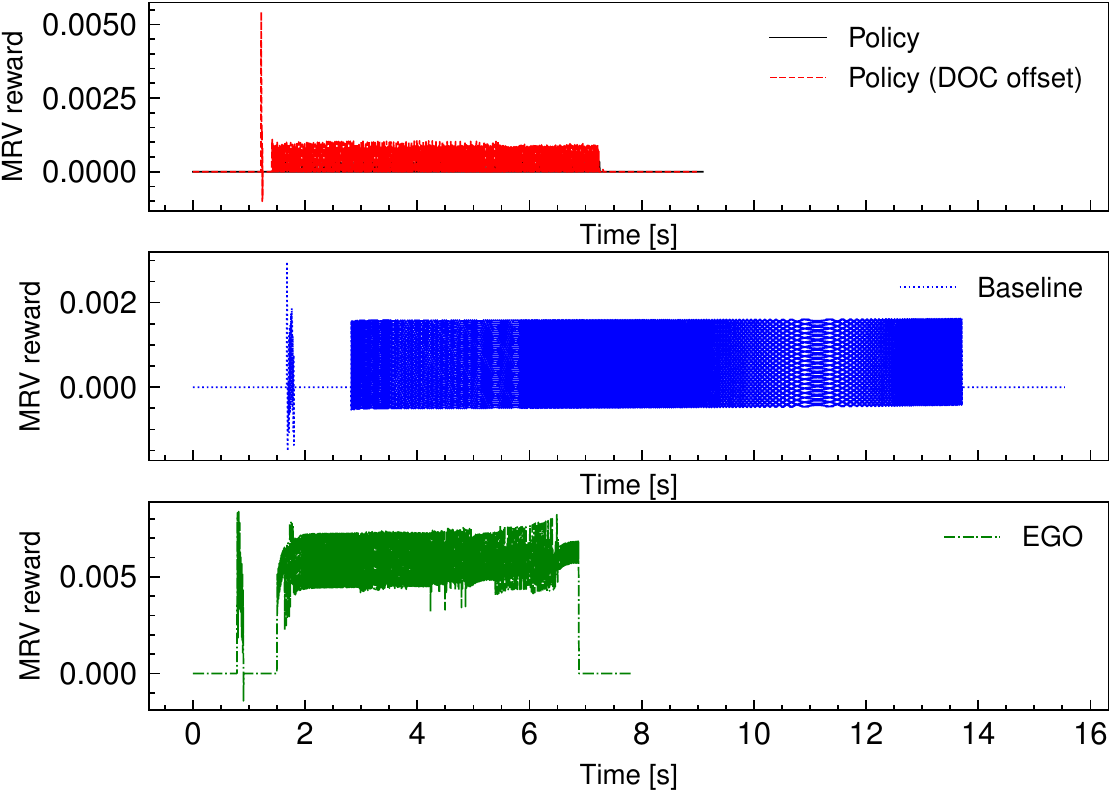}
        \caption{MRV, Top: policy, Middle: baseline, Bottom: EGO}
        \label{fig:Results-Obs-Expt-2-MRV-Policies}
    \end{subfigure}
    \begin{subfigure}[t]{0.49\columnwidth}
        \centering\includegraphics[width=\textwidth]{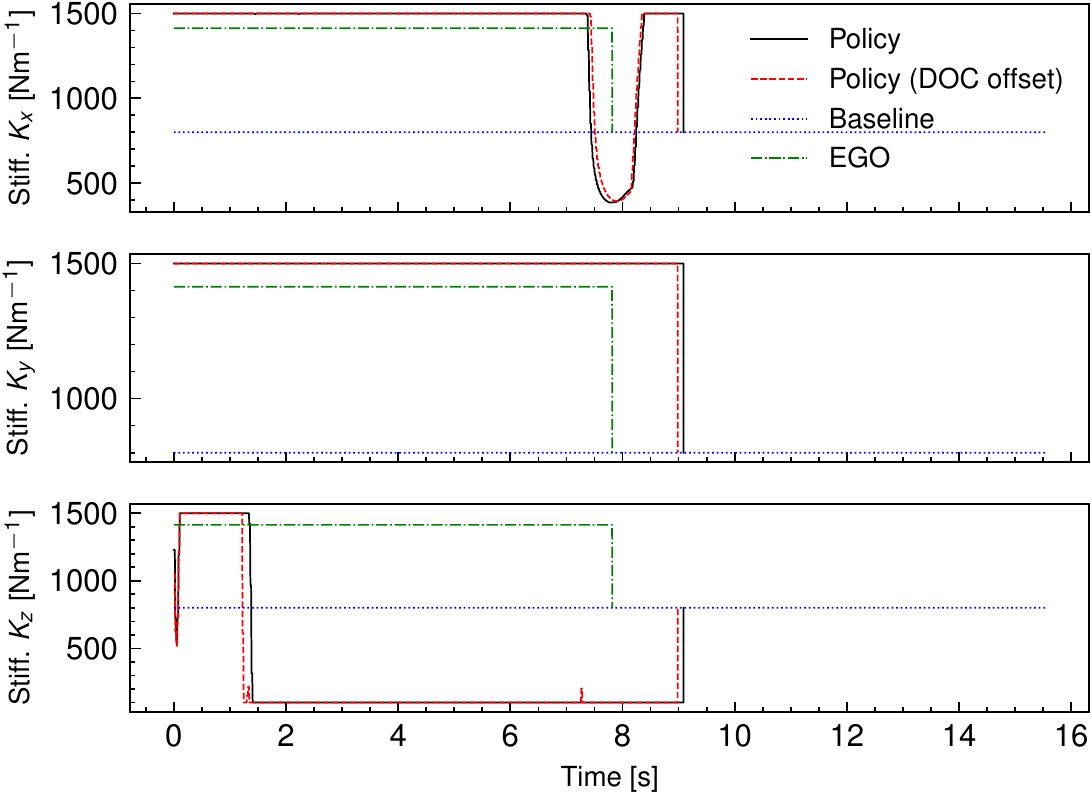}
        \caption{Implemented stiffness actions, all}
        \label{fig:Results-Stiffness-Expt-2}
    \end{subfigure}
	\caption[Observations during rollout for cutting of material (in experiment 2).]{Observations during rollout for cutting of material ($\vec{K}_{c}=\begin{bmatrix}368.4 & 759.6 & 0.03994\end{bmatrix}^{\tpose}$Nm$^{-2}$, $\vec{K}_{e}=\begin{bmatrix}3.306 & 3.509 & -0.007470\end{bmatrix}^{\tpose}$ Nm$^{-1}$).}
	\label{fig:Results-Obs-Expt-2}
\end{figure}
\begin{figure}
	\centering
      \begin{subfigure}[t]{0.49\columnwidth}
        \centering\includegraphics[width=\textwidth]{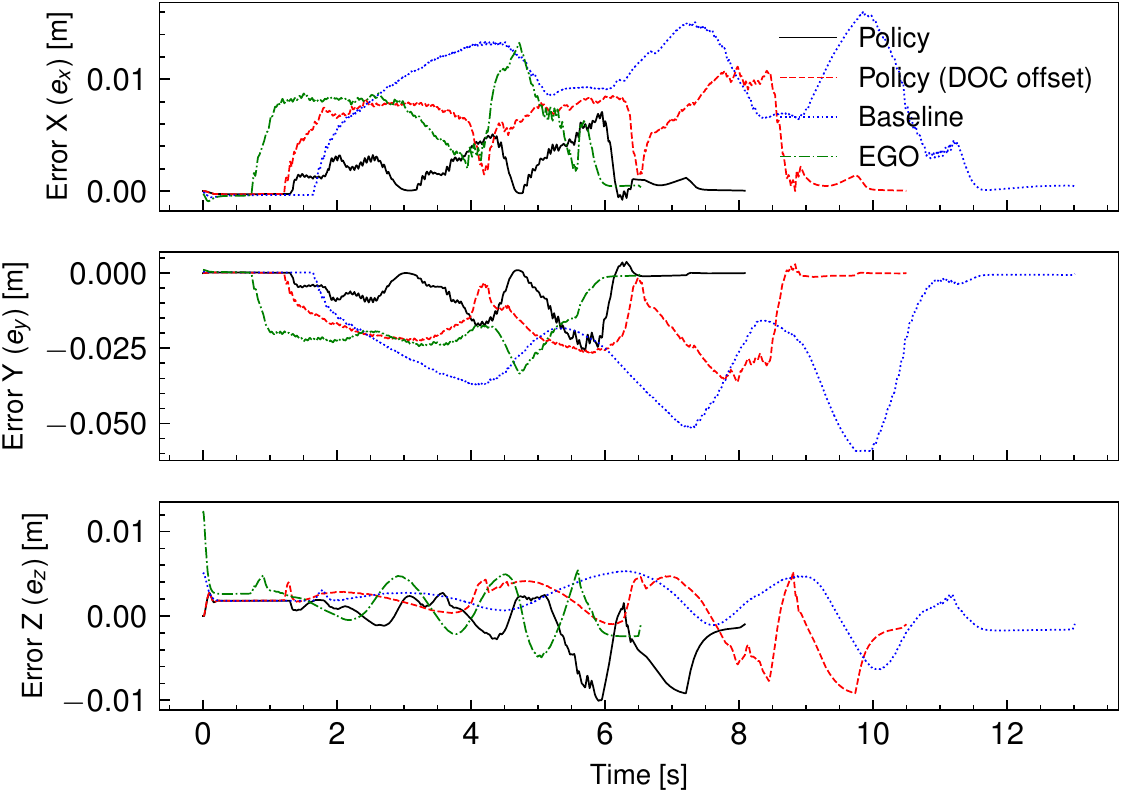}
        \caption{Path error, all}
        \label{fig:Results-Obs-Expt-3-Error}
    \end{subfigure}
    \begin{subfigure}[t]{0.49\columnwidth}
        \centering\includegraphics[width=\textwidth]{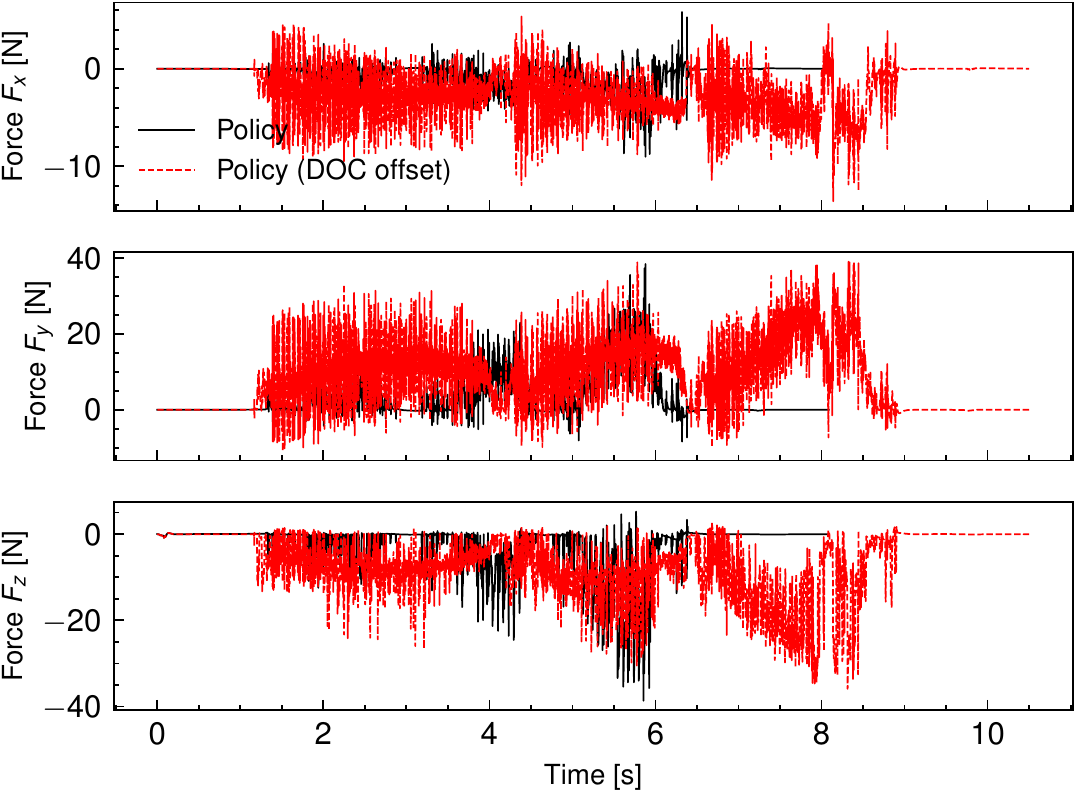}
        \caption{Force, policies}
        \label{fig:Results-Obs-Expt-3-Force-Policies}
    \end{subfigure}
    \begin{subfigure}[t]{0.49\columnwidth}
        \centering\includegraphics[width=\textwidth]{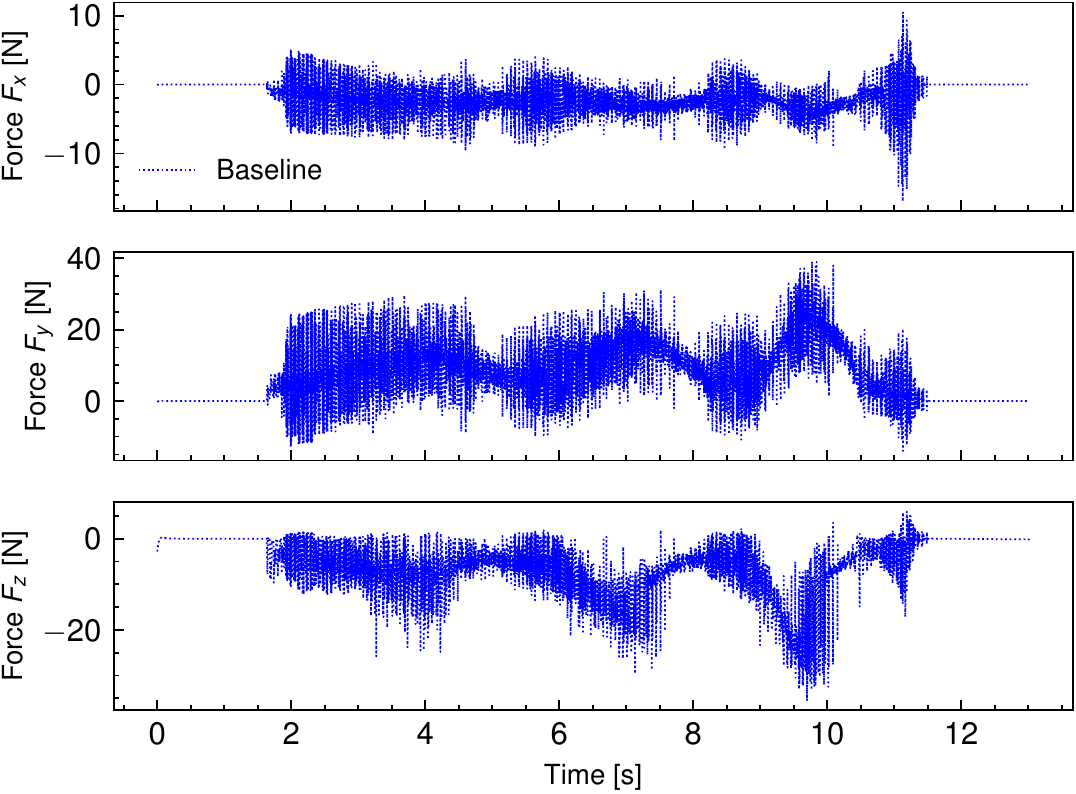}
        \caption{Force, baseline}
        \label{fig:Results-Obs-Expt-3-Force-Baseline}
    \end{subfigure}
    \begin{subfigure}[t]{0.49\columnwidth}
        \centering\includegraphics[width=\textwidth]{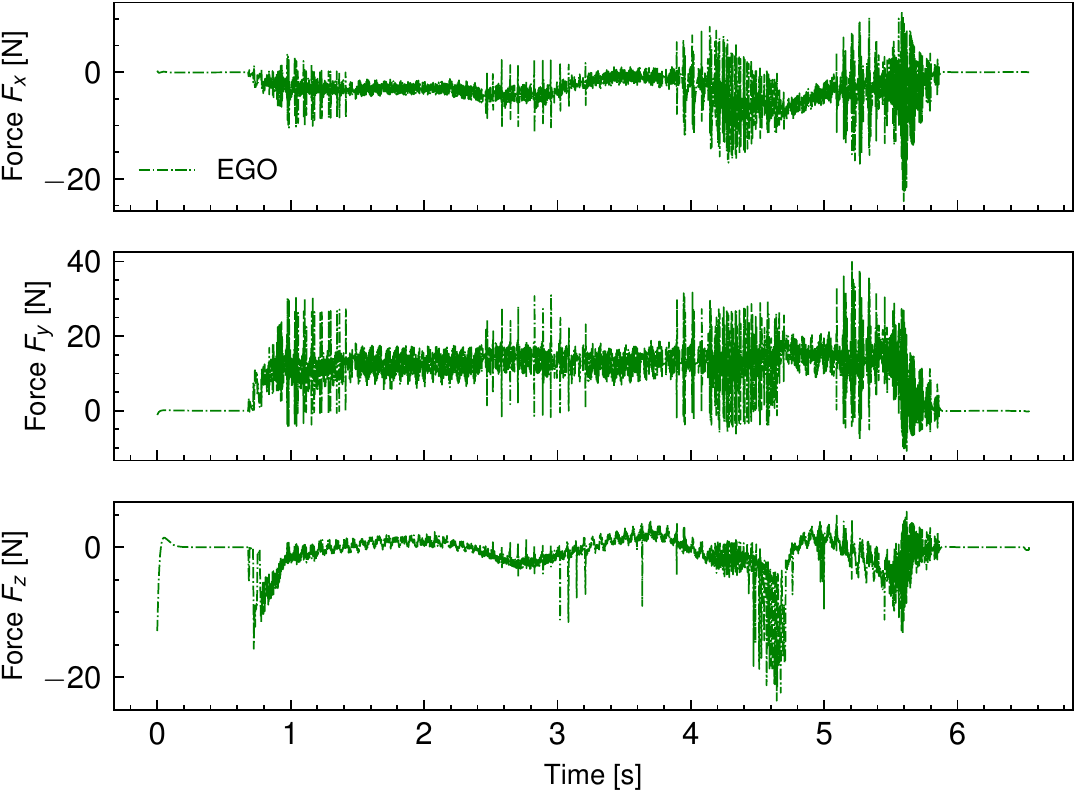}
        \caption{Force, EGO}
        \label{fig:Results-Obs-Expt-3-Force-EGO}
    \end{subfigure}
    \begin{subfigure}[t]{0.49\columnwidth}
        \centering\includegraphics[width=\textwidth]{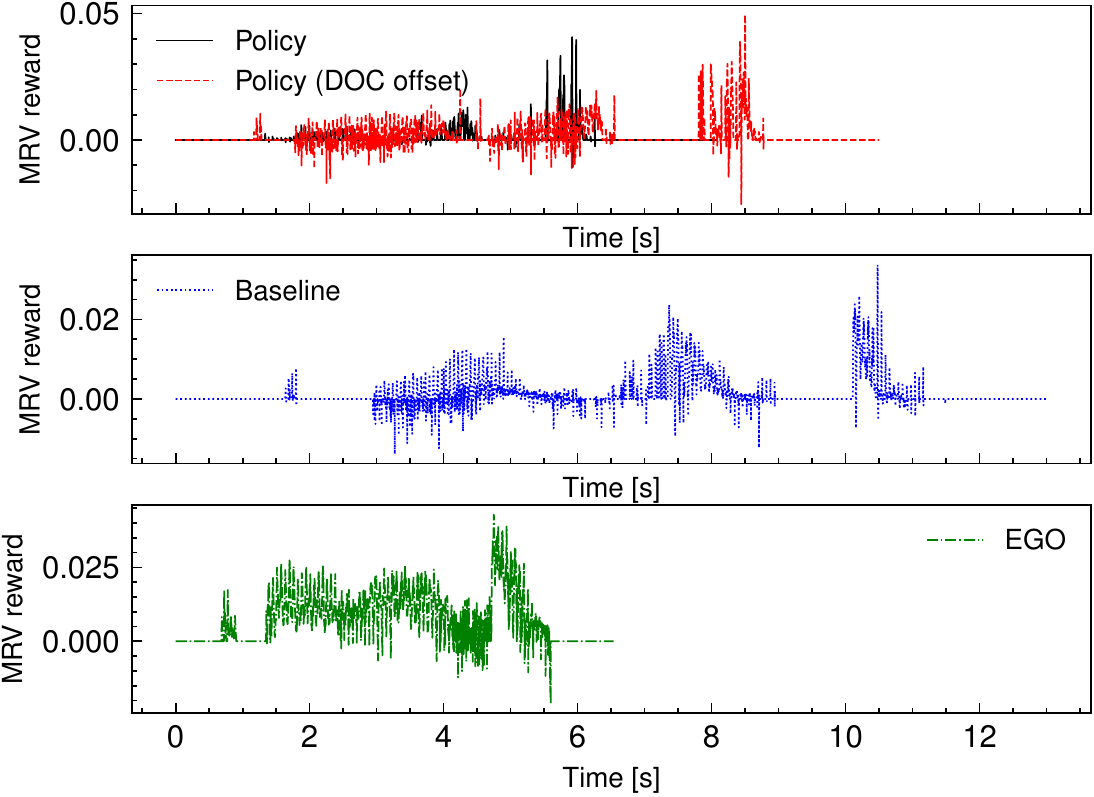}
        \caption{MRV, Top: policy, Middle: baseline, Bottom: EGO}
        \label{fig:Results-Obs-Expt-3-MRV-Policies}
    \end{subfigure}
    \begin{subfigure}[t]{0.49\columnwidth}
        \centering\includegraphics[width=\textwidth]{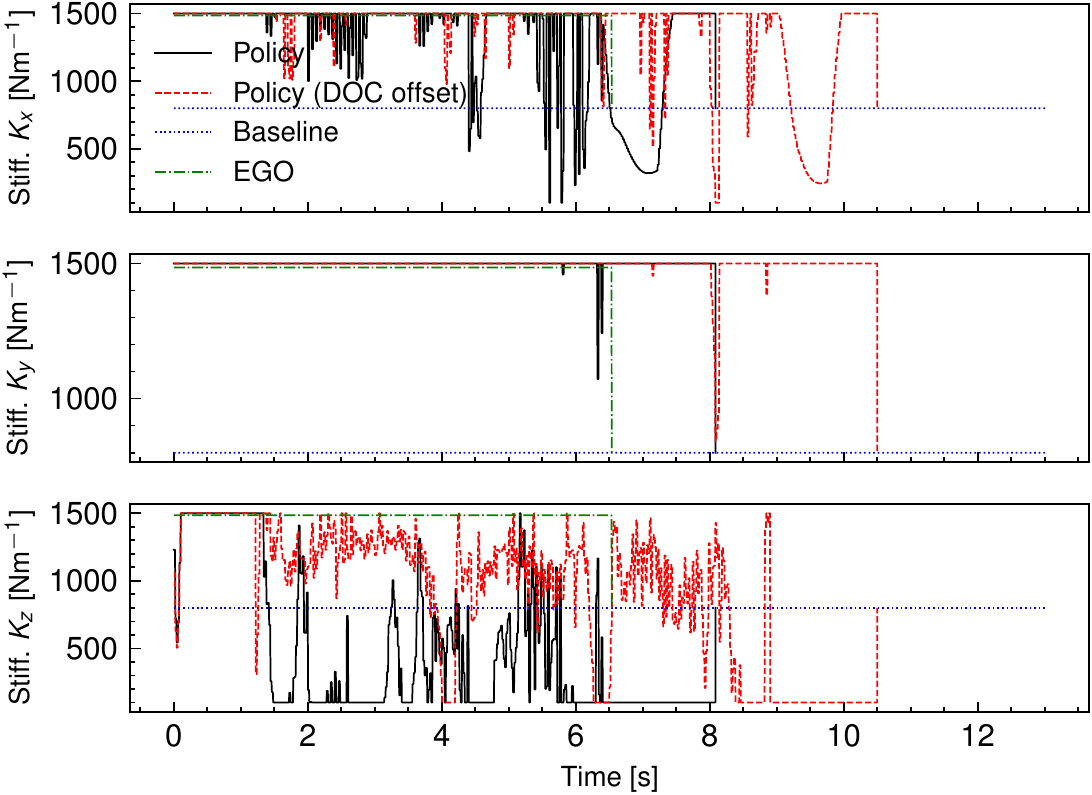}
        \caption{Implemented stiffness actions, all}
        \label{fig:Results-Stiffness-Expt-3}
    \end{subfigure}
	\caption[Observations during rollout for cutting of material (in experiment 3).]{Observations during rollout for cutting of material ($\vec{K}_{c}=\begin{bmatrix}343.7 & 788.0 & -0.04609\end{bmatrix}^{\tpose}$Nm$^{-2}$, $\vec{K}_{e}=\begin{bmatrix}9.203 & 3.984 & -0.0005049\end{bmatrix}^{\tpose}$ Nm$^{-1}$).}
	\label{fig:Results-Obs-Expt-3}
\end{figure}
\begin{figure}
	\centering
     \begin{subfigure}[t]{0.49\columnwidth}
        \centering\includegraphics[width=\textwidth]{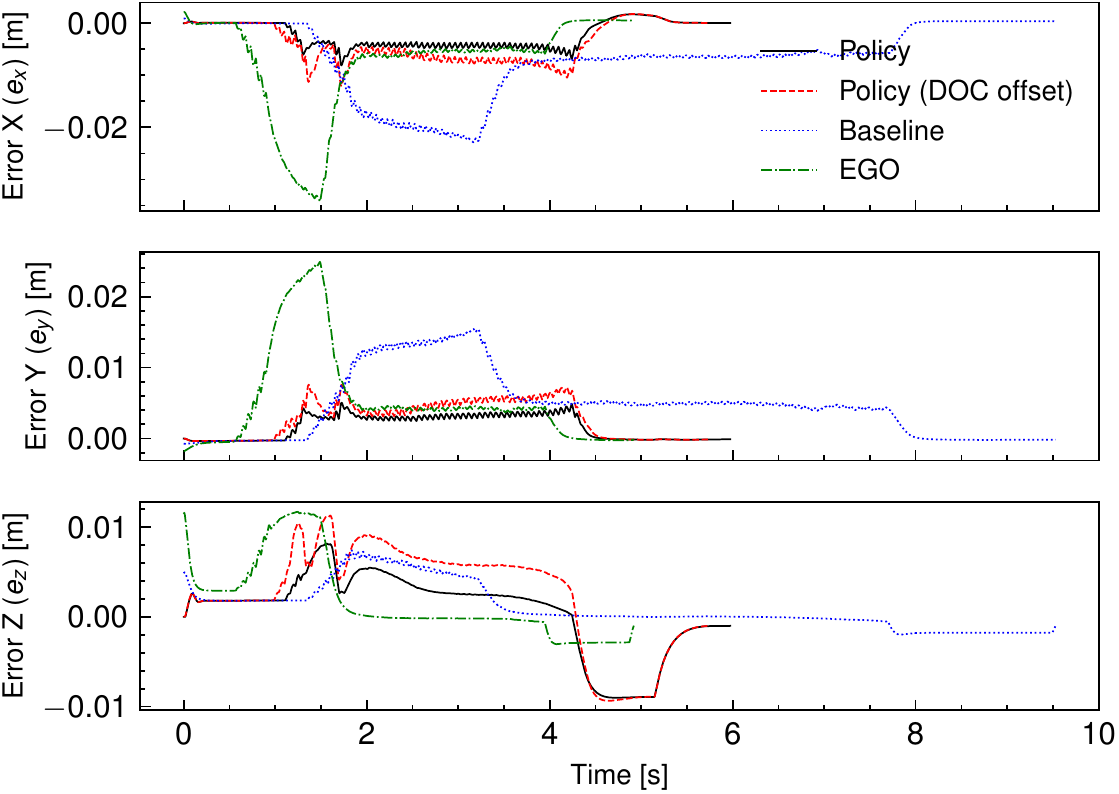}
        \caption{Path error, all}
        \label{fig:Results-Obs-Expt-4-Error}
    \end{subfigure}
    \begin{subfigure}[t]{0.49\columnwidth}
        \centering\includegraphics[width=\textwidth]{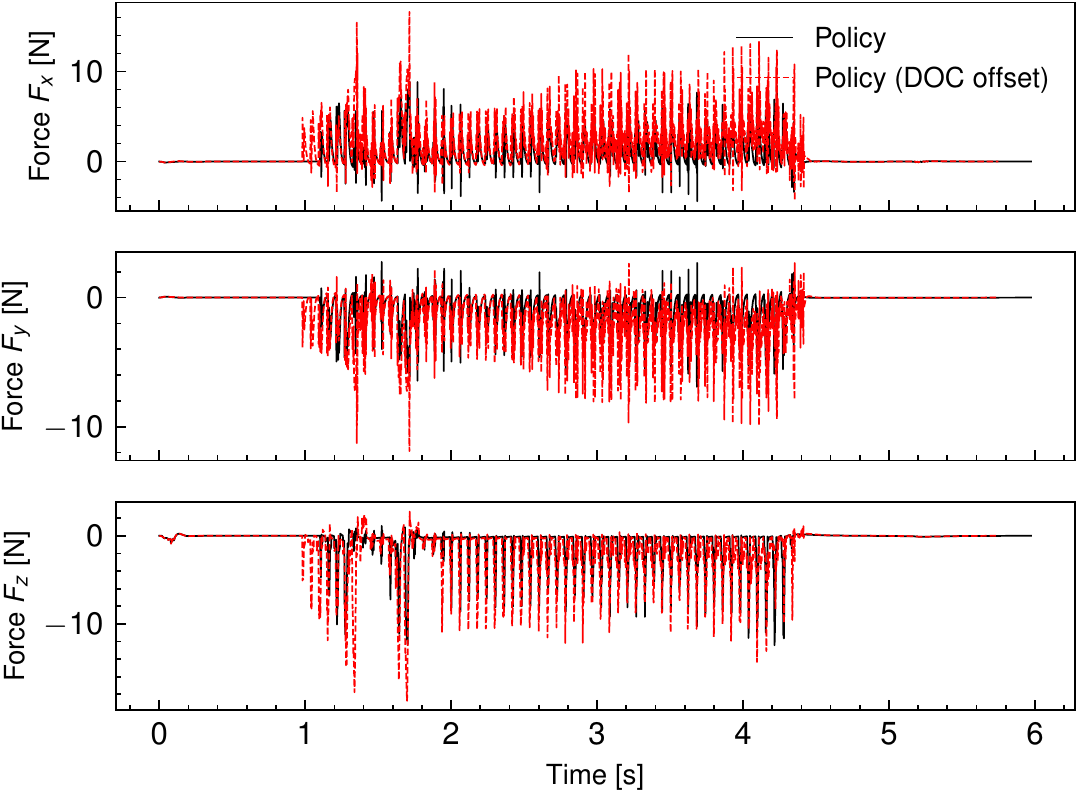}
        \caption{Force, policies}
        \label{fig:Results-Obs-Expt-4-Force-Policies}
    \end{subfigure}
    \begin{subfigure}[t]{0.49\columnwidth}
        \centering\includegraphics[width=\textwidth]{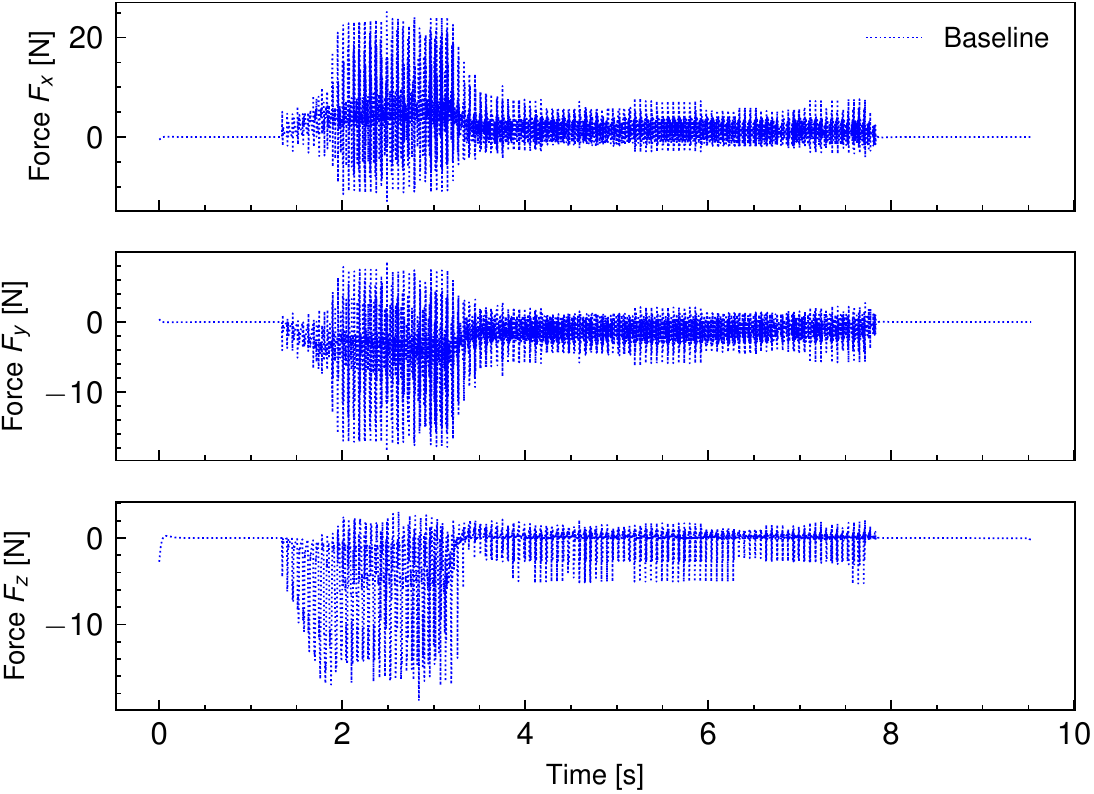}
        \caption{Force, baseline}
        \label{fig:Results-Obs-Expt-4-Force-Baseline}
    \end{subfigure}
    \begin{subfigure}[t]{0.49\columnwidth}
        \centering\includegraphics[width=\textwidth]{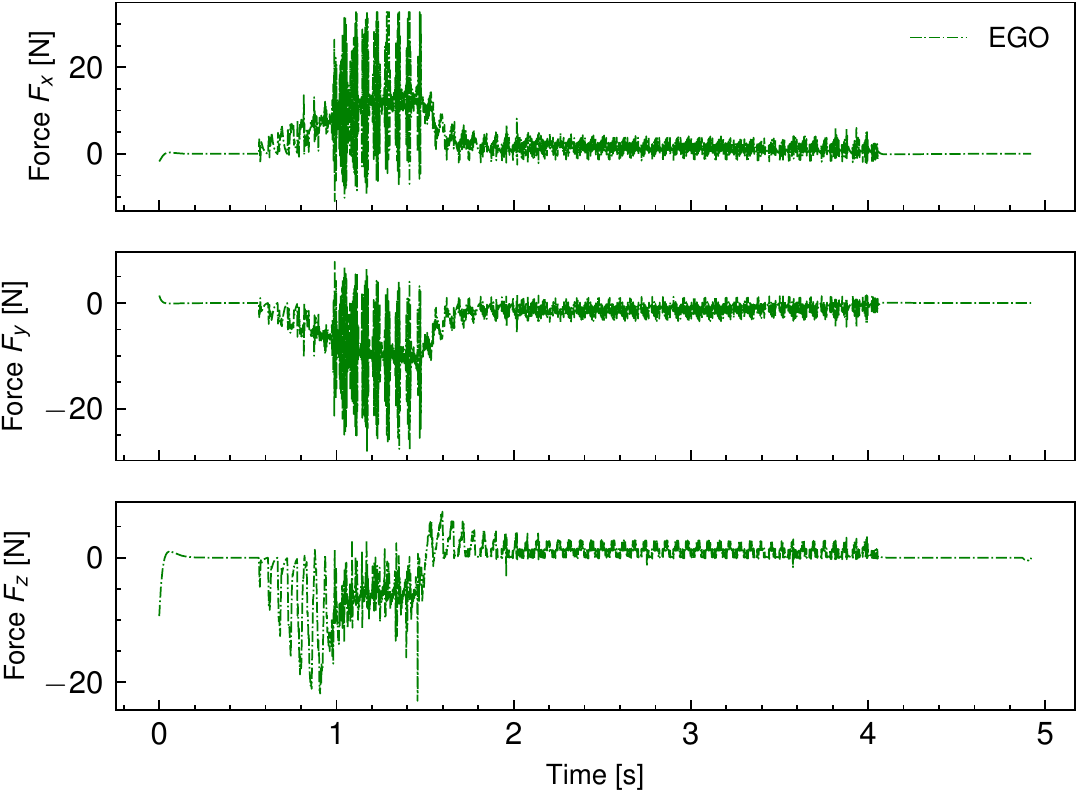}
        \caption{Force, EGO}
        \label{fig:Results-Obs-Expt-4-Force-EGO}
    \end{subfigure}
    \begin{subfigure}[t]{0.49\columnwidth}
        \centering\includegraphics[width=\textwidth]{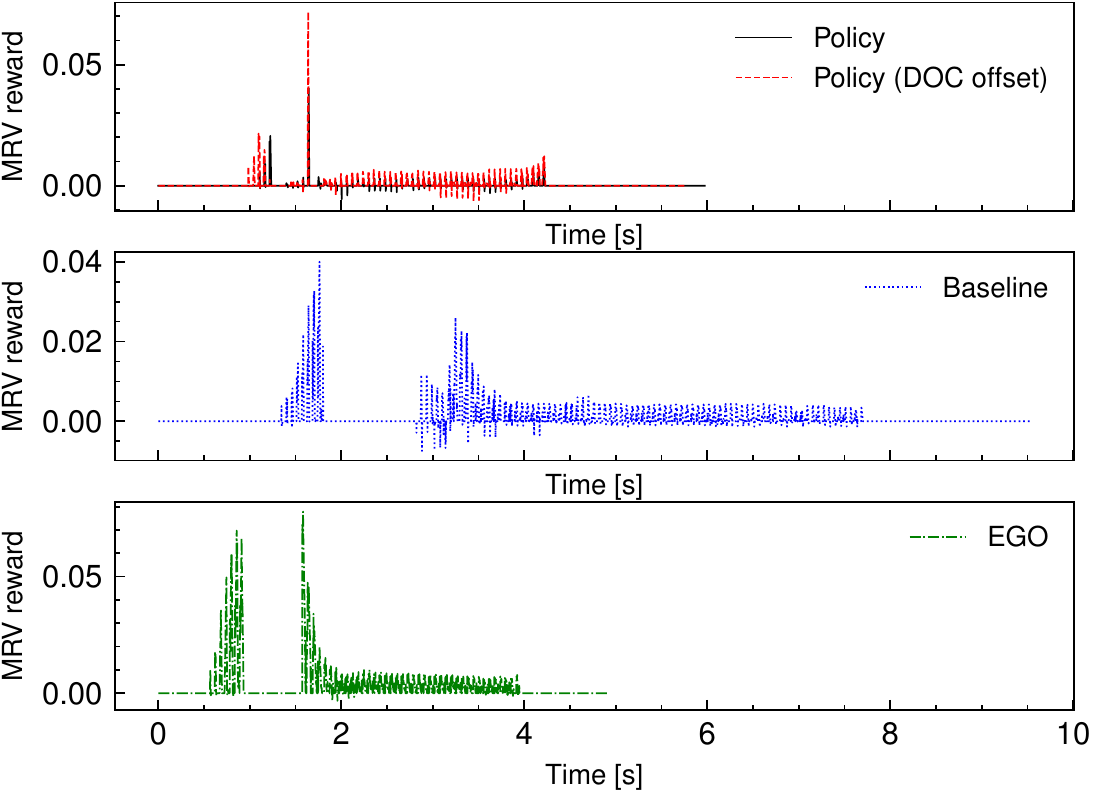}
        \caption{MRV, Top: policy, Middle: baseline, Bottom: EGO}
        \label{fig:Results-Obs-Expt-4-MRV-Policies}
    \end{subfigure}
    \begin{subfigure}[t]{0.49\columnwidth}
        \centering\includegraphics[width=\textwidth]{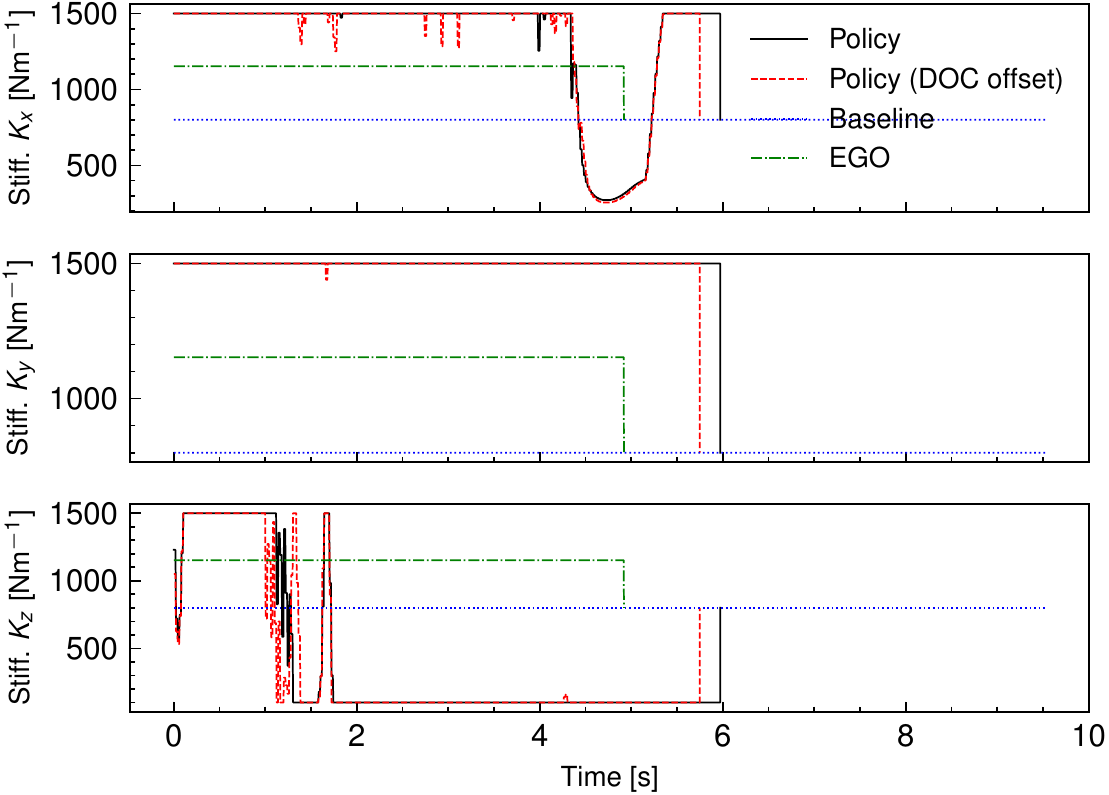}
        \caption{Implemented stiffness actions, all}
        \label{fig:Results-Stiffness-Expt-4}
    \end{subfigure}
	\caption[Observations during rollout for cutting of material (in experiment 4).]{Observations during rollout for cutting of material ($\vec{K}_{c}=\begin{bmatrix}463.4 & 997.7 & 0.09269\end{bmatrix}^{\tpose}$Nm$^{-2}$, $\vec{K}_{e}=\begin{bmatrix}3.253 & 6.923 & 0.0002610\end{bmatrix}^{\tpose}$ Nm$^{-1}$).}
	\label{fig:Results-Obs-Expt-4}
\end{figure}

Figures \ref{fig:Results-Reward-Contour-Expt-1}, \ref{fig:Results-Reward-Contour-Expt-2}, \ref{fig:Results-Reward-Contour-Expt-3}, \ref{fig:Results-Reward-Contour-Expt-4} show the estimated negative reward (cost) surface contour for four selected rollouts, showing the resultant optimal process parameters. The evolution of parameter selection of the policy during each rollout is overlaid as a quiver plot. Figures \ref{fig:Results-Obs-Expt-1}, \ref{fig:Results-Obs-Expt-2}, \ref{fig:Results-Obs-Expt-3}, \ref{fig:Results-Obs-Expt-4} show the path deviation, force, material removal and controller stiffness for the selected rollouts, comparing between the described parameter selection strategies. Comparison of these figures indicates the selection of feed rate is close to the optimal behaviour over all case studies. The selected feed rates correspond closely with regions of low cost, corroborated by the similar task duration of the policy and EGO over all trials. Comparing the tracking performance between each strategy indicates improved tracking of the desired path in the X and Y directions, while increasing tracking error in the normal direction.

The selection of RDOC by the learned policy particularly differs from the optimal behaviour, assuming a low depth of cut throughout each task, however, outperforms a naive selection of DOC throughout all tasks. From the partial dependence plots for depth of cut in Figure \ref{fig:Results-Reward-Contour-Expt-1}, \ref{fig:Results-Reward-Contour-Expt-2}, it is clear that lower depths of cut are favoured in these scenarios to minimise process force. Although some deviation is observed in Figure \ref{fig:Results-Reward-Contour-Expt-4}, corresponding to the cost surface favouring higher RDOC, the implemented behaviour is still highly conservative. This is particularly apparent in Figure \ref{fig:Results-Obs-Expt-2}, where the user RDOC offset is required to achieve a similar level of material removal as the baseline. The propensity to favour low RDOC throughout the selected case studies suggests the susceptibility of the method to local minima, contrasting with EGO. Further examination of the DOC variation over the selected rollouts reveals a loop-like structure in which the DOC is rapidly increased. Throughout each operation, small deviations from the path are observed. Close to the end of the path, this results in a ``corner-cutting'' behaviour, where the controller withdraws from the surface before reaching the path endpoint. Hence, the loop-like RDOC structure suggests a compensatory strategy for the compliance characteristics of the robot, allowing better coverage of the entire desired cutting path. Thus, in spite of the conservative RDOC behaviour, the selection of RDOC offset by the agent may be a useful strategy to compensate for path planning errors, e.g. in the case of damaged / deformed components. This informs applications for a baseline parameter adjustment method or means of data collection with reduced user intervention.

Overall, a key advantage of the proposed method is that the performance of a milling task can be improved in isolation without any prior knowledge. Since no rollouts are computed in advance as required with the EGO approach, the computational overhead is much lower. Nonetheless, the overall greater performance of EGO, coupled with the results for the policy with DOC offset may suggest a hybrid strategy based on combination of offline process modelling, with recourse to the learned parameter selection strategy where data are unavailable. This would allow such datasets to be constructed and added to a global knowledge base over time, reflecting the approaches in \cite{CognitiveRoboticsBasic, CognitiveRoboticsRevision} for disassembly applications. An alternative could be to employ EGO as a supervisor to accelerate the learning process and alleviate the problem of local minima, while preserving some of our method's advantages.

\section{Conclusion\label{sec:Conclusion}}

We propose a novel learning-based approach to milling parameter selection based on a mechanistic milling force simulation. We demonstrate the replicability of real-world results in the simulation environment and successful learning of a variable operational space control (OSC) policy over a wide distribution of materials and surface profiles. We furthermore address the issue of stability for variable OSC policies using the concept of energy tanks (ETs). Although the general concept of ET control is not new, this work solves relevant problems for reinforcement-learning-based interaction control and demonstrates the applicability of ET-OSC to already trained policies. For the simulated milling task, a favourable comparison with a constant parameter benchmark and greatly improved task consistency implies the generality of the proposed approach. Although an efficient global optimisation (EGO) strategy based on prior knowledge outperforms the proposed method, our approach has reduced computational overhead, and is independent of prior knowledge, material properties and workpiece geometry, informing potential applications as a conservative baseline parameter adjustment method or means of data collection for unknown components with reduced user intervention. Future work will demonstrate generality of our approach to real-world robotic cutting demonstrations using a variety of different rotary contact tools. Furthermore, while this work demonstrates generalisation across a range of single-component materials, the extension to complex products comprising multiple layers or composite materials could be investigated. Moreover, an online framework could use a combination of our approach for online parameter adjustment, guided by EGO where material models are available, with recourse to the proposed method where these are unavailable.

\bibliographystyle{IEEEtran}
\bibliography{main}

\begin{IEEEbiography}[{\includegraphics[width=1in,height=1.25in,clip,keepaspectratio]{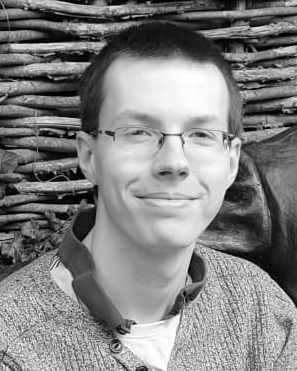}}]{Jamie Hathaway}
Jamie Hathaway is an PhD candidate at the University of Birmingham as part of the Faraday Institution RELIB (Reuse and Recycling of Lithium-ion Batteries) project. He received the MEng degree in Nuclear Engineering in 2020 from the University of Birmingham. His research interests primarily focus on data-driven methods for modelling and intelligent robotic control, and their application to contact-rich tasks. His current work focuses on learning and demonstration-based methods to develop generalised interaction control strategies for robotising the process of battery pack disassembly.
\end{IEEEbiography}

\begin{IEEEbiography}[{\includegraphics[width=1in,height=1.25in,clip,keepaspectratio]{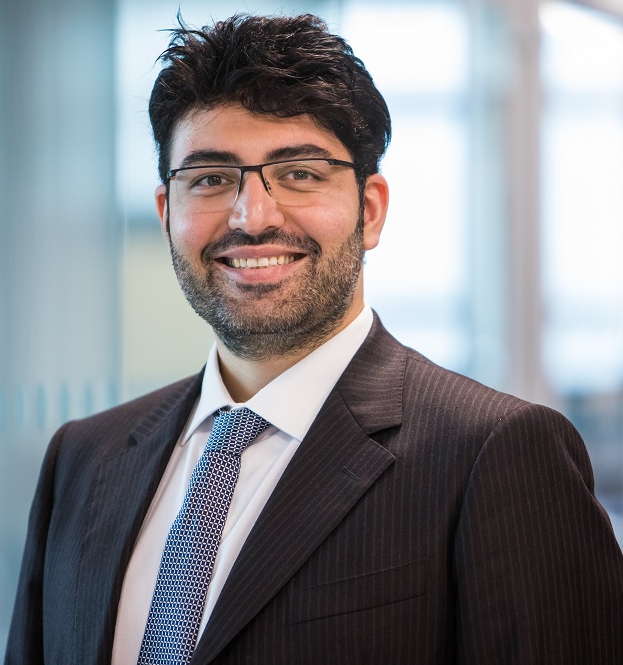}}]{Alireza Rastegarpanah}
Dr Rastegarpanah is a senior robotic researcher at the Faraday Institution-University of Birmingham who is leading the robotics team at a project called "Reuse and Recycling of Lithium-ion batteries (RELIB)" with a focus on automating the process of testing, disassembly and sorting Electrical Vehicle Lithium-ion batteries using advanced robotics and AI techniques. Dr Alireza Rastegarpanah is an interdisciplinary engineer with diverse research interests broadly centres on robotics, physical human-robot interaction, machine vision, machine learning and robotic manipulation. Currently his research comprises two main streams: (i) developing adaptive learning-based control strategies for disassembly of complex products, and (ii) Developing Neural Network models for predicting the state of health of EV batteries.
\end{IEEEbiography}

\begin{IEEEbiography}
[{\includegraphics[width=1in,height=1.25in,clip,keepaspectratio]{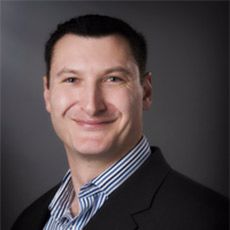}}]{Rustam Stolkin}
Rustam Stolkin received the M.Eng. degree in engineering science from the University of Oxford, Oxford, U.K. in 1998, and the Ph.D. degree in computer vision from the University College London, London, U.K., in 2004.
He is the Chair of Robotics with the University of Birmingham, Birmingham, U.K.; the Royal Society Industry Fellow; the Chair of the Expert Group on Robotic and Remote Systems for the OECD’s Nuclear Energy Agency; and founded the U.K. National Centre for Nuclear Robotics in 2017. His research interests include computer vision and image processing, machine learning and AI, robotic grasping and manipulation, and human–robot interaction.
\end{IEEEbiography}
\vfill

\end{document}